%% file: arxiv.tex
\documentclass{article}

\usepackage{graphicx}
\usepackage{booktabs} 
\usepackage{amsmath,amsfonts}
\usepackage{mathtools}
\usepackage{subcaption}

\usepackage{hyperref}

\usepackage[accepted]{icml2019}

\newcommand{\bx}{\mathbf{x}}
\newcommand{\bX}{\mathbf{X}}

\newcommand{\cI}{\mathcal{I}}

\newcommand{\cK}{\mathcal{K}}

\newcommand{\cS}{\mathcal{S}}
\newcommand{\cY}{\mathcal{Y}}

\newcommand{\hy}{\hat{y}}

\newcommand\given[1][]{\:#1\vert\:}

\icmltitlerunning{Generalized Linear Rule Models}

\begin{document}

\twocolumn[
\icmltitle{Generalized Linear Rule Models}



\icmlsetsymbol{equal}{*}

\begin{icmlauthorlist}
\icmlauthor{Dennis Wei}{IBM}
\icmlauthor{Sanjeeb Dash}{IBM}
\icmlauthor{Tian Gao}{IBM}
\icmlauthor{Oktay G\"{u}nl\"{u}k}{IBM}
\end{icmlauthorlist}

\icmlaffiliation{IBM}{IBM Research, Yorktown Heights, NY, USA}

\icmlcorrespondingauthor{Dennis Wei}{dwei@us.ibm.com}

\icmlkeywords{Machine Learning, ICML}

\vskip 0.3in
]



\printAffiliationsAndNotice{}  

\begin{abstract}
This paper considers generalized linear models using rule-based features, also referred to as rule ensembles, for regression and probabilistic classification. Rules facilitate model interpretation while also capturing nonlinear dependences and interactions. Our problem formulation accordingly trades off rule set complexity and prediction accuracy. Column generation is used to optimize over an exponentially large space of rules without pre-generating a large subset of candidates or greedily boosting rules one by one. The column generation subproblem is solved using either integer programming or a heuristic optimizing the same objective. In experiments involving logistic and linear regression, the proposed methods obtain better accuracy-complexity trade-offs than existing rule ensemble algorithms. At one end of the trade-off, the methods are competitive with less interpretable benchmark models.
\end{abstract}

\section{Introduction}
\label{sec:intro}

Decision rules have served as important building blocks for supervised learning models.  
They are often combined via logical operations into decision lists \cite{rivest1987,angelino2017,yang2017} and rule sets \cite{lakkaraju2016,wang2017}, models whose appeal stems from the human-interpretability of their constituent rules. 

In this paper, we consider models that are linear combinations of decision rules, also referred to as rule ensembles, within the framework of generalized linear models (GLM) \cite{mccullagh1989generalized}.  Rule ensembles retain interpretability while allowing modelling flexibility since rules are able to capture nonlinear dependences and interactions.  Our problem formulation accordingly trades off predictive accuracy against the complexity of the rule ensemble, measured in terms of the number of rules as well as their lengths, i.e.,~the number of elementary conditions on individual variables in the conjunction.  We make use of GLMs to address the real-valued prediction tasks of regression and probabilistic classification, where class probability estimates are desired.

The main challenge in learning rule-based models is due to the exponential size of the space of rules, corresponding to all possible conjunctions of the input features.  Predominant approaches include pre-selecting a (often large) subset of candidate rules and greedy optimization in which rules are added one by one but not revised.  The former includes optimization methods that select from among the candidates \cite{friedman2008,lakkaraju2016,wang2017} while the latter includes sequential covering \cite{cohen1995,clark1991,fuernkranz2012} and boosting.

Our main contribution herein is to propose an approach that avoids both of the above alternatives.  The technique of column generation (CG) is used to intelligently search the space of rules and produce useful ones as needed, as opposed to using a large, fixed set of candidates.  Instead of boosting, a GLM is re-fit as rules are generated, which allows existing rules to be reweighted and even discarded.  The CG subproblem is formulated as an integer program (not of exponential size) and is solved using either integer programming or a heuristic that targets the same objective.  We also discuss a non-CG algorithm that uses only first-degree rules, i.e.,~those with a single condition, and optionally numerical features as-is.  This latter algorithm produces a kind of generalized additive model (GAM) \cite{hastie1990}.

Experiments are presented involving the two most common cases of GLMs, logistic and linear regression.  The proposed methods are seen to yield better performance-complexity trade-offs than existing rule ensemble algorithms.  At the performance-maximizing end of the trade-off, the methods are competitive with less interpretable benchmark models such as tree ensembles and nonlinear support vector machines (SVM). The trade-off curves also suggest that substantially simpler models are often available at a relatively small cost in performance.

\subsection{Related Work}
\label{sec:intro:relWork}

Of the rule ensemble algorithms that have been proposed, the most closely related to the current proposal is RuleFit \cite{friedman2008}, which also fits linear models to a set of rules.  This set however is predetermined by first inducing a large number (hundreds) of decision trees from the data and then extracting rules corresponding to nodes of the trees.  As will be seen in Section~\ref{sec:eval}, this approach yields more complex rule ensembles for the same performance. \citet{rueckert2006} also propose fitting linear models to an increasing set of rules but the rules are generated in an unsupervised manner.

Boosting algorithms, which do not modify previously added rules, are represented by SLIPPER \cite{cohen1999}, MLRules and ENDER \cite{dembczynski2010}.  Other algorithms include HKL \cite{jawanpuria2011}, which is also able to effectively optimize over an exponential number of conjunctions but relies on a regularizer with a special group structure.  While often cited as a rule ensemble method, \citet{weiss2000} actually learn ensembles of disjunctive normal forms (DNF), which are more akin to tree ensembles.

Even though branch-and-bound and CG have been used before in ML, e.g.,~for boosting \cite{demiriz2002} and in particular for rule learning \cite{angelino2017, Rudin2018}, we believe our work is the first application of CG to a nonlinear optimization problem in ML.
Our CG approach is inspired by its recent use in learning  disjunctive/conjunctive normal form rules \cite{dash2018}, with similar benefits. While that work applied CG to a linear program, here we have a (convex) nonlinear problem. 
For nonlinear problems, even though the general framework has been discussed for OR problems \cite{Garcia2003}, there are only a few practical applications, mostly nonlinear variants of the VRP \cite{Borndorfer2013, Fortz2010}.
We also note that whether CG can be successfully applied to MINLPs has been posed as an open problem in a 2018 Dagstuhl Seminar \cite{Bonami2018}.

\section{Generalized Linear Rule Models}
\label{sec:prob}

We consider the standard supervised learning problem of predicting a target variable $Y \in \cY$ using input features $\bX = (X_1,\dots,X_d)$, given a training dataset of i.i.d.~samples $(\bx_i, y_i)$, $\bx_i = (x_{i1}, \dots, x_{id})$, $i=1,\dots,n$. The output space $\cY$ may be discrete or continuous.  We assume that all features $X_j$ have been binarized to take values in $\{0,1\}$.  For categorical features, this is achieved through the usual ``one-hot'' coding into indicators $X_j = x$ for all categories $x$ as well as their negations $X_j \neq x$.  Numerical features are binarized through bi-directional comparisons to a set of thresholds, e.g.,~$X_j \leq 1$, $X_j \leq 2.3$ and $X_j > 1$, $X_j > 2.3$.  Further details on binarization are given in Section~\ref{sec:eval}.

In this section, we recall aspects of generalized linear models (GLMs) and introduce notation needed to define them over a feature space of rules. 
Let $\mathcal{K}$ denote the set of conjunctions of $\bX$ to be considered.  
We defer discussion of the choice of $\cK$ to Section~\ref{sec:instants} but note that it is not necessary to limit its size. Denote by $A_k$ the variable corresponding to conjunction $k \in \cK$, and $a_{ik} \in \{0,1\}$ the value taken by $A_k$ in instance $i$.  Let $k=0$ be the index of the empty conjunction $A_0 \equiv 1$. 

For a GLM, 
we posit that $Y$ conditioned on $\bX$ follows an exponential family distribution given by 
\begin{equation}\label{eqn:pY}
p_{Y\given\bX}(y \given \bx) = h(y) \exp\left( \eta y - \Phi(\eta) \right),
\end{equation}
where the canonical parameter $\eta$ is a linear combination of the conjunctions $A_k$ of $\bX$, 
\begin{equation}\label{eqn:eta}
\eta = \sum_{k\in\cK} \beta_k A_k,
\end{equation}
and $\Phi(\eta)$ is the log-partition function. The terms with nonzero coefficients $\beta_k$ define an ensemble of rules mapping conjunctions to real values $\beta_k$, which are then linearly combined.  The distribution may have parameters in addition to $\eta$ (e.g.,~the variance in the Gaussian case) but these are either assumed known or their estimation can be separated from that of $\eta$.  The prediction function is given by the conditional mean of $Y$,
\begin{equation}\label{eqn:yhat}
\hy(\bX) = \mathbb{E}[Y\given\bX] = \Phi'(\eta),
\end{equation}
where the second equality holds for \eqref{eqn:pY}.

The coefficients $\beta_k$ in \eqref{eqn:eta} are determined by minimizing the negative log-likelihood corresponding to \eqref{eqn:pY} on the training data.  Since $\cK$ is potentially large, a sparse solution is essential and is obtained through $\ell_1$ regularization.  The optimization problem is therefore 
\begin{equation}\label{eqn:regLL}
    \min_{\beta} \;\; \frac{1}{n} \sum_{i=1}^n \left[ \Phi\left( \sum_{k\in\mathcal{K}} \beta_k a_{ik} \right) - y_i \sum_{k\in\mathcal{K}} \beta_k a_{ik} \right] + \sum_{k\in\mathcal{K}} \lambda_k \lvert\beta_k\rvert.
\end{equation}
The factor $h(y)$ in \eqref{eqn:pY} is not a function of $\beta$ and is omitted.  Each regularization parameter $\lambda_k$ depends on the number of literals of conjunction $k$ as specified later.  It is a property of exponential families that the log-partition function $\Phi(\eta)$ is convex.  By affine composition property of convex functions the problem \eqref{eqn:regLL} is therefore convex.

We specialize the foregoing to the two most common cases of GLMs, logistic and linear regression.  For logistic regression, the log-partition function $\Phi(\eta) = \log(1 + e^\eta)$.  Substituting this into \eqref{eqn:regLL}, the quantity in square brackets becomes the familiar expression
\begin{equation}\label{eqn:logR}
\log\Big( 1 + \exp\Big((-1)^{y_i} \sum_{k\in\mathcal{K}} \beta_k a_{ik}\Big) \Big),
\end{equation}
where $y_i \in \{0, 1\}$. For linear regression, $\Phi(\eta) = \eta^2 / 2$ and the bracketed quantity becomes (after adding back $y_i^2 / 2$)
\begin{equation}\label{eqn:linR}
\frac{1}{2}\Big( y_i - \sum_{k\in\mathcal{K}} \beta_k a_{ik} \Big)^2.
\end{equation}

\section{Model Instantiations}
\label{sec:instants}
We discuss two instantiations of generalized linear rule models (GLRM). In Section~\ref{sec:instants:LR1}, the set of conjunctions $\cK$ is restricted to first-degree or singleton conjunctions, i.e.,~those with a single condition on an individual feature. Section~\ref{sec:prob:CG} considers the general case with no restriction on $\cK$.

\subsection{Generalized Additive Model Using First-Degree Rules}
\label{sec:instants:LR1}

In the first case, the conjunctions $A_k$ correspond to the binarized features $X_j$ themselves. In terms of the original unbinarized features, conditions are placed on only one feature at a time and so the resulting GLRM is free of interaction terms.  On the other hand, if features are binarized as discussed at the beginning of Section~\ref{sec:prob}, then the GLRM is a type of generalized additive model (GAM), i.e.,~a sum of univariate functions.  For numerical features, first-degree rules correspond to step functions, which can be linearly combined into arbitrary piecewise-constant functions with discontinuities at the binarization thresholds. For categorical features, any function can be realized.

\citet{friedman2008} discuss a similar type of model obtained by restricting their decision trees to depth $1$ (decision stumps). There are two differences however. First, the singleton rules herein are systematically enumerated as opposed to generated by a randomized tree induction procedure.  Second, from every pair of complementary singleton rules (e.g.,~$X_j \leq 1$, $X_j > 1$), we remove one member as otherwise the pair together with the empty conjunction $A_0 \equiv 1$ are collinear. The results in Section~\ref{sec:eval} suggest that this removal of linearly dependent rules contributes toward sparser, simpler rule ensembles.

In addition to first-degree rules, following \citet{friedman2008} we may also include in the feature space any numerical features as they are, without binarization. This model variant is also evaluated in Section~\ref{sec:eval}.

\subsection{General Rule Ensemble Using Column Generation}
\label{sec:prob:CG}

We now let $\cK$ be the set of all possible conjunctions of $\bX$ to obtain rule ensembles with no restrictions. Since $\cK$ is now exponentially large, it is intractable even to enumerate the variables in problem \eqref{eqn:regLL} (unless the feature dimension $d$ is very small). We exploit the technique of column generation (CG) to tackle this problem.

Column generation was originally developed to solve linear programs (LPs) with a very large number of columns \cite{gilmore1961, IPref}.
Using the fact that optimal solutions of LPs are sparse, the main idea is to first solve a restricted problem with a small number of candidate columns and then generate some of the missing columns based on the optimal dual solution  of this restricted problem.
Using the dual solution, one can compute the marginal benefit (or, partial derivative) of introducing a missing column to the restricted problem. If  partial derivative for the most promising missing column is non-negative, then the procedure terminates. 
The crucial component of this approach is to formulate a column generation problem that can search through all of the missing columns 
without complete enumeration. 

We next describe how to adopt this idea to solve the generalized linear model \eqref{eqn:regLL}.
To derive the column generation subproblem, it is helpful to express $\beta_k$ as $\beta_k = \beta^+_k - \beta^-_k$ for $\beta^+_k$, $\beta^-_k \geq 0$. Problem \eqref{eqn:regLL} becomes 
\begin{multline}\label{eqn:regLLpm}
    \min_{\beta^+, \beta^- \geq 0} \;\; \frac{1}{n} \sum_{i=1}^n \left[ \Phi\left( \sum_{k\in\mathcal{K}} (\beta^+_k - \beta^-_k) a_{ik} \right)\right.\\ \left.{} - y_i \sum_{k\in\mathcal{K}} (\beta^+_k - \beta^-_k) a_{ik} \right] + \sum_{k\in\mathcal{K}} \lambda_k (\beta^+_k + \beta^-_k).    
\end{multline}

Suppose that a restricted version of \eqref{eqn:regLLpm} has been solved for a subset $\mathcal{S} \subset \mathcal{K}$ of the set of conjunctions, yielding $\beta^\pm_k = (\beta^\pm_k)^*$ for $k \in \mathcal{S}$.  We extend this to a solution for \eqref{eqn:regLLpm} by setting $\beta^\pm_k = 0$ for $k \notin \mathcal{S}$ and wish to determine the optimality of the extended solution. Since \eqref{eqn:regLLpm} is also a convex problem with non-negativity constraints, a necessary and sufficient condition of optimality is for the partial derivatives of the objective with respect to $\beta^\pm_k$ to be zero if $\beta^\pm_k > 0$ and non-negative if $\beta^\pm_k = 0$.  This condition is true for $k \in \mathcal{S}$ due to optimality for the restricted problem.  For $k \notin \mathcal{S}$, $\beta^\pm_k = 0$ and we are thus required to check non-negativity of the derivatives. 

The partial derivative w.r.t.~$\beta^+_k$ of the objective in \eqref{eqn:regLLpm} is
\begin{align*}
    &\frac{1}{n} \sum_{i=1}^n \left[ \Phi'\left( \sum_{k'\in\mathcal{K}} \beta_{k'} a_{ik'} \right) a_{ik} - y_i a_{ik} \right] + \lambda_k\\
    &\quad = \frac{1}{n} \sum_{i=1}^n \bigl(\hy(\bx_i) - y_i\bigr) a_{ik} + \lambda_k
    = \frac{1}{n} \sum_{i=1}^n r_i a_{ik} + \lambda_k,
\end{align*}
using \eqref{eqn:eta} and \eqref{eqn:yhat} to obtain the first equality and defining the prediction \emph{residual} $r_i = \hy(\bx_i) - y_i$ in the second.  The partial derivative with respect to $\beta^-_k$ is the same except that the residuals are negated.  Non-negativity of all partial derivatives can thus be determined by solving 
the pair of problems 
\begin{equation}\label{eqn:pricing1}
    \min_{k\in\mathcal{K}} \; \pm \frac{1}{n} \sum_{i=1}^n r_i a_{ik} + \lambda_k.
\end{equation}
If both optimal values in \eqref{eqn:pricing1} are non-negative, then all derivatives are indeed non-negative and we conclude that the extended solution is optimal for \eqref{eqn:regLLpm}.  On the other hand, if the objective in \eqref{eqn:pricing1} is negative for some $k \notin \mathcal{S}$ and the `$+$' sign (say), then the partial derivative w.r.t.~$\beta^+_k$ is negative and $\beta^+_k$ can be added to the restricted problem (i.e.,~$k$ added to $\mathcal{S}$) to potentially improve the current solution.

We now use the fact that $\mathcal{K}$ is a set of conjunctions to avoid solving \eqref{eqn:pricing1} by enumeration.  This involves encoding a conjunction by the set of participating features and relating the values $a_{ik}$ to the feature values $x_{ij}$.  Let $z_j \in \{0,1\}$ represent whether feature $j$ is selected in a conjunction.  We assume that the regularization parameter $\lambda_k$ is an affine function of the degree of the conjunction, $\lambda_k = \lambda_0 + \lambda_1 \sum_j z_j$ with $\lambda_0$, $\lambda_1 \geq 0$  
(other affine functions of $z_j$ are possible).  
Define $\bar{x}_{ij} = 1 - x_{ij}$, $\mathcal{I}_+ = \{i: r_i > 0\}$, and $\mathcal{I}_- = \{i: r_i < 0\}$.  Then \eqref{eqn:pricing1} can be reformulated as 
\begin{equation}\label{eqn:pricing2}
    \begin{split}
        \min_{a,z} \quad &\pm \frac{1}{n} \sum_{i=1}^n r_i a_i + \lambda_0 + \lambda_1 \sum_{j=1}^d z_j\\
        \text{s.t.} \quad &a_i + \sum_{j=1}^d \bar{x}_{ij} z_j \geq 1, \quad a_i \geq 0, \qquad i \in \mathcal{I}_+\\
        &a_i + z_j \leq 1, \qquad i \in \mathcal{I}_-, \quad j : \bar{x}_{ij} = 1\\
        &z_j \in \{0,1\}, \qquad j = 1,\dots,d,
    \end{split}
\end{equation}
where the subscript $k$ has been dropped in favor of encoding by $\{z_j\}$.  The constraints in \eqref{eqn:pricing2} ensure that $a_i$ acts as the conjunction of the selected $x_{ij}$'s.  For $i \in \mathcal{I}_+$, $a_i = 1$ only if all selected features have $x_{ij} = 1$ ($\bar{x}_{ij} z_j = 0 \; \forall j$), otherwise $a_i = 0$ since $r_i a_i$ is minimized in the objective.  For $i \in \mathcal{I}_-$, $r_i < 0$, $a_i$ is maximized, and the corresponding constraint enforces the same behavior for $a_i$.

{
Therefore, our column generation algorithm alternates between solving the restricted log-likelihood problem \eqref{eqn:regLL} and searching for new columns by solving \eqref{eqn:pricing2} for both signs.  We initialize the restricted set $\cS$ to be the set of first-degree rules discussed in Section~\ref{sec:instants:LR1}, optionally including original numerical features as well. 
The algorithm terminates with a certificate that problem \eqref{eqn:regLL} is solved to optimality if the optimal value of 
problem \eqref{eqn:pricing2} is  non-negative for both signs.
For practical reasons we also have a secondary termination criteria that depends on the total number of column generation iterations and the total CPU time spent.

This algorithm is guaranteed to terminate in {\em finite} time as there are only a finite number of candidate columns (conjunctions) and the column generation procedure would not generate the same conjunction more than once as the partial derivative of the conjunctions in the restricted problem are guaranteed to be non-negative.
We note that finite termination is not guaranteed for models with different regularization parameters, for example for $\ell_2$-regularization, as repeating a conjunction with a non-zero weight would improve the optimal value of \eqref{eqn:regLLpm} by simply splitting the weight of an original conjunction into two equal parts. After splitting the first term in \eqref{eqn:regLLpm} would stay the same whereas the regularization penalty would decrease.

Once the column generation algorithm terminates, we solve the  log-likelihood problem \eqref{eqn:regLL} one last time to de-bias the solution.
In this final run we restrict conjunctions to the ones with $\beta^\pm_k>10^{-5}$ in the last round and we drop the regularization term in the objective.
}

\section{Column Generation Approaches}
\label{sec:CG}
The column generation subproblem \eqref{eqn:pricing2} is solved using either integer programming (IP) or a heuristic algorithm. 
{
We have implemented our column  generation procedure in Java using LibLinear \cite{REF08a} to solve the regularized logistic regression problem \eqref{eqn:regLL} and using Cplex callable library (version 12.7.1)  to solve the integer program \eqref{eqn:pricing2} for column generation.
As LibLinear package only allows simple $\ell_1$-regularization (as opposed to using different weights $\lambda_k$ that depend on the complexity of the conjunction $k$), we scale the $a_{ik}$ values by $1/\lambda_k$.


} 

The proposed heuristic algorithm performs a limited search of the rule space, proceeding in order of increasing conjunction degree $D = \sum_j z_j$ from $1$ up to a maximum $D_{\max}$.  To describe the heuristic, we define the children of a conjunction as those conjunctions that involve one additional feature, i.e.,~have one additional $z_j = 1$ in terms of the representation in \eqref{eqn:pricing2}.  At each degree $D$, only those conjunctions that are children of a chosen parent (of degree $D-1$) are evaluated.  These children are evaluated for two purposes: 1) To determine whether any improve upon the incumbent solution, defined as the best solution observed thus far; 2) to select a parent conjunction for the next highest degree.  Evaluation for the first purpose is based only on the objective value achieved in \eqref{eqn:pricing2}, while evaluation for the second also considers a lower bound on future objective values, as discussed next.

We illustrate the objective value and lower bound calculations for degree $1$ conjunctions, i.e.,~children of the initial empty conjunction. Higher degrees are analogous.  Objective values of children can be computed via their increments and decrements relative to the value of the parent.  For $D = 1$, setting $z_j = 1$ forces $a_i = 0$ for $i$ such that $x_{ij} = 0$ ($\bar{x}_{ij} = 1$).  The change in value of child $j$ is thus 
\begin{equation}\label{eqn:objChange}
\Delta v(j) = \lambda_1 - \sum_{i\in\cI_+} r_i \bar{x}_{ij} - \sum_{i\in\cI_-} r_i \bar{x}_{ij}.
\end{equation}
A lower bound on future objective values resulting from setting $z_j = 1$, i.e.,~values of descendants of child $j$, can be obtained by optimistically assuming that with the addition of one more feature (at a further cost of $\lambda_1$), all positive $r_i$ can be eliminated from the objective function in \eqref{eqn:pricing2} while no negative $r_i$ are eliminated beyond those due to setting $z_j = 1$ itself. Expressed in the same relative terms as \eqref{eqn:objChange}, this lower bound is 
\begin{equation}\label{eqn:lowerBound}
\mathrm{LB}(j) = 2\lambda_1 - \sum_{i\in\cI_+} r_i - \sum_{i\in\cI_-} r_i \bar{x}_{ij}.
\end{equation}

To determine the parent for the next degree, we first eliminate all children of the current parent whose lower bounds $LB(j)$ are not less than the value of the incumbent solution, since these cannot lead to improvement.  Any remaining children are evaluated using the average of $\Delta v(j)$ and $\mathrm{LB}(j)$ and the child with the lowest such average is selected.  The motivation is to consider not only the children's current values but also a crude but easily computed estimate of potential future values.  Other convex combinations of $\Delta v(j)$ and $\mathrm{LB}(j)$ have not been explored.

Once a new parent conjunction is chosen, corresponding to setting a $z_j = 1$, two operations are performed to reduce the dimensions of the problem and to render it in the same form as for $D = 1$.  First, indices (rows) $i$ where $\bar{x}_{ij} = 1$ are removed since $a_i$ is forced to zero as noted above.  Second, setting $z_j = 1$ may make other features $j'$ redundant and these can be removed (by setting $z_{j'} = 0$). 
For example, if binary feature $j$ corresponds to one category of an original categorical feature, then we may set $z_{j'} = 0$ for other categories of the same feature because the conjunction of $j$ and $j'$ would be identically zero.  We refer to \citet{su2016} for a fuller discussion of these redundancies.


We have explored additional variations of the heuristic algorithm as discussed in Appendix~\ref{sec:heuristicVars}. 

\section{Numerical Evaluation}
\label{sec:eval}

We report on numerical experiments involving both logistic regression \eqref{eqn:logR} (i.e.,~classification) and linear regression \eqref{eqn:linR}.  We tested the $4$ methods discussed in Section~\ref{sec:instants}: logistic/linear regression on singleton rules (without CG, abbreviated LR1), logistic/linear regression on general rules (with CG, abbreviated LRR), and the same two with the addition of any numerical features originally present in the data (LR1N, LRRN). Column generation is done using the heuristic in Section~\ref{sec:CG} but we also report a preliminary result using an IP version of LRR (LRRI). 
We compared these methods to RuleFit \cite{friedman2008}, which is the closest existing method, 
and specifically a Python implementation\footnote{\url{https://github.com/christophM/rulefit}} because the original R code is no longer supported.  We also tried to test HKL \cite{jawanpuria2011} but encountered a no longer supported toolbox as well as numerical problems.  Beyond rule ensembles, we also compared to gradient boosted classification/regression trees (GBT) and support vector machines (SVM) with radial basis function (RBF) kernels.  These are less interpretable models intended to provide a benchmark for prediction performance.  $10$-fold cross-validation (CV) is used to estimate all test performance metrics.

Categorical and numerical features were binarized as described at the beginning of Section~\ref{sec:prob}, using sample deciles as thresholds for numerical features.  To control for the effect of discretization, numerical features were discretized using the same quantile thresholds for all rule- and tree-based methods in the comparison.  Excluded from this treatment are SVMs and the variant of RuleFit that uses numerical features in addition to rules (abbreviated RuleFitN).

\begin{figure*}[ht]
  \centering
  \begin{subfigure}[b]{0.7\columnwidth}
  \includegraphics[width=\columnwidth]{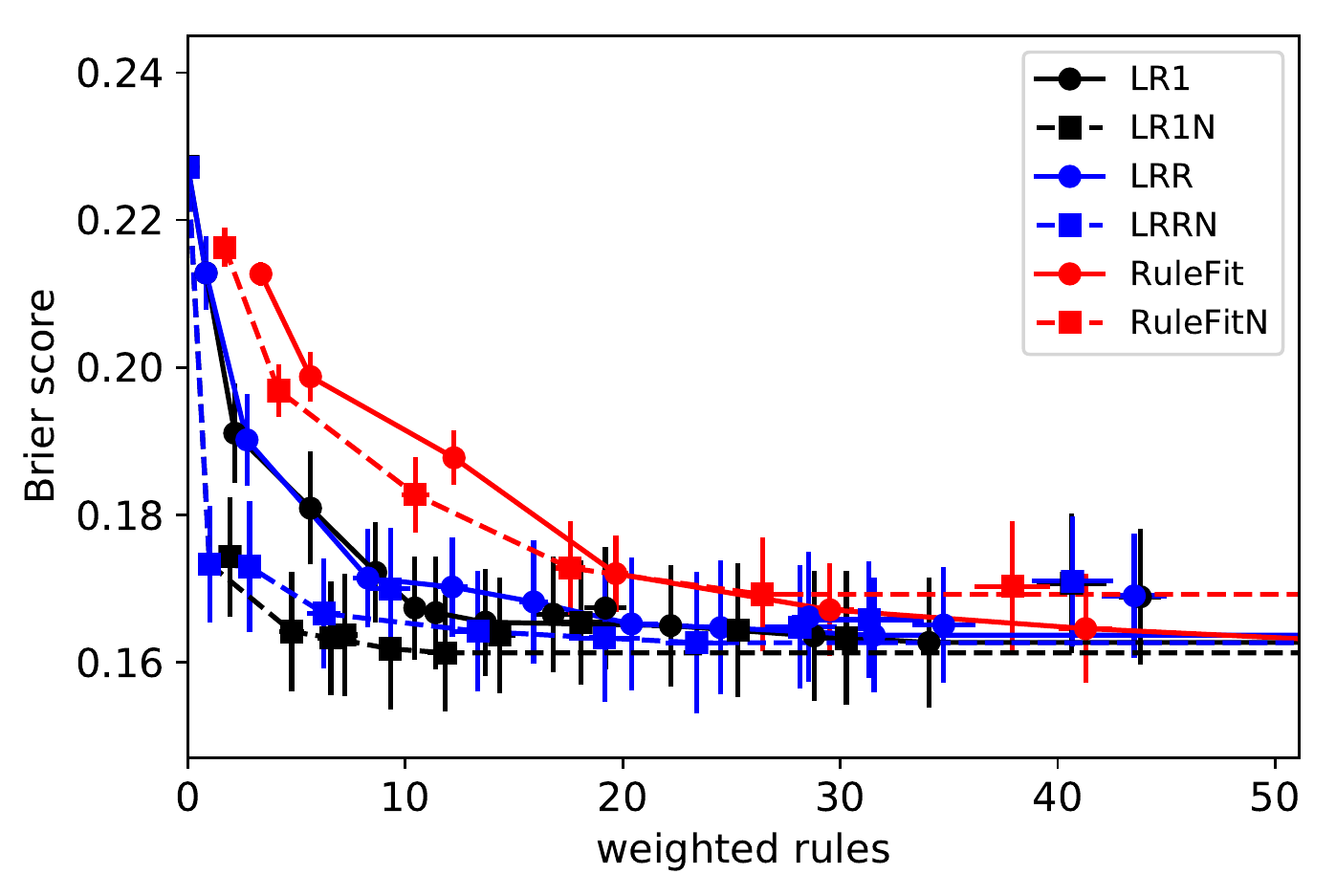}
  \caption{Pima Indians diabetes}
  \label{fig:paretoClassBrier:pima0}
  \end{subfigure}
  \begin{subfigure}[b]{0.7\columnwidth}
  \includegraphics[width=\columnwidth]{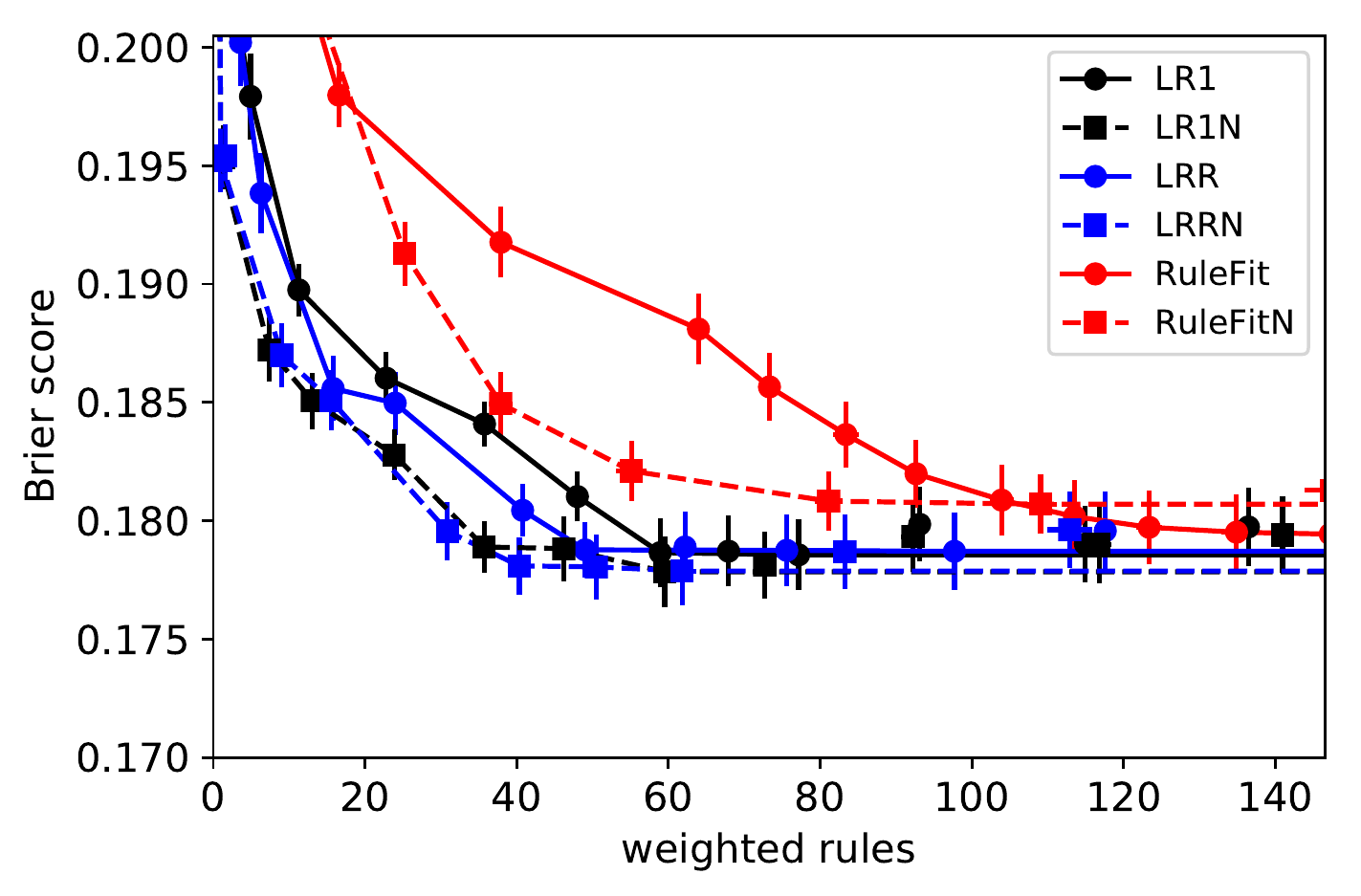}
  \caption{FICO Explainable Machine Learning Challenge}
  \label{fig:paretoClassBrier:FICO0}
  \end{subfigure}
  \begin{subfigure}[b]{0.7\columnwidth}
  \includegraphics[width=\columnwidth]{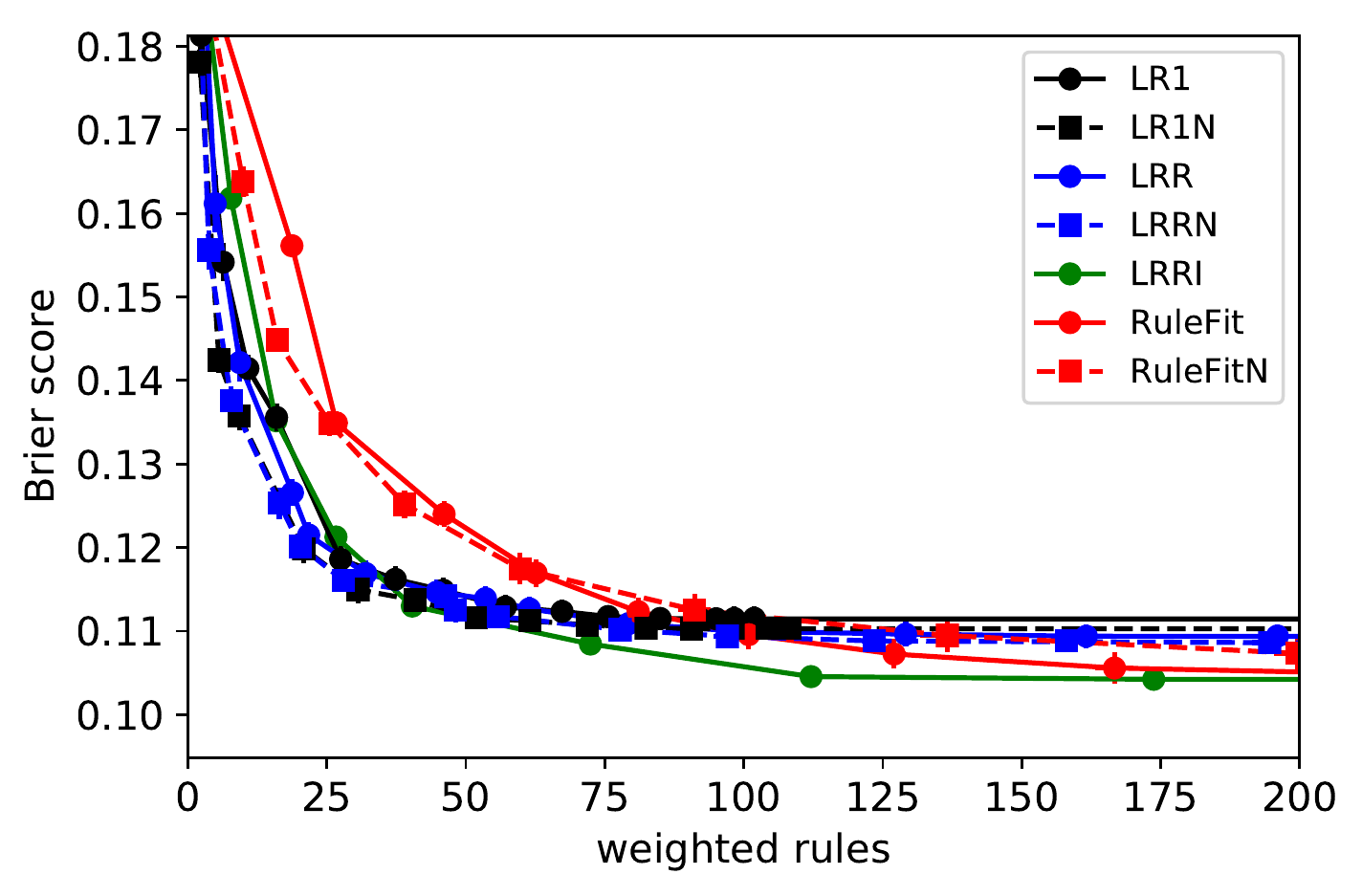}
  \caption{MAGIC gamma telescope}
  \label{fig:paretoClassBrier:magic0}
  \end{subfigure}
  \begin{subfigure}[b]{0.7\columnwidth}
  \includegraphics[width=\columnwidth]{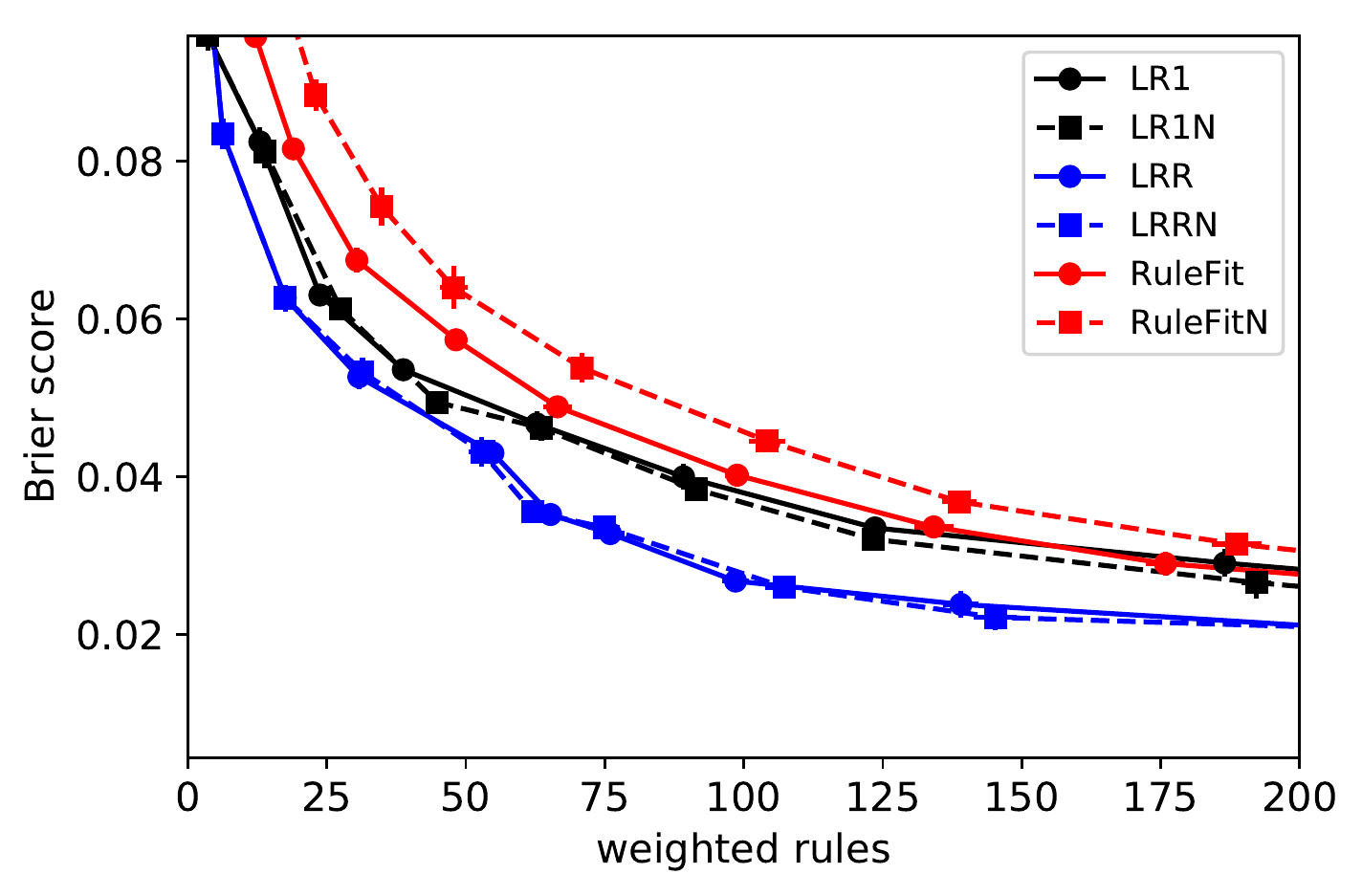}
  \caption{Musk molecules}
  \label{fig:paretoClassBrier:musk0}
  \end{subfigure}
  \caption{Trade-offs between Brier score and weighted number of rules. Pareto efficient points are connected by line segments. Horizontal and vertical bars represent standard errors in the means.} 
  \label{fig:paretoClassBrier}
\end{figure*}

\begin{table*}[ht]
\caption{Mean test Brier scores (standard error in parentheses). Best values in \textbf{bold}.}
\label{tbl:Brier}
\begin{center}
\begin{tiny}
\begin{tabular}{lrrrrrrrr}
\toprule
dataset&\multicolumn{1}{c}{LR1}&\multicolumn{1}{c}{LRR}&\multicolumn{1}{c}{RuleFit}&\multicolumn{1}{c}{LR1N}&\multicolumn{1}{c}{LRRN}&\multicolumn{1}{c}{RuleFitN}&\multicolumn{1}{c}{GBT}&\multicolumn{1}{c}{SVM}\\
\midrule
banknote&$2.58$ ($0.75$) E-3&$2.58$ ($0.98$) E-3&$3.46$ ($0.99$) E-3&$\mathbf{7.43}$ ($1.94$) \textbf{E-5}&$1.54$ ($0.95$) E-3&$1.53$ ($0.83$) E-3&$4.50$ ($1.04$) E-3&$3.58$ ($2.47$) E-4\\
heart&$1.33$ ($0.09$) E-1&$1.33$ ($0.10$) E-1&$\mathbf{1.22}$ ($0.09$) \textbf{E-1}&$1.34$ ($0.10$) E-1&$1.32$ ($0.09$) E-1&$1.25$ ($0.10$) E-1&$1.33$ ($0.08$) E-1&$1.26$ ($0.12$) E-1\\
ILPD&$1.86$ ($0.06$) E-1&$1.82$ ($0.05$) E-1&$2.50$ ($0.00$) E-1&$1.86$ ($0.06$) E-1&$1.82$ ($0.05$) E-1&$2.50$ ($0.00$) E-1&$\mathbf{1.78}$ ($0.03$) \textbf{E-1}&$1.94$ ($0.02$) E-1\\
ionosphere&$7.26$ ($0.97$) E-2&$6.63$ ($0.95$) E-2&$6.31$ ($1.39$) E-2&$7.07$ ($1.11$) E-2&$6.99$ ($1.09$) E-2&$4.95$ ($1.30$) E-2&$6.98$ ($1.08$) E-2&$\mathbf{3.96}$ ($1.18$) \textbf{E-2}\\
liver&$2.50$ ($0.12$) E-1&$2.47$ ($0.09$) E-1&$2.55$ ($0.11$) E-1&$2.57$ ($0.13$) E-1&$2.50$ ($0.11$) E-1&$2.48$ ($0.16$) E-1&$2.47$ ($0.06$) E-1&$\mathbf{2.35}$ ($0.08$) \textbf{E-1}\\
pima&$1.63$ ($0.09$) E-1&$1.66$ ($0.09$) E-1&$1.62$ ($0.09$) E-1&$1.61$ ($0.08$) E-1&$1.65$ ($0.08$) E-1&$1.66$ ($0.11$) E-1&$1.66$ ($0.07$) E-1&$\mathbf{1.57}$ ($0.10$) \textbf{E-1}\\
tic-tac-toe&$1.62$ ($0.30$) E-2&$1.71$ ($0.41$) E-2&$\mathbf{3.78}$ ($2.58$) \textbf{E-7}&$1.62$ ($0.30$) E-2&$1.71$ ($0.41$) E-2&$\mathbf{3.78}$ ($2.58$) \textbf{E-7}&$8.22$ ($1.60$) E-3&$1.53$ ($0.35$) E-2\\
transfusion&$1.66$ ($0.05$) E-1&$1.55$ ($0.04$) E-1&$1.62$ ($0.09$) E-1&$\mathbf{1.53}$ ($0.03$) \textbf{E-1}&$1.55$ ($0.03$) E-1&$1.66$ ($0.11$) E-1&$1.60$ ($0.03$) E-1&$1.68$ ($0.03$) E-1\\
WDBC&$2.52$ ($0.61$) E-2&$1.76$ ($0.38$) E-2&$2.01$ ($0.35$) E-2&$2.08$ ($0.61$) E-2&$\mathbf{1.41}$ ($0.34$) \textbf{E-2}&$2.62$ ($0.44$) E-2&$3.09$ ($0.37$) E-2&$1.63$ ($0.26$) E-2\\
\midrule
adult&$1.05$ ($0.01$) E-1&$1.05$ ($0.01$) E-1&$1.07$ ($0.01$) E-1&$9.78$ ($0.09$) E-2&$9.78$ ($0.09$) E-2&$\mathbf{9.01}$ ($0.09$) \textbf{E-2}&$1.04$ ($0.01$) E-1&$1.11$ ($0.01$) E-1\\
bank-mkt&$1.00$ ($0.00$) E-1&$7.79$ ($0.07$) E-2&$1.73$ ($0.00$) E-1&$1.00$ ($0.00$) E-1&$\mathbf{7.77}$ ($0.07$) \textbf{E-2}&$9.99$ ($0.00$) E-2&$7.98$ ($0.07$) E-2&$8.94$ ($0.06$) E-2\\
gas&$3.93$ ($0.41$) E-3&$\mathbf{3.40}$ ($0.35$) \textbf{E-3}&$4.19$ ($0.49$) E-3&$3.99$ ($0.33$) E-3&$4.40$ ($0.34$) E-3&$3.70$ ($0.30$) E-3&$5.72$ ($0.50$) E-3&$4.11$ ($0.46$) E-3\\
magic&$1.11$ ($0.01$) E-1&$1.09$ ($0.01$) E-1&$1.06$ ($0.02$) E-1&$1.10$ ($0.01$) E-1&$1.09$ ($0.02$) E-1&$\mathbf{9.33}$ ($0.17$) \textbf{E-2}&$9.56$ ($0.17$) E-2&$9.34$ ($0.15$) E-2\\
mushroom&$5.05$ ($1.75$) E-7&$2.02$ ($1.10$) E-7&$\mathbf{0.00}$ ($0.00$) \textbf{E-7}&$5.05$ ($1.75$) E-7&$2.02$ ($1.10$) E-7&$\mathbf{0.00}$ ($0.00$) \textbf{E-7}&$4.69$ ($2.51$) E-4&$3.10$ ($1.73$) E-4\\
musk&$2.53$ ($0.31$) E-2&$1.37$ ($0.09$) E-2&$2.02$ ($0.11$) E-2&$3.00$ ($0.49$) E-2&$\mathbf{1.20}$ ($0.10$) \textbf{E-2}&$1.76$ ($0.12$) E-2&$4.27$ ($0.33$) E-2&$1.30$ ($0.38$) E-2\\
FICO&$1.79$ ($0.01$) E-1&$1.79$ ($0.02$) E-1&$1.80$ ($0.02$) E-1&$1.78$ ($0.02$) E-1&$\mathbf{1.78}$ ($0.01$) \textbf{E-1}&$1.79$ ($0.01$) E-1&$1.80$ ($0.01$) E-1&$1.88$ ($0.01$) E-1\\
\midrule
mean rank&$5.81$&$4.12$&$4.84$&$4.75$&$\mathbf{3.56}$&$3.59$&$5.31$&$4.00$\\
\bottomrule
\end{tabular}
\end{tiny}
\end{center}
\vskip -0.1in
\end{table*}

\begin{table*}[ht]
\caption{Mean weighted number of rules (standard error in parentheses) corresponding to Table~\ref{tbl:Brier}. Best values in \textbf{bold}.}
\label{tbl:BrierWeighted}
\begin{center}
\begin{scriptsize}
\begin{tabular}{lrrrrrr}
\toprule
dataset&\multicolumn{1}{c}{LR1}&\multicolumn{1}{c}{LRR}&\multicolumn{1}{c}{RuleFit}&\multicolumn{1}{c}{LR1N}&\multicolumn{1}{c}{LRRN}&\multicolumn{1}{c}{RuleFitN}\\
\midrule
banknote&$32.3$ ($0.8$)&$54.9$ ($2.4$)&$57.8$ ($0.7$)&$\mathbf{16.4}$ ($0.4$)&$50.6$ ($3.3$)&$1124.9$ ($67.7$)\\
heart&$13.4$ ($2.7$)&$7.6$ ($0.7$)&$34.3$ ($0.9$)&$14.3$ ($2.1$)&$\mathbf{6.0}$ ($0.8$)&$59.4$ ($2.4$)\\
ILPD&$14.6$ ($3.7$)&$11.6$ ($1.4$)&$\mathbf{0.0}$ ($0.0$)&$14.7$ ($4.9$)&$11.3$ ($1.3$)&$\mathbf{0.0}$ ($0.0$)\\
ionosphere&$\mathbf{114.4}$ ($28.5$)&$122.8$ ($35.4$)&$1007.3$ ($12.0$)&$130.3$ ($23.7$)&$122.6$ ($44.1$)&$983.4$ ($145.2$)\\
liver&$28.9$ ($5.8$)&$16.9$ ($2.6$)&$66.7$ ($11.7$)&$25.7$ ($5.7$)&$\mathbf{14.0}$ ($2.0$)&$89.7$ ($36.8$)\\
pima&$22.1$ ($2.3$)&$25.9$ ($1.4$)&$64.8$ ($1.1$)&$\mathbf{12.1}$ ($1.0$)&$18.7$ ($2.9$)&$183.8$ ($34.9$)\\
tic-tac-toe&$\mathbf{21.6}$ ($0.0$)&$78.2$ ($6.3$)&$1640.7$ ($99.2$)&$\mathbf{21.6}$ ($0.0$)&$78.2$ ($6.3$)&$1640.7$ ($99.2$)\\
transfusion&$15.2$ ($4.3$)&$17.6$ ($1.1$)&$64.8$ ($1.1$)&$24.3$ ($2.5$)&$\mathbf{12.4}$ ($1.9$)&$183.8$ ($34.9$)\\
WDBC&$145.7$ ($18.0$)&$271.4$ ($13.6$)&$809.4$ ($89.6$)&$\mathbf{86.1}$ ($12.6$)&$248.4$ ($28.0$)&$562.3$ ($83.3$)\\
\midrule
adult&$87.1$ ($1.6$)&$\mathbf{83.2}$ ($5.3$)&$102.4$ ($4.6$)&$85.8$ ($2.4$)&$104.7$ ($4.3$)&$719.9$ ($58.6$)\\
bank-mkt&$\mathbf{0.0}$ ($0.0$)&$68.9$ ($3.6$)&$\mathbf{0.0}$ ($0.0$)&$\mathbf{0.0}$ ($0.0$)&$61.8$ ($3.6$)&$0.2$ ($0.0$)\\
gas&$\mathbf{483.7}$ ($8.0$)&$694.8$ ($2.1$)&$2663.1$ ($235.9$)&$950.2$ ($12.9$)&$1145.2$ ($25.6$)&$2920.8$ ($125.7$)\\
magic&$\mathbf{93.1}$ ($2.2$)&$202.1$ ($13.5$)&$496.7$ ($5.9$)&$97.9$ ($2.9$)&$219.9$ ($21.5$)&$947.4$ ($7.0$)\\
mushroom&$\mathbf{24.7}$ ($0.6$)&$30.1$ ($1.6$)&$1308.9$ ($207.5$)&$\mathbf{24.7}$ ($0.6$)&$30.1$ ($1.6$)&$1308.9$ ($207.5$)\\
musk&$\mathbf{263.0}$ ($39.3$)&$1255.7$ ($25.4$)&$1152.1$ ($298.9$)&$313.9$ ($101.6$)&$1348.3$ ($50.0$)&$2000.3$ ($314.8$)\\
FICO&$92.6$ ($5.9$)&$72.9$ ($5.1$)&$239.8$ ($2.5$)&$81.0$ ($5.2$)&$\mathbf{51.8}$ ($2.7$)&$183.4$ ($3.0$)\\
\midrule
mean rank&$\mathbf{2.31}$&$3.19$&$4.66$&$2.56$&$2.94$&$5.34$\\
\bottomrule
\end{tabular}
\end{scriptsize}
\end{center}
\vskip -0.1in
\end{table*}

\subsection{Classification}
\label{sec:eval:class}

For classification, we used the same $16$ datasets considered in \cite{dash2018}, which  also appeared in other recent works on rule-based models \cite{su2016,wang2017}.  One of these datasets comes from the recent FICO Explainable Machine Learning Challenge \cite{FICO2018}. 

In the first experiment, we evaluated the performance-complexity trade-offs of the four proposed methods (LR1, LRR, LR1N, LRRN) as well as RuleFit.  For performance metrics, we report both accuracy and Brier score (i.e.,~mean squared error (MSE)), the latter a well-known metric for probabilistic outputs \cite{hernandez-orallo2012} as produced by logistic regression-based models.  To measure rule ensemble complexity, we consider not only the number of rules (with nonzero coefficients) but also their lengths in terms of number of conditions.  Specifically we define the \emph{weight} of a rule similarly to the regularization parameter $\lambda_k$ as $1 + w \sum_j z_j$, where we take $w = 0.2$ as the weight on the degree.  Coefficients corresponding to numerical features receive a weight of $1$.  For consistency with this definition, the parameters $\lambda_k$ used by all methods in this comparison are set proportional to the weights, i.e.,~with $\lambda_1 / \lambda_0 = 0.2$.  By varying the remaining free parameter $\lambda_0$, we sweep out trade-offs between performance and complexity.  RuleFit has an additional parameter, the mean tree size, that is recommended for tuning in \cite{friedman2008}.  We have done so based on test set results, which gives RuleFit a slight advantage.

\begin{figure*}[t]
  \centering
  \begin{subfigure}[b]{0.685\columnwidth}
  \includegraphics[width=\columnwidth]{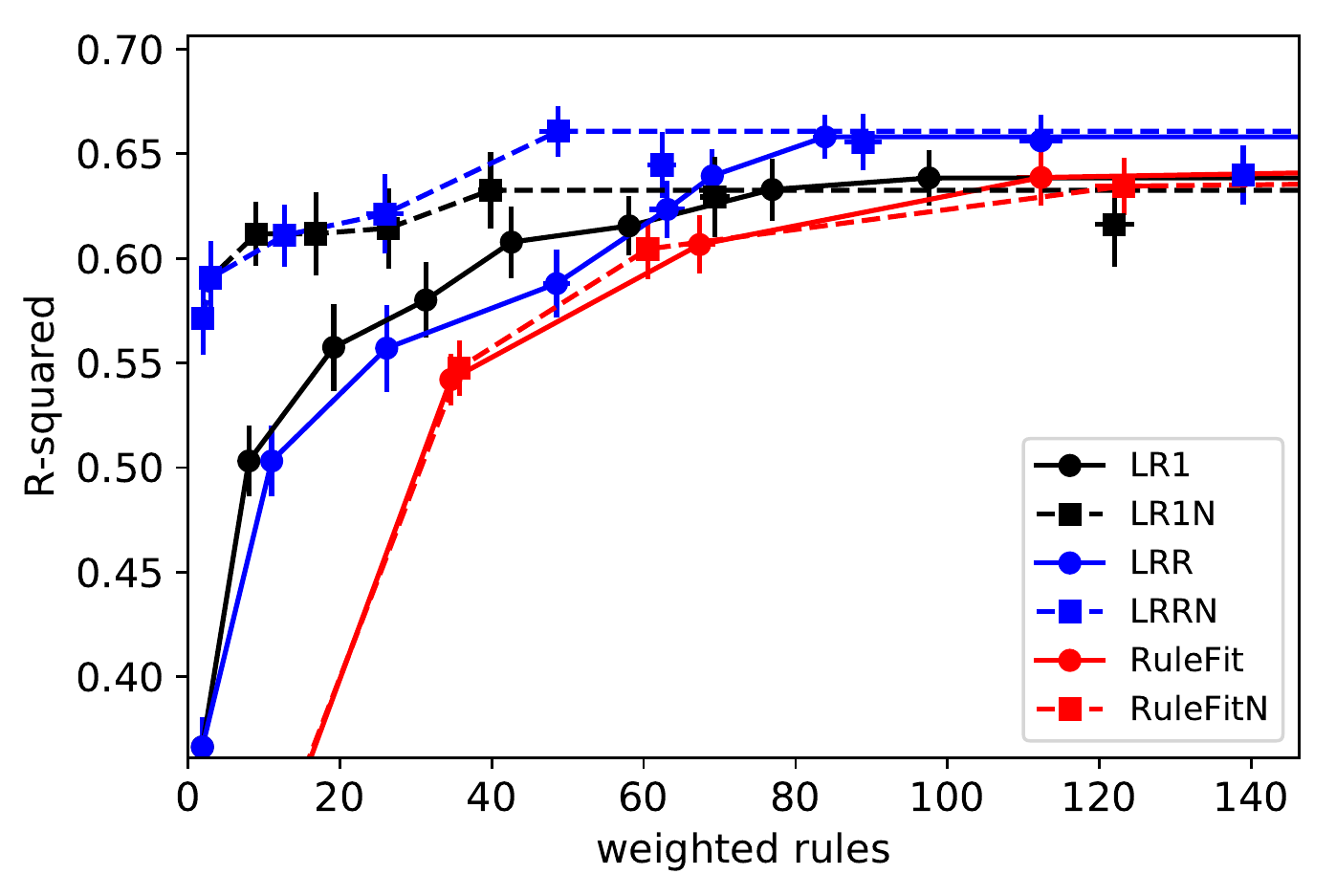}
  \caption{Communities and Crime}
  \label{fig:paretoRegress:crime0}
  \end{subfigure}
  \begin{subfigure}[b]{0.685\columnwidth}
  \includegraphics[width=\columnwidth]{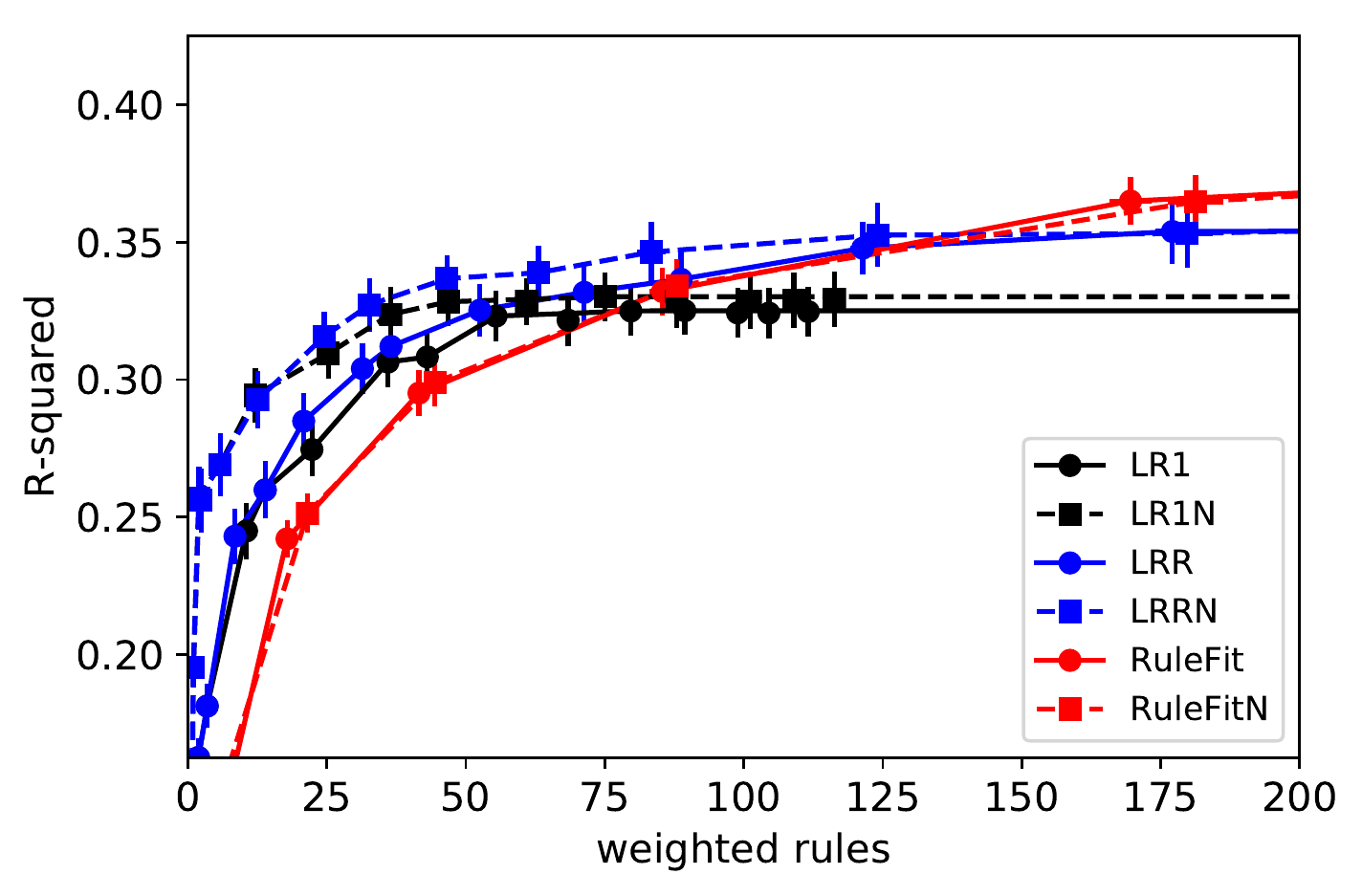}
  \caption{Wine Quality}
  \label{fig:paretoRegress:wine0}
  \end{subfigure}
  \begin{subfigure}[b]{0.685\columnwidth}
  \includegraphics[width=\columnwidth]{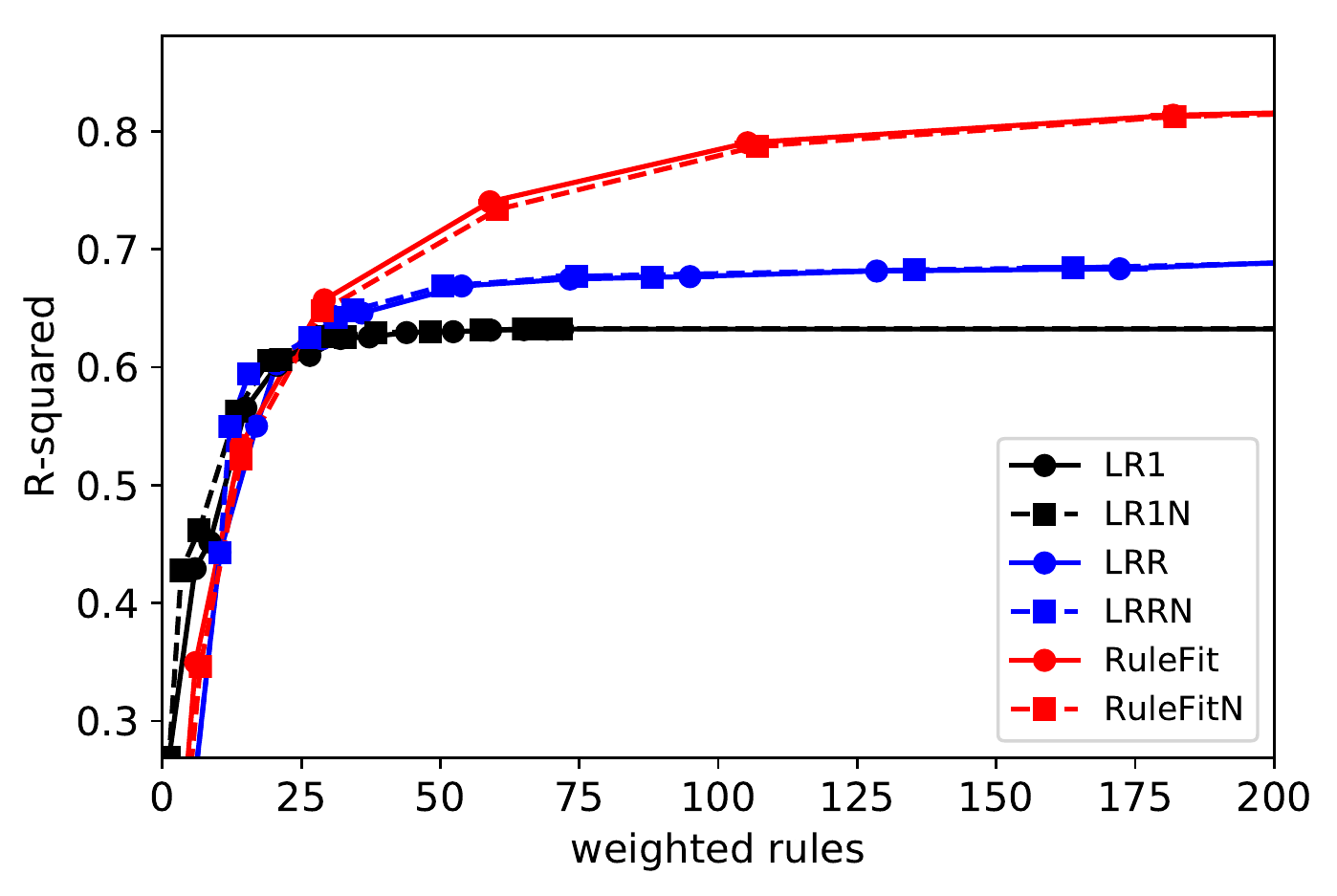}
  \caption{Bike Sharing}
  \label{fig:paretoRegress:bike0}
  \end{subfigure}
  \caption{Trade-offs between coefficient of determination $R^2$ and weighted number of rules. 
  } 
  \label{fig:paretoRegress}
\end{figure*}

Figure~\ref{fig:paretoClassBrier} shows the resulting trade-offs with Brier score for $4$ of the $16$ datasets.  The other $12$ plots as well as those for accuracy are in Appendix~\ref{sec:eval:add}. Pareto-efficient points, i.e.,~those not dominated by points with both lower Brier score and lower complexity, have been connected with line segments for the sole purpose of visualization.  RuleFit obtains inferior trade-offs on most of the tested datasets.  Even in cases such as Figure~\ref{fig:paretoClassBrier:magic0} where RuleFit eventually attains a lower Brier score, the initial part of the trade-off is worse.  Note  that the variants employing numerical features generally fare better, as expected.  Figure~\ref{fig:paretoClassBrier:magic0} indicates that IP CG (LRRI) can improve upon the heuristic in some cases.

Next we discuss the differences between LR1(N) and LRR(N).  A general observation is that LRR, which uses CG to produce higher-degree rules, tends to achieve better trade-offs on larger datasets, as exemplified by musk in Figure~\ref{fig:paretoClassBrier:musk0}.  This can be explained by the fact that LRR effectively considers a much larger feature space than LR1.  If the training sample size is sufficient to support this, then the greater power of the higher-degree rules found by LRR generalizes to test data. The presence of strong interactions in the data also favors LRR.  On the other hand, on smaller datasets such as in Figure~\ref{fig:paretoClassBrier:pima0}, LRR may overfit and achieve a worse trade-off relative to LR1.  In Figure~\ref{fig:paretoClassBrier:FICO0}, LRR outperforms LR1 but the advantage disappears for LRRN compared to LR1N, a pattern that also occurs on other datasets.  In Figure~\ref{fig:paretoClassBrier:magic0}, the advantage of LRR(N) lies in attaining a slightly lower minimal Brier score.

In a second experiment, we aim to maximize performance (minimize Brier score or maximize accuracy) by performing nested CV on the training set to select $\lambda_0$ and applying the resulting model to the test set.  Since performance is now the primary criterion, we broaden the comparison to include GBT and SVM.  For GBT, the maximum tree depth was tuned and the number of trees was also determined via a stopping criterion on a validation set, up to a maximum of $500$ trees.  For SVM, the regularization parameter $C$ and kernel width $\gamma$ were tuned and Platt scaling \cite{platt1999} was used to calibrate the output scores as probabilities.

Table~\ref{tbl:Brier} shows the resulting mean Brier scores while Table~\ref{tbl:BrierWeighted} shows the weighted number of rules at which the Brier scores were achieved. To help summarize these results, we report the mean rank of each method as well as Friedman tests on these mean ranks, as suggested by a reviewer and following \cite{demsar2006}. The overall conclusion is that when tuned for maximum performance, the proposed methods, especially LRRN, compete well with benchmark models and do so using significantly fewer rules than RuleFit. 

For Table~\ref{tbl:Brier}, the Friedman statistic is $12.41$, corresponding to a p-value of $0.082$ using the $F$-distribution approximation. Hence the null hypothesis of no significant differences is rejected at the $0.10$ level but not at $0.05$. A post-hoc test, comparing all other algorithms to LRRN and correcting for multiple comparisons using Holm's procedure, shows that the only significant difference at the $0.10$ level is with LR1, i.e.,~when excluding both higher-degree rules and numerical features. The mean ranks also show that LRR(N) outperforms LR1(N) although most of the differences are not statistically significant.

In Table~\ref{tbl:BrierWeighted}, it is clear that RuleFit produces much more (3-4 times) complex classifiers than all four proposed methods. The Friedman statistic of $34.01$ (p-value $\sim 10^{-6}$) is significant as expected. Post-hoc comparisons to LRRN confirm that RuleFit and RuleFitN are significantly more complex, this time at the $0.05$ level and again with Holm's correction. Note that including original numerical features does not significant increase classifier complexity (comparing LR1 with LR1N and LRR with LRRN). 



\subsection{Regression}
\label{sec:eval:regress}

\begin{table*}[ht]
\caption{Mean test $R^2$ (\%, standard error in parentheses). Best values in \textbf{bold}.}
\label{tbl:R2}
\begin{center}
\begin{scriptsize}
\begin{tabular}{lrrrrrrrr}
\toprule
dataset&\multicolumn{1}{c}{LR1}&\multicolumn{1}{c}{LRR}&\multicolumn{1}{c}{RuleFit}&\multicolumn{1}{c}{LR1N}&\multicolumn{1}{c}{LRRN}&\multicolumn{1}{c}{RuleFitN}&\multicolumn{1}{c}{GBT}&\multicolumn{1}{c}{SVM}\\
\midrule
abalone&$51.2$ ($1.2$)&$51.2$ ($1.3$)&$51.1$ ($1.1$)&$55.4$ ($1.3$)&$55.5$ ($1.3$)&$54.7$ ($1.3$)&$51.8$ ($1.0$)&$\mathbf{56.7}$ ($1.0$)\\
boston&$77.9$ ($4.1$)&$76.4$ ($4.2$)&$84.3$ ($2.3$)&$78.8$ ($3.5$)&$78.2$ ($3.7$)&$\mathbf{84.3}$ ($2.2$)&$79.3$ ($3.5$)&$82.0$ ($2.7$)\\
bike&$63.2$ ($0.5$)&$69.1$ ($0.5$)&$83.0$ ($0.4$)&$63.3$ ($0.5$)&$68.9$ ($0.5$)&$83.0$ ($0.4$)&$\mathbf{83.9}$ ($0.3$)&$54.3$ ($0.5$)\\
california&$70.2$ ($0.4$)&$73.3$ ($0.4$)&$75.2$ ($0.3$)&$72.7$ ($0.3$)&$75.6$ ($0.3$)&$\mathbf{76.6}$ ($0.3$)&$75.6$ ($0.4$)&$76.4$ ($0.4$)\\
crime&$60.6$ ($1.7$)&$63.9$ ($1.3$)&$64.1$ ($1.5$)&$61.3$ ($2.2$)&$\mathbf{65.7}$ ($1.2$)&$63.5$ ($1.6$)&$63.9$ ($1.2$)&$54.5$ ($1.0$)\\
parkinsons&$17.8$ ($0.6$)&$45.2$ ($1.3$)&$25.3$ ($3.4$)&$17.8$ ($0.6$)&$46.1$ ($1.2$)&$\mathbf{49.2}$ ($7.9$)&$29.3$ ($0.8$)&$7.0$ ($0.3$)\\
wine&$32.3$ ($1.0$)&$35.5$ ($1.2$)&$33.6$ ($0.9$)&$32.9$ ($0.9$)&$35.6$ ($1.4$)&$32.6$ ($1.1$)&$\mathbf{38.0}$ ($1.1$)&$36.5$ ($0.8$)\\
MEPS&$16.4$ ($1.5$)&$16.4$ ($1.4$)&$15.3$ ($1.6$)&$\mathbf{16.6}$ ($1.5$)&$16.5$ ($1.4$)&$14.5$ ($1.4$)&$15.7$ ($1.5$)&$5.5$ ($0.6$)\\
\midrule
mean rank&$6.75$&$4.88$&$4.50$&$5.00$&$\mathbf{3.00}$&$3.50$&$3.38$&$5.00$\\
\bottomrule
\end{tabular}
\end{scriptsize}
\end{center}
\vskip -0.1in
\end{table*}

\begin{table*}[h!]
\caption{Mean weighted number of rules (standard error in parentheses) corresponding to Table~\ref{tbl:R2}. Best values in \textbf{bold}.}
\label{tbl:R2Weighted}
\begin{center}
\begin{scriptsize}
\begin{tabular}{lrrrrrr}
\toprule
dataset&\multicolumn{1}{c}{LR1}&\multicolumn{1}{c}{LRR}&\multicolumn{1}{c}{RuleFit}&\multicolumn{1}{c}{LR1N}&\multicolumn{1}{c}{LRRN}&\multicolumn{1}{c}{RuleFitN}\\
\midrule
abalone&$\mathbf{69.4}$ ($1.0$)&$141.7$ ($13.0$)&$90.1$ ($3.5$)&$74.9$ ($0.9$)&$144.1$ ($13.8$)&$83.4$ ($2.1$)\\
boston&$56.3$ ($3.0$)&$96.1$ ($21.3$)&$432.3$ ($41.1$)&$\mathbf{50.2}$ ($3.1$)&$97.1$ ($24.9$)&$285.8$ ($18.2$)\\
bike&$\mathbf{69.7}$ ($1.0$)&$287.1$ ($14.7$)&$558.3$ ($29.6$)&$70.7$ ($0.9$)&$285.0$ ($16.0$)&$568.9$ ($28.0$)\\
california&$82.1$ ($0.6$)&$219.8$ ($2.9$)&$315.3$ ($11.3$)&$\mathbf{78.7}$ ($0.7$)&$204.4$ ($3.8$)&$447.2$ ($15.9$)\\
crime&$237.4$ ($2.8$)&$76.8$ ($5.8$)&$161.7$ ($3.7$)&$130.9$ ($16.5$)&$\mathbf{55.1}$ ($5.9$)&$168.5$ ($5.0$)\\
parkinsons&$\mathbf{2.4}$ ($0.0$)&$136.5$ ($5.9$)&$14.0$ ($2.0$)&$\mathbf{2.4}$ ($0.0$)&$134.1$ ($4.5$)&$389.8$ ($120.2$)\\
wine&$75.7$ ($2.2$)&$188.0$ ($10.9$)&$112.2$ ($7.9$)&$\mathbf{61.8}$ ($1.7$)&$189.8$ ($9.5$)&$108.2$ ($11.6$)\\
MEPS&$99.4$ ($2.2$)&$80.6$ ($7.4$)&$115.1$ ($3.3$)&$104.9$ ($2.7$)&$\mathbf{78.9}$ ($7.6$)&$107.2$ ($3.1$)\\
\midrule
mean rank&$2.31$&$3.75$&$4.62$&$\mathbf{1.94}$&$3.50$&$4.88$\\
\bottomrule
\end{tabular}
\end{scriptsize}
\end{center}
\vskip -0.1in
\end{table*}

For regression, we experimented with an additional $8$ datasets, $7$ of which are drawn from previous works on rule ensembles \cite{friedman2008,dembczynski2010} and the UCI repository \cite{dua2017}.  The last dataset comes from the Medical Expenditure Panel Survey (MEPS) \cite{MEPS} of the US Department of Health and Human Services, specifically panel 19 from the year 2015.  The task is to predict the annual healthcare expenditure of individuals based on demographics and self-reported medical conditions.

The same two experiments are conducted for regression, using the linear regression variants \eqref{eqn:linR} of both the proposed approaches as well as RuleFit. The coefficient of determination $R^2$ is chosen as the performance metric and the model complexity metric is the same as in Section~\ref{sec:eval:class}.  

In Figure~\ref{fig:paretoRegress}, we show trade-offs between $R^2$ and weighted number of rules for $3$ of the datasets; the other $5$ can be found in Appendix~\ref{sec:eval:add}. RuleFit is a stronger competitor in regression due to sometimes achieving higher $R^2$ at higher complexities. On $4$ of $8$ datasets, the curves for RuleFit(N) remain below their LRR(N) counterparts as in Figure~\ref{fig:paretoRegress:crime0}. On another $2$ datasets, the curves cross at moderate to high complexities as in Figure~\ref{fig:paretoRegress:wine0}, while in Figure~\ref{fig:paretoRegress:bike0}, RuleFit obtains much higher $R^2$. Among the proposed methods, the advantage of LRR(N) vs.~LR1(N) is generally larger in regression than in classification (cf.~Figure~\ref{fig:paretoClassBrier}), indicating the benefit of generating higher-degree rules. A possible explanation may be that interaction terms matter more in regression where the output range is wider.

Tables~\ref{tbl:R2} and \ref{tbl:R2Weighted} show the results of selecting parameter $\lambda_0$ through nested CV to maximize $R^2$. The same overall conclusion as in Section~\ref{sec:eval:class} holds for LRRN in particular, namely that it yields highly competitive $R^2$ values while using fewer rules than RuleFit. The Friedman statistic for Table~\ref{tbl:R2} is $13.63$ (p-value $0.046$), and in post-hoc comparisons to LRRN, the only significant difference ($0.05$ level) is again with LR1. Among the proposed methods, advantages due to CG and/or numerical features in Figure~\ref{fig:paretoRegress} carry over into Table~\ref{tbl:R2}.  
In Table~\ref{tbl:R2Weighted}, RuleFit(N) again yields more complex solutions on average, although the difference is not as large as in Table~\ref{tbl:BrierWeighted}. The Friedman statistic is $16.16$ (p-value $0.002$); however, no post-hoc comparisons with LRRN (which occupies a middle position) as the reference show statistically significant differences.

\section{Discussion}
\label{sec:concl}

The numerical results in Section~\ref{sec:eval} may raise the question of the interpretability of rule ensembles with hundreds of rules. First we note that because of the shape of many of the trade-off curves, with steep improvement at low complexity followed by a flatter region, it may be possible to obtain substantially simpler models, with say a few tens of rules, that are not too far from maximum performance.  Apart from model simplification, the fact that a rule ensemble is also a linear model facilitates model inspection, for example by focusing on the rules corresponding to the largest coefficients $\beta_k$. \citet{friedman2008} discuss a slightly more refined measure of rule importance, which is also used to assess the importance of (original) input features. Based on the discussion in Section~\ref{sec:instants}, we also suggest a division between singleton rules and linear terms on the one hand, and higher-degree rules on the other.  Since the former constitute a GAM, their effect can be summarized visually by univariate plots.  The higher-degree rules can be ranked and the interactions studied further as described in \cite{friedman2008}. 

The following extensions are suggested for future work: 1) We have used a fixed binarization of the features to facilitate formulation of the column generation subproblem \eqref{eqn:pricing2} as an IP. However, if CG is done using a heuristic, it may be possible to refine the binarization with each CG iteration. Doing so may improve results, particularly for regression.
2) In the experiments in Section~\ref{sec:eval}, we have fixed the ratio $\lambda_1 / \lambda_0$ to match the definition of rule weight. One may also vary this ratio to encourage or discourage longer rules, or tune it to the level of interaction present in a dataset.

\section*{Acknowledgements}
We thank Karthikeyan Natesan Ramamurthy for help with the MEPS dataset.

\bibliography{main}
\bibliographystyle{icml2019}

\clearpage
\appendix
\input{appendix.tex}

\end{document}

%% file: appendix.tex
\section{Variations on the Column Generation Heuristic}
\label{sec:heuristicVars}

We have explored the following three variations of the heuristic algorithm for column generation described in Section~\ref{sec:CG}: 
\begin{enumerate}
    \item The algorithm can return the best $K$ solutions that it finds instead of a single incumbent solution, potentially reducing the number of CG iterations needed.  We have observed however that these solutions tend to correspond to very similar conjunctions and are hence highly correlated.  By allowing multiple such columns to enter together, sparsity suffers because the $\ell_1$ regularization in \eqref{eqn:regLL} has difficulty favoring sparse linear combinations of highly correlated columns over dense ones.  For this reason we  kept $K = 1$.
    \item The algorithm can be generalized to a beam search by considering the children of $B > 1$ parent conjunctions at each degree instead of a single parent.  The best $B$ children according to a combination of metrics \eqref{eqn:objChange} and \eqref{eqn:lowerBound} are then chosen to become the next parents.  To date however, we have not found setting $B > 1$ to be beneficial.
    \item The algorithm can be terminated early once a solution with negative objective value is found since any such solution corresponds to a descent direction for problem \eqref{eqn:regLLpm}.  Termination can be immediate or occur after the current degree.  While early termination speeds up each CG iteration, the number of iterations tends to increase because the generated columns are of lower quality. 
\end{enumerate}

\section{Additional Numerical Results}
\label{sec:eval:add}

\subsection{Classification}

\begin{figure*}[t]
  \centering
  \begin{subfigure}[b]{0.9\columnwidth}
  \includegraphics[width=0.9\columnwidth]{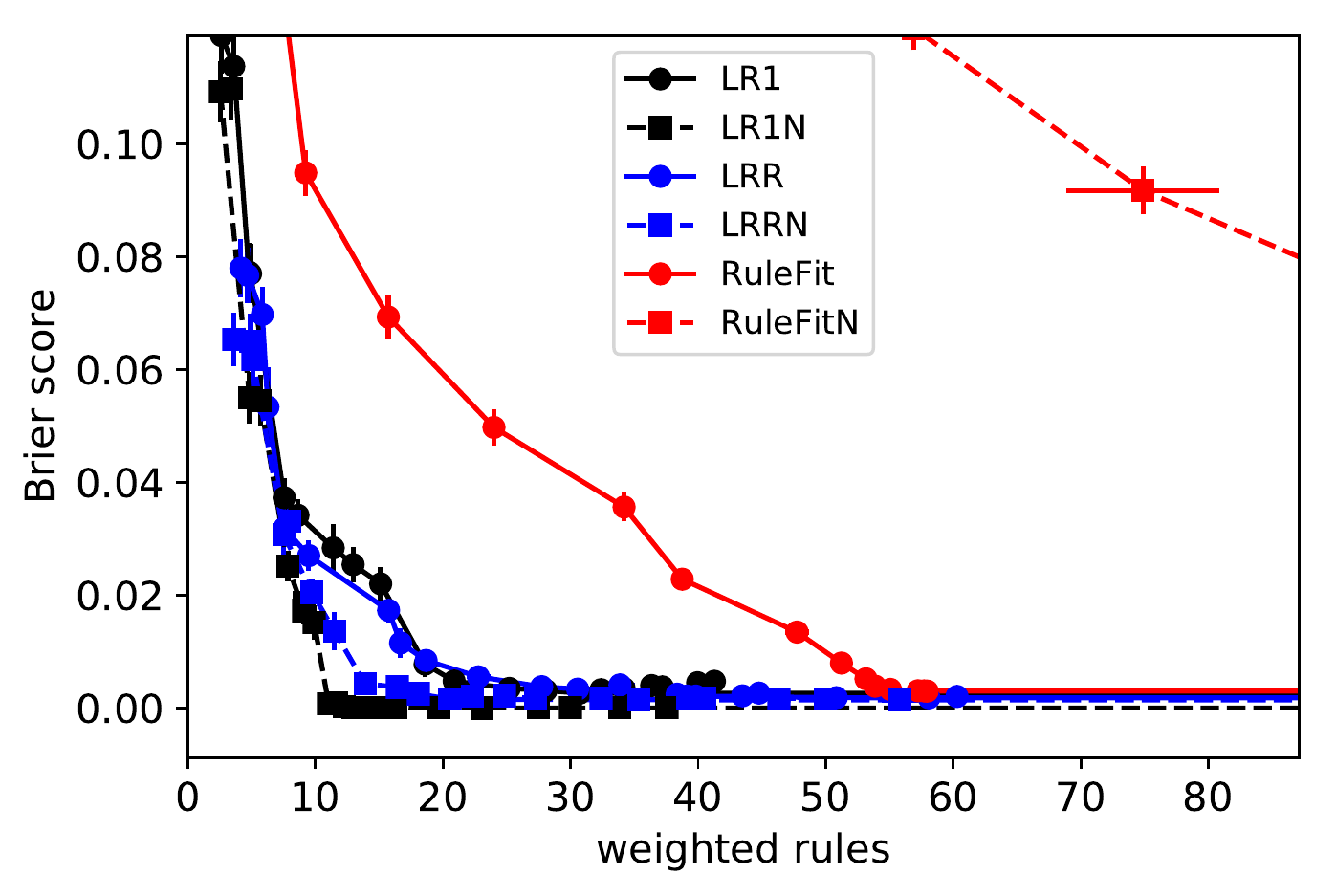}
  \caption{banknote}
  \label{fig:paretoClassBrier:banknote}
  \end{subfigure}
  \begin{subfigure}[b]{0.9\columnwidth}
  \includegraphics[width=0.9\columnwidth]{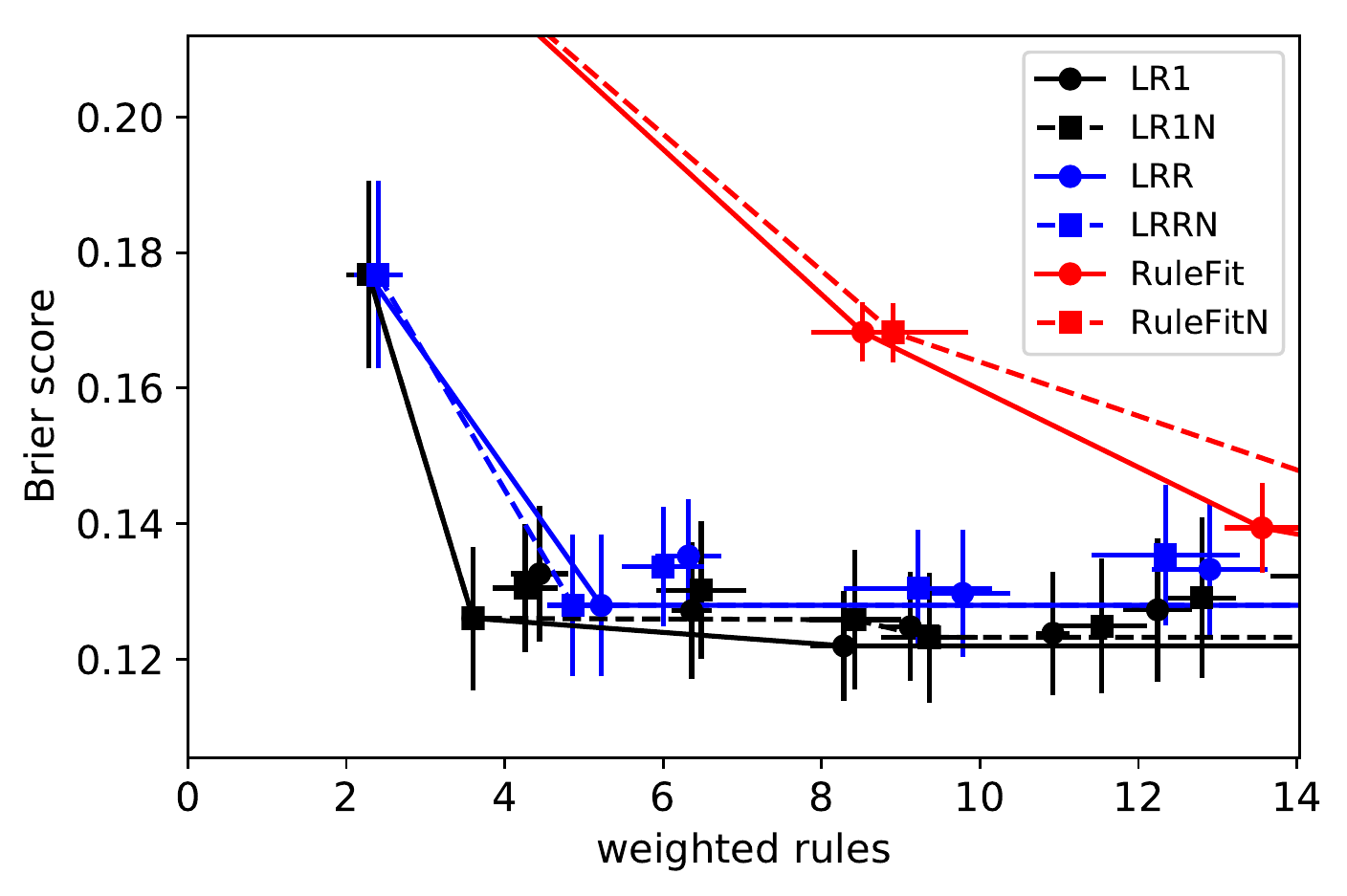}
  \caption{heart}
  \label{fig:paretoClassBrier:heart}
  \end{subfigure}
  \begin{subfigure}[b]{0.9\columnwidth}
  \includegraphics[width=0.9\columnwidth]{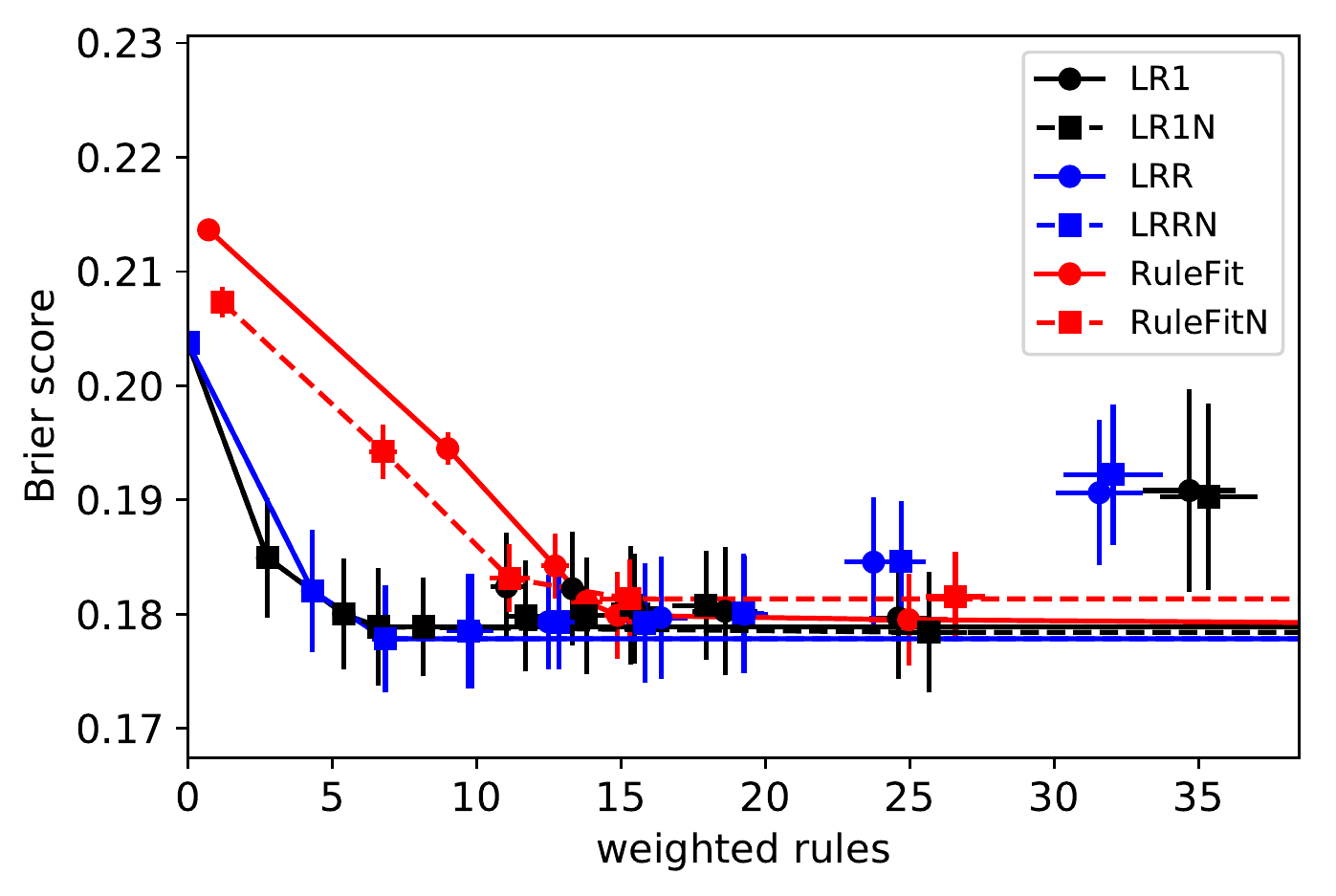}
  \caption{ILPD}
  \label{fig:paretoClassBrier:ILPD}
  \end{subfigure}
  \begin{subfigure}[b]{0.9\columnwidth}
  \includegraphics[width=0.9\columnwidth]{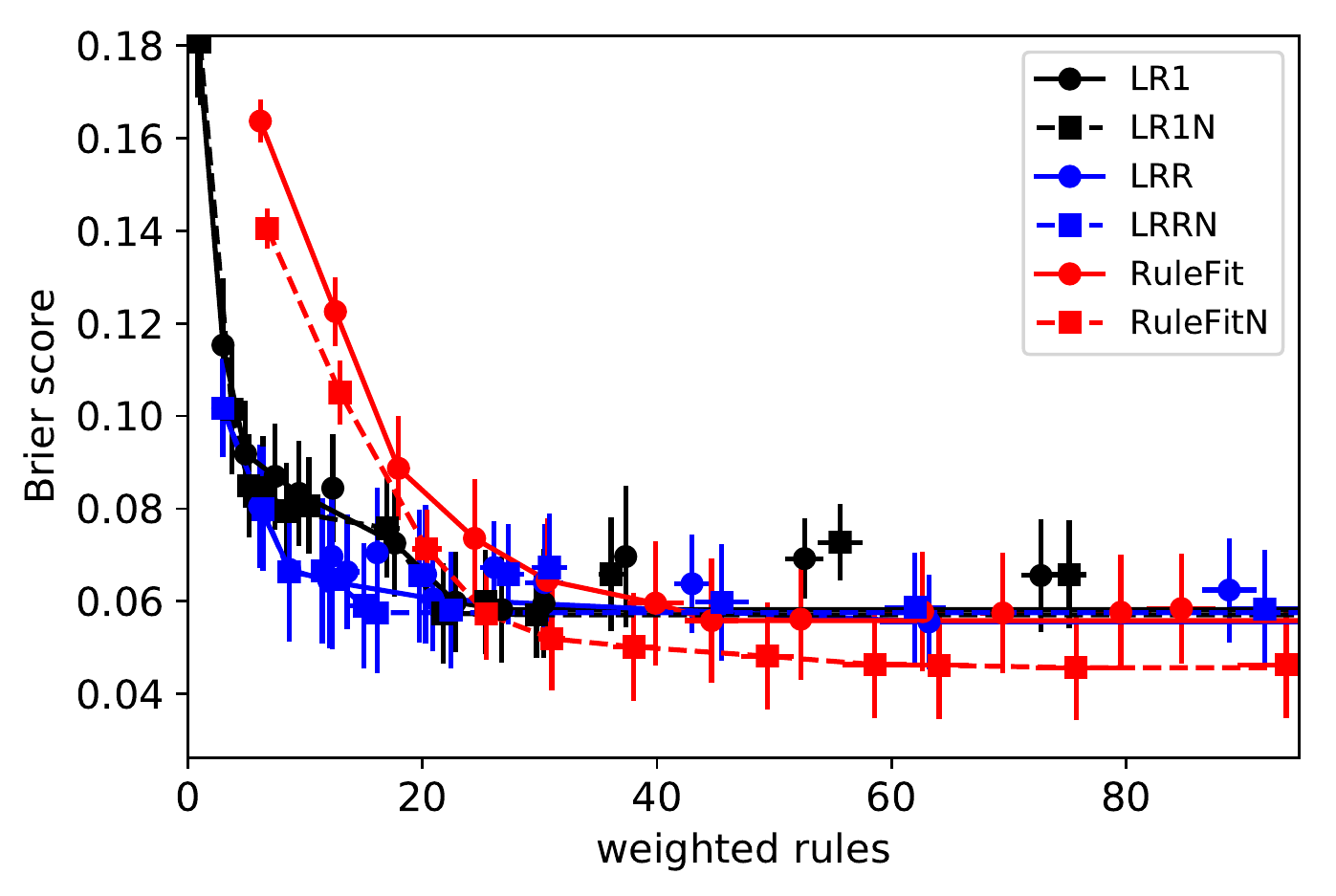}
  \caption{ionosphere}
  \label{fig:paretoClassBrier:ionosphere}
  \end{subfigure}
  \begin{subfigure}[b]{0.9\columnwidth}
  \includegraphics[width=0.9\columnwidth]{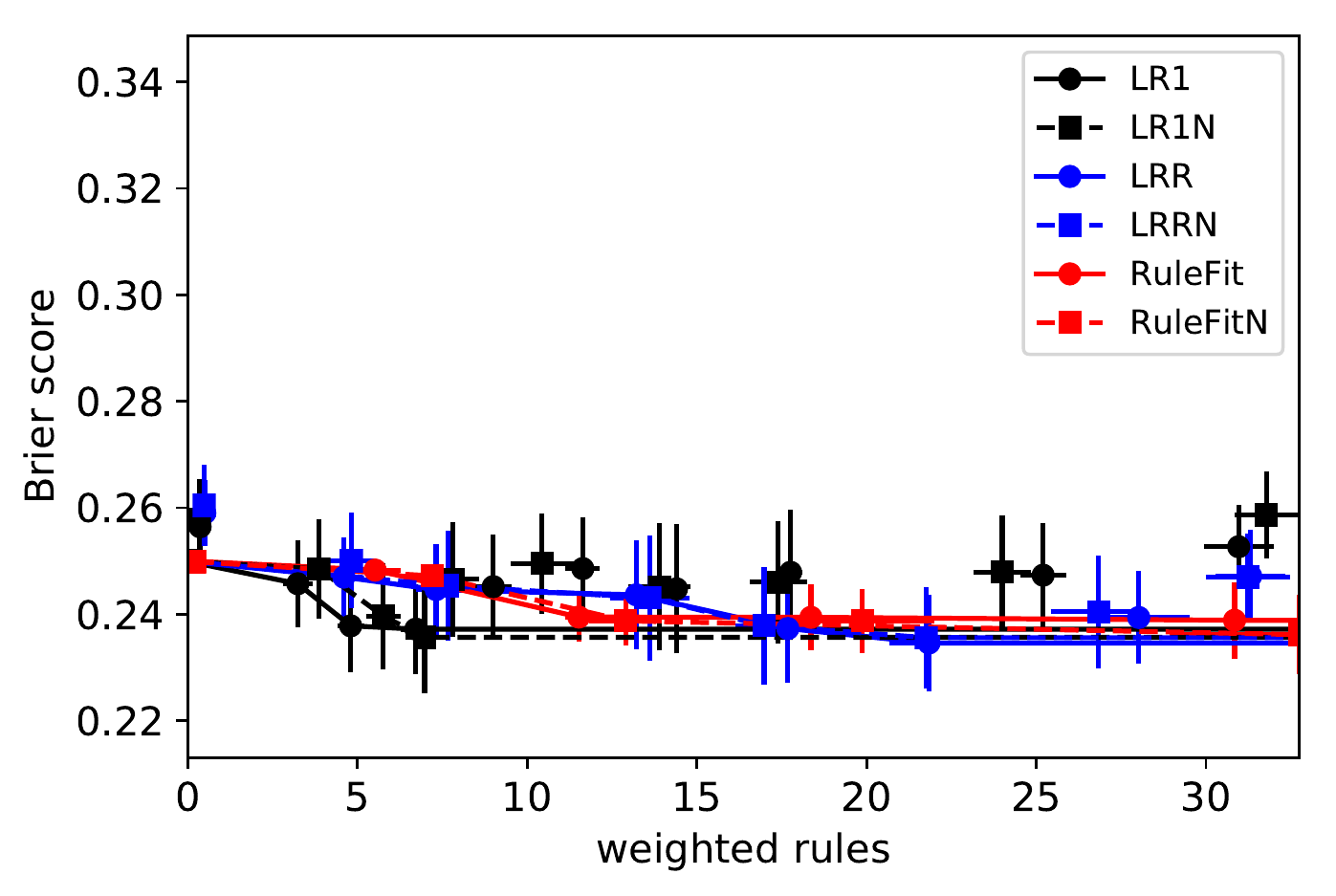}
  \caption{liver}
  \label{fig:paretoClassBrier:bupa}
  \end{subfigure}
  \begin{subfigure}[b]{0.9\columnwidth}
  \includegraphics[width=0.9\columnwidth]{paretoLRR_Brier_pima.pdf}
  \caption{pima}
  \label{fig:paretoClassBrier:pima}
  \end{subfigure}
  \begin{subfigure}[b]{0.9\columnwidth}
  \includegraphics[width=0.9\columnwidth]{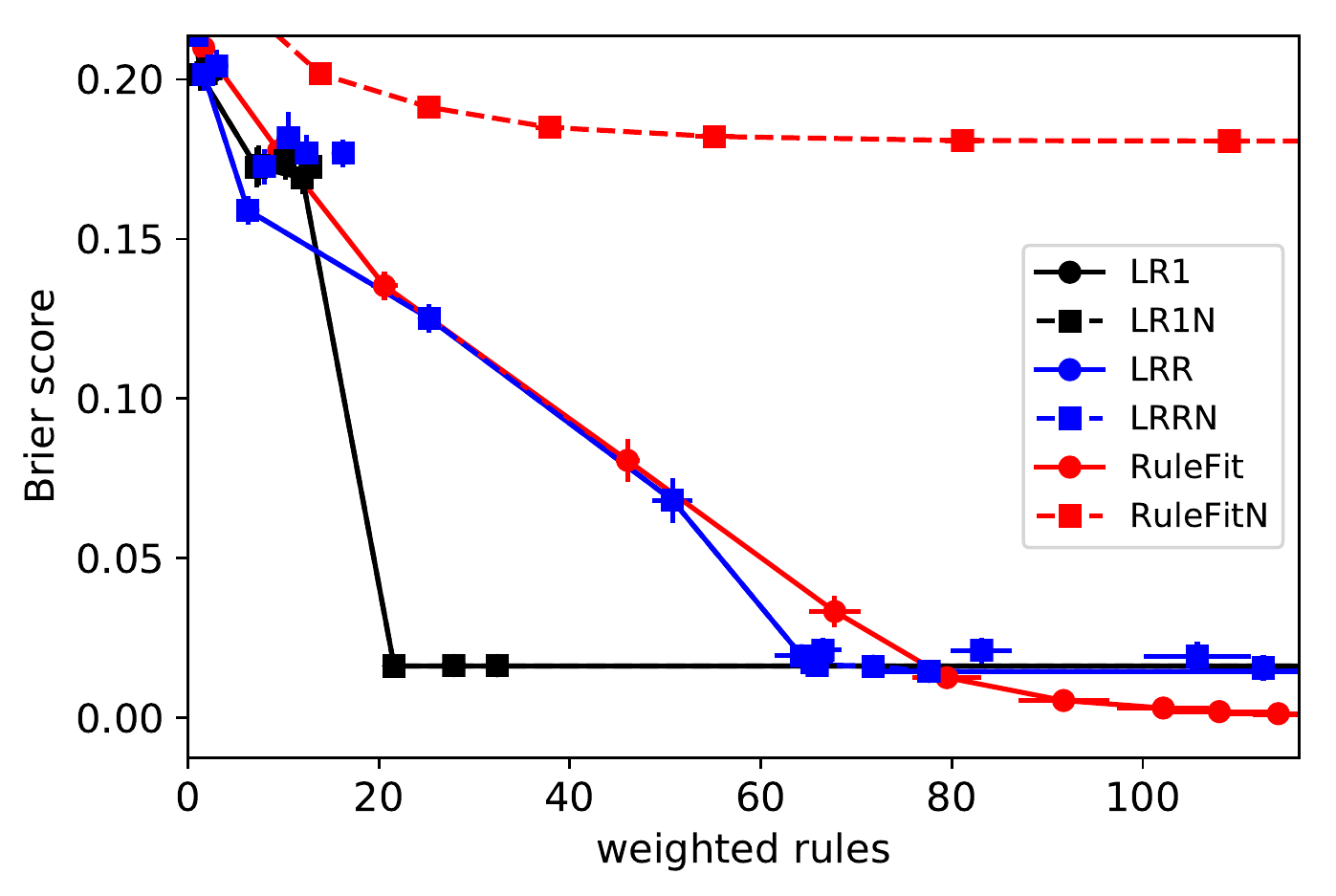}
  \caption{tic-tac-toe}
  \label{fig:paretoClassBrier:tic-tac-toe}
  \end{subfigure}
  \begin{subfigure}[b]{0.9\columnwidth}
  \includegraphics[width=0.9\columnwidth]{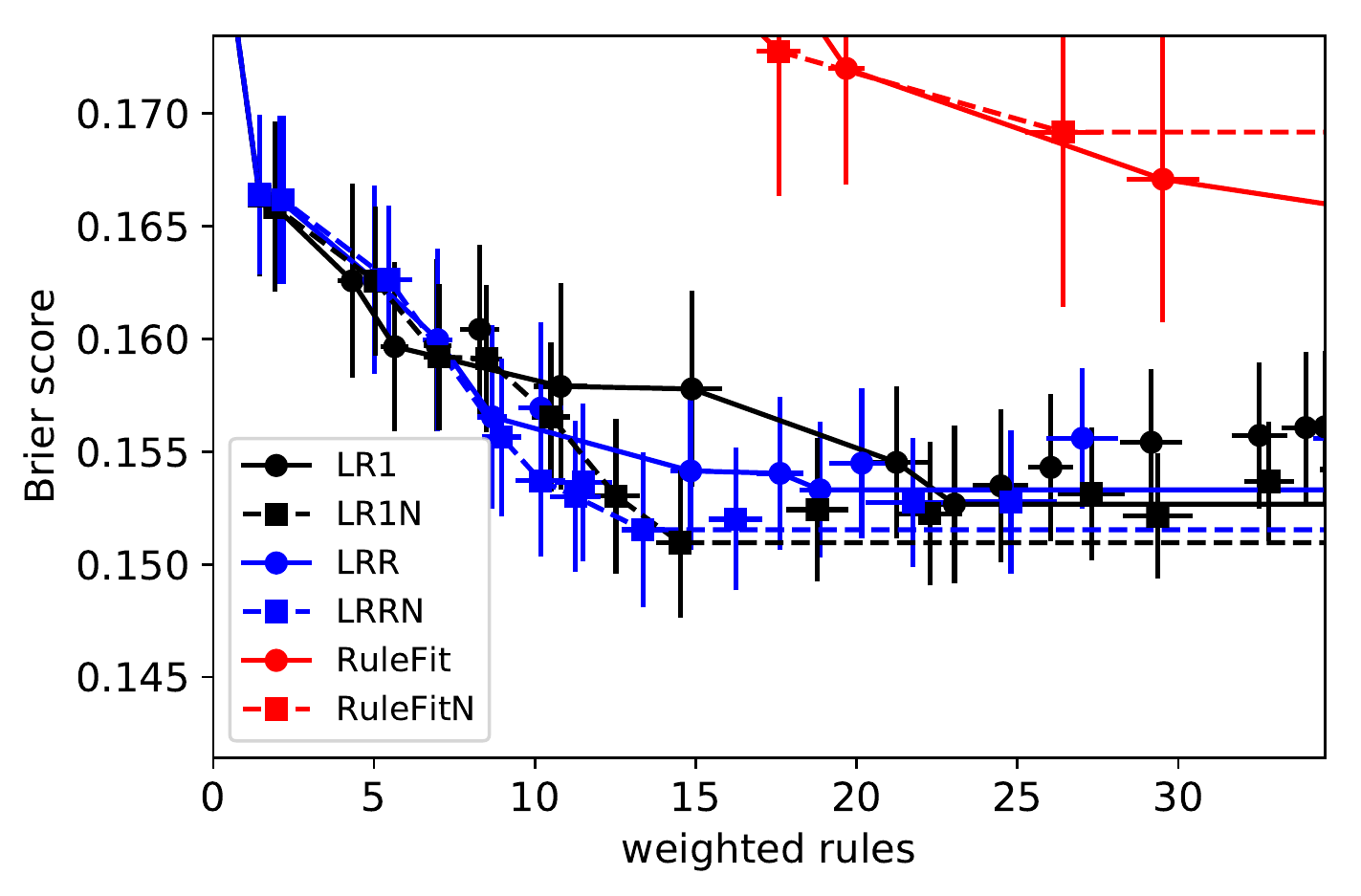}
  \caption{transfusion}
  \label{fig:paretoClassBrier:transfusion}
  \end{subfigure}
  \caption{Trade-offs between Brier score and weighted number of rules on classification datasets. Pareto efficient points are connected by line segments. Horizontal and vertical bars represent standard errors in the means.} 
  \label{fig:paretoClassBrier1}
\end{figure*}

\begin{figure*}[t]
  \centering
  \begin{subfigure}[b]{0.9\columnwidth}
  \includegraphics[width=0.9\columnwidth]{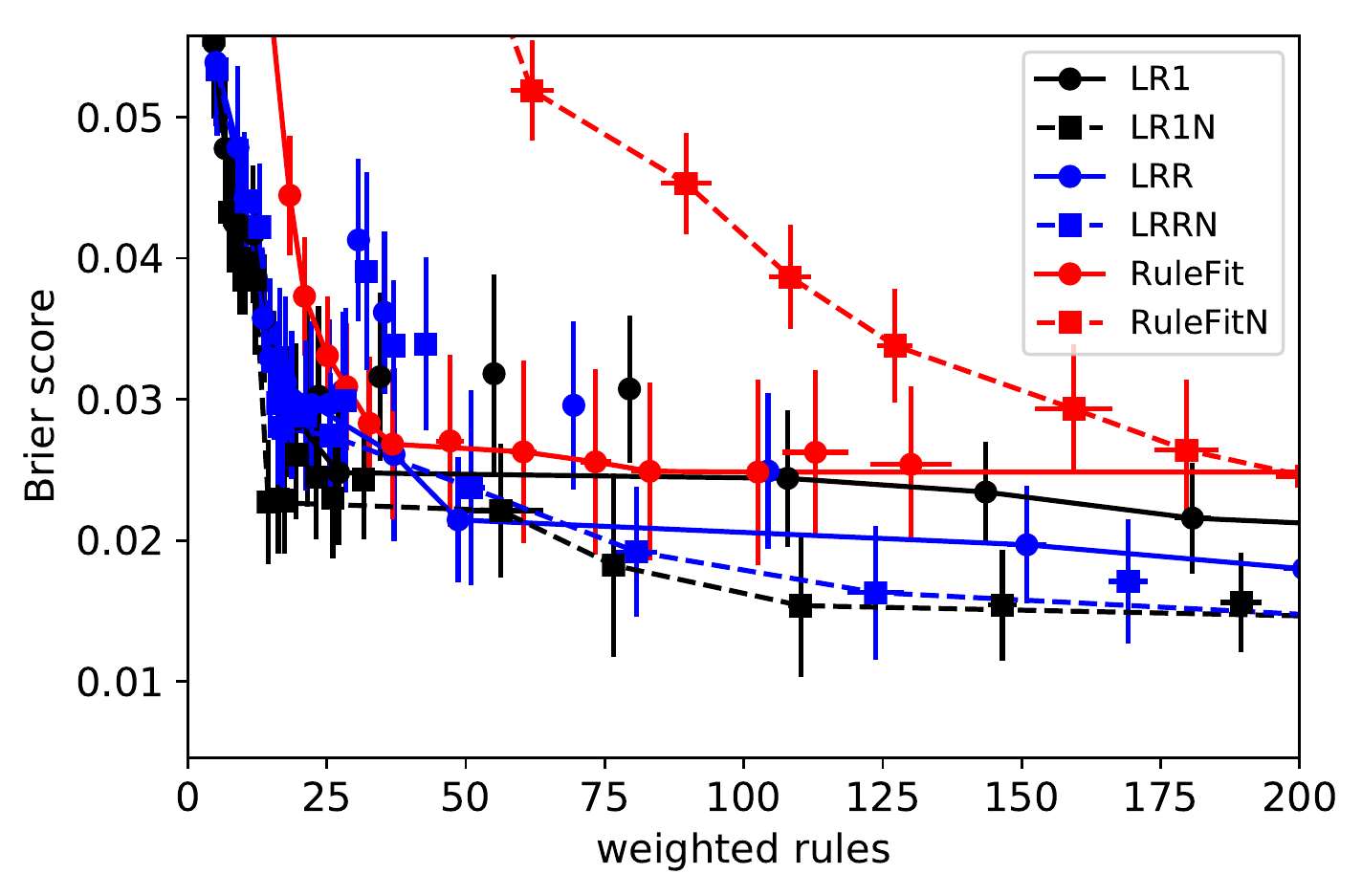}
  \caption{WDBC}
  \label{fig:paretoClassBrier:wdbc}
  \end{subfigure}
  \begin{subfigure}[b]{0.9\columnwidth}
  \includegraphics[width=0.9\columnwidth]{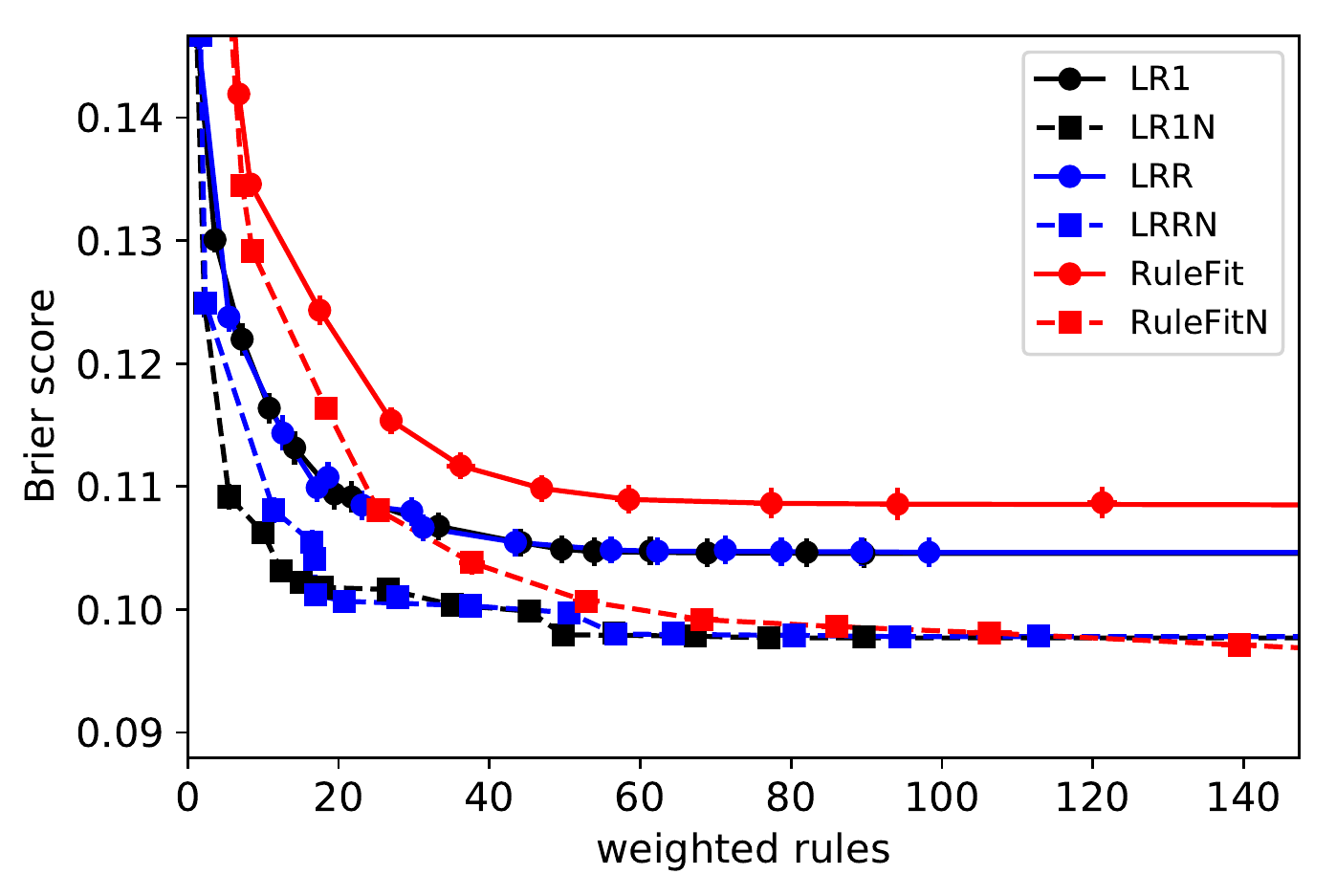}
  \caption{adult}
  \label{fig:paretoClassBrier:adult}
  \end{subfigure}
  \begin{subfigure}[b]{0.9\columnwidth}
  \includegraphics[width=0.9\columnwidth]{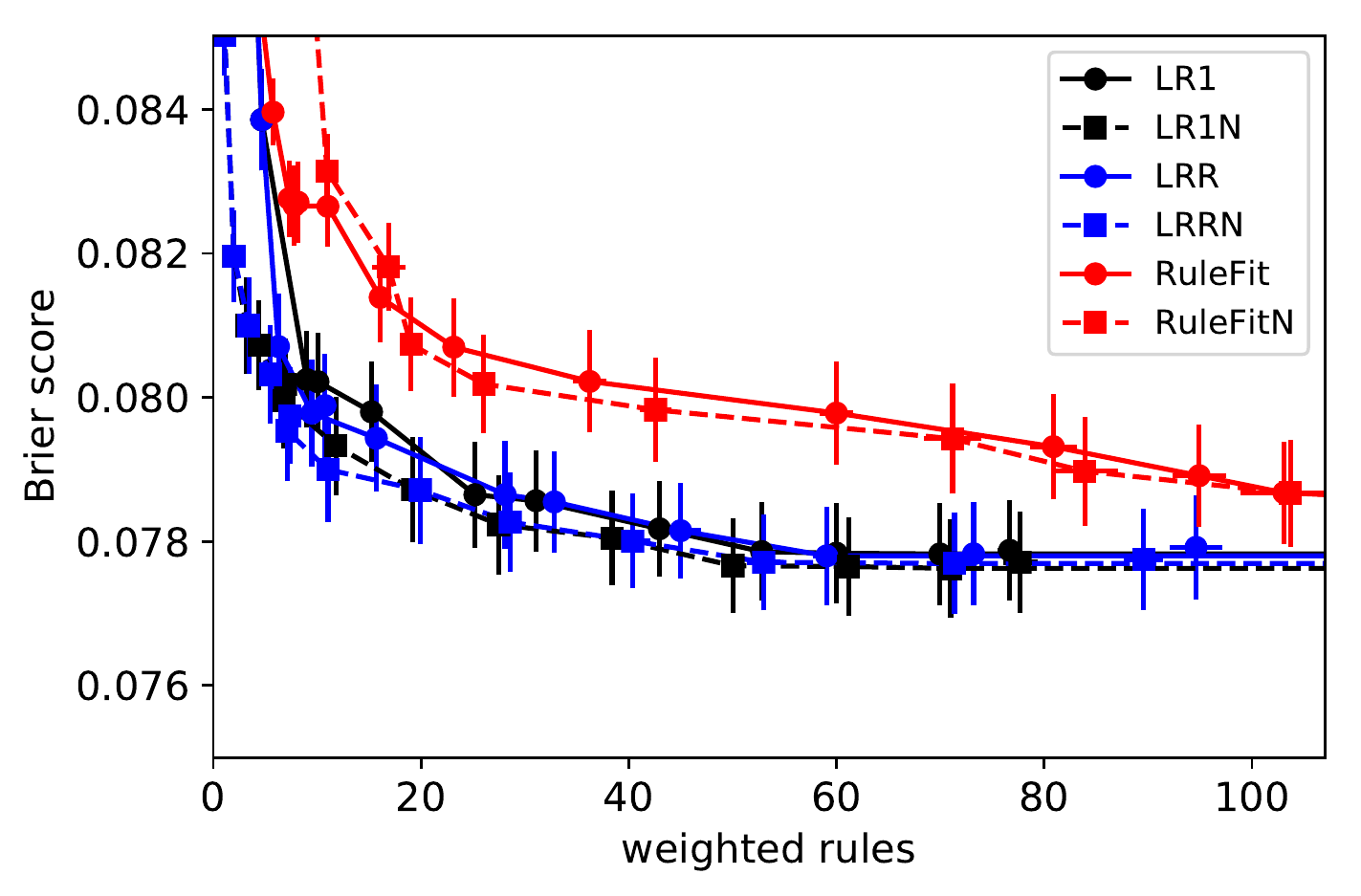}
  \caption{bank-marketing}
  \label{fig:paretoClassBrier:bank}
  \end{subfigure}
  \begin{subfigure}[b]{0.9\columnwidth}
  \includegraphics[width=0.9\columnwidth]{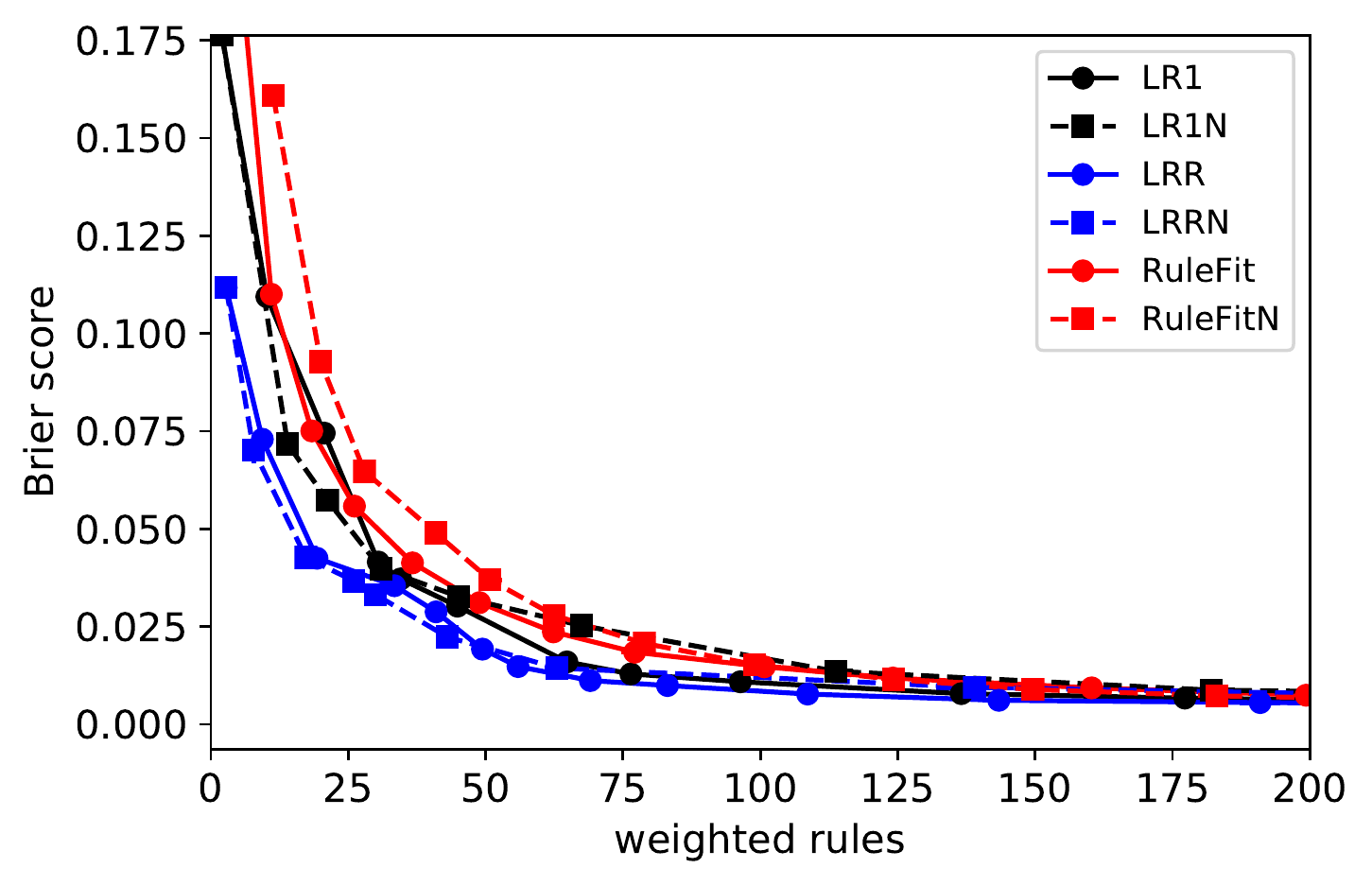}
  \caption{gas}
  \label{fig:paretoClassBrier:gas}
  \end{subfigure}
  \begin{subfigure}[b]{0.9\columnwidth}
  \includegraphics[width=0.9\columnwidth]{paretoLRR_Brier_magic.pdf}
  \caption{magic}
  \label{fig:paretoClassBrier:magic}
  \end{subfigure}
  \begin{subfigure}[b]{0.9\columnwidth}
  \includegraphics[width=0.9\columnwidth]{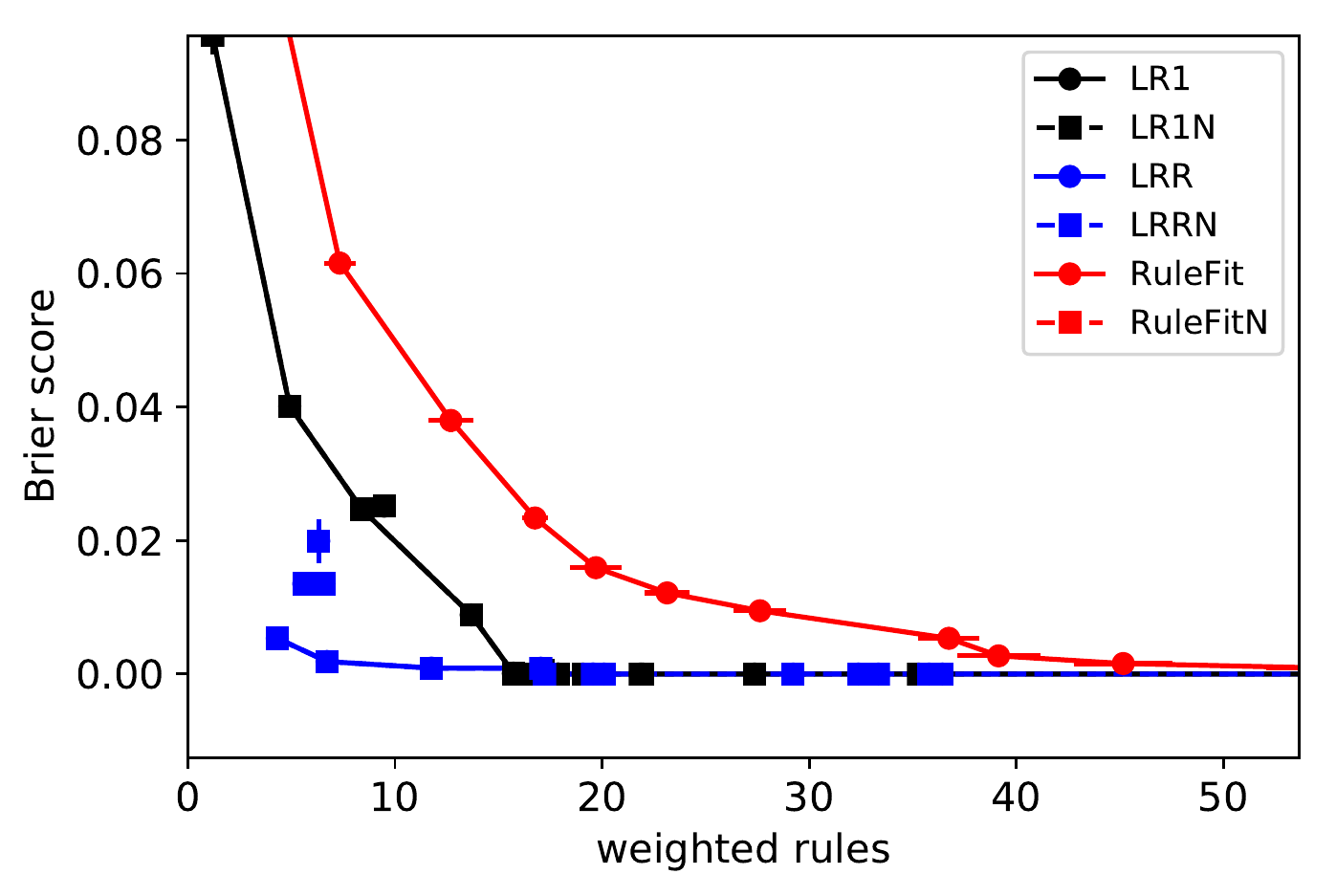}
  \caption{mushroom}
  \label{fig:paretoClassBrier:mushroom}
  \end{subfigure}
  \begin{subfigure}[b]{0.9\columnwidth}
  \includegraphics[width=0.9\columnwidth]{paretoLRR_Brier_musk.pdf}
  \caption{musk}
  \label{fig:paretoClassBrier:musk}
  \end{subfigure}
  \begin{subfigure}[b]{0.9\columnwidth}
  \includegraphics[width=0.9\columnwidth]{paretoLRR_Brier_FICO.pdf}
  \caption{FICO}
  \label{fig:paretoClassBrier:FICO}
  \end{subfigure}
  \caption{Trade-offs between Brier score and weighted number of rules on classification datasets. Pareto efficient points are connected by line segments. Horizontal and vertical bars represent standard errors in the means.} 
  \label{fig:paretoClassBrier2}
\end{figure*}

\begin{figure*}[t]
  \centering
  \begin{subfigure}[b]{0.9\columnwidth}
  \includegraphics[width=0.9\columnwidth]{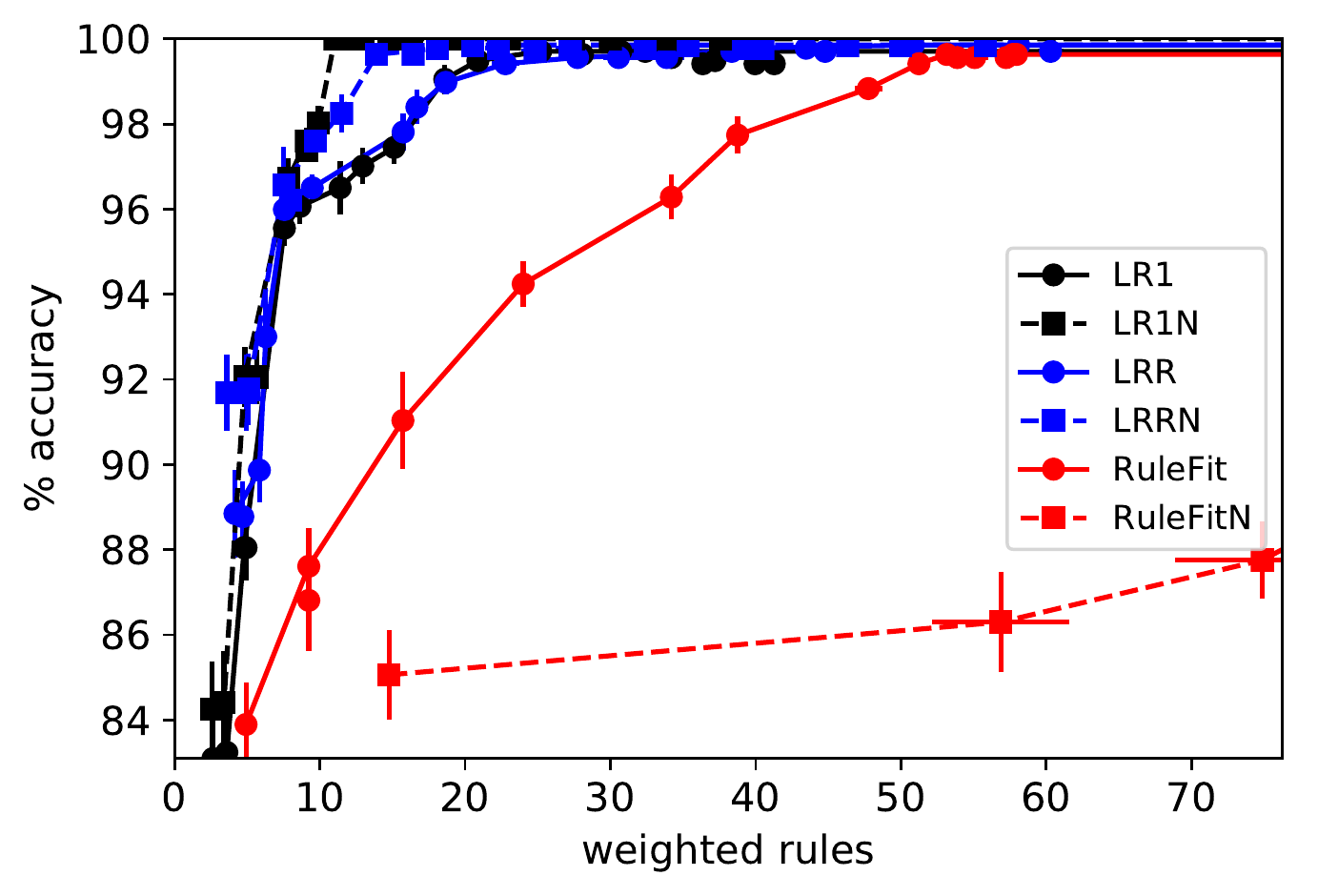}
  \caption{banknote}
  \label{fig:paretoClassAcc:banknote}
  \end{subfigure}
  \begin{subfigure}[b]{0.9\columnwidth}
  \includegraphics[width=0.9\columnwidth]{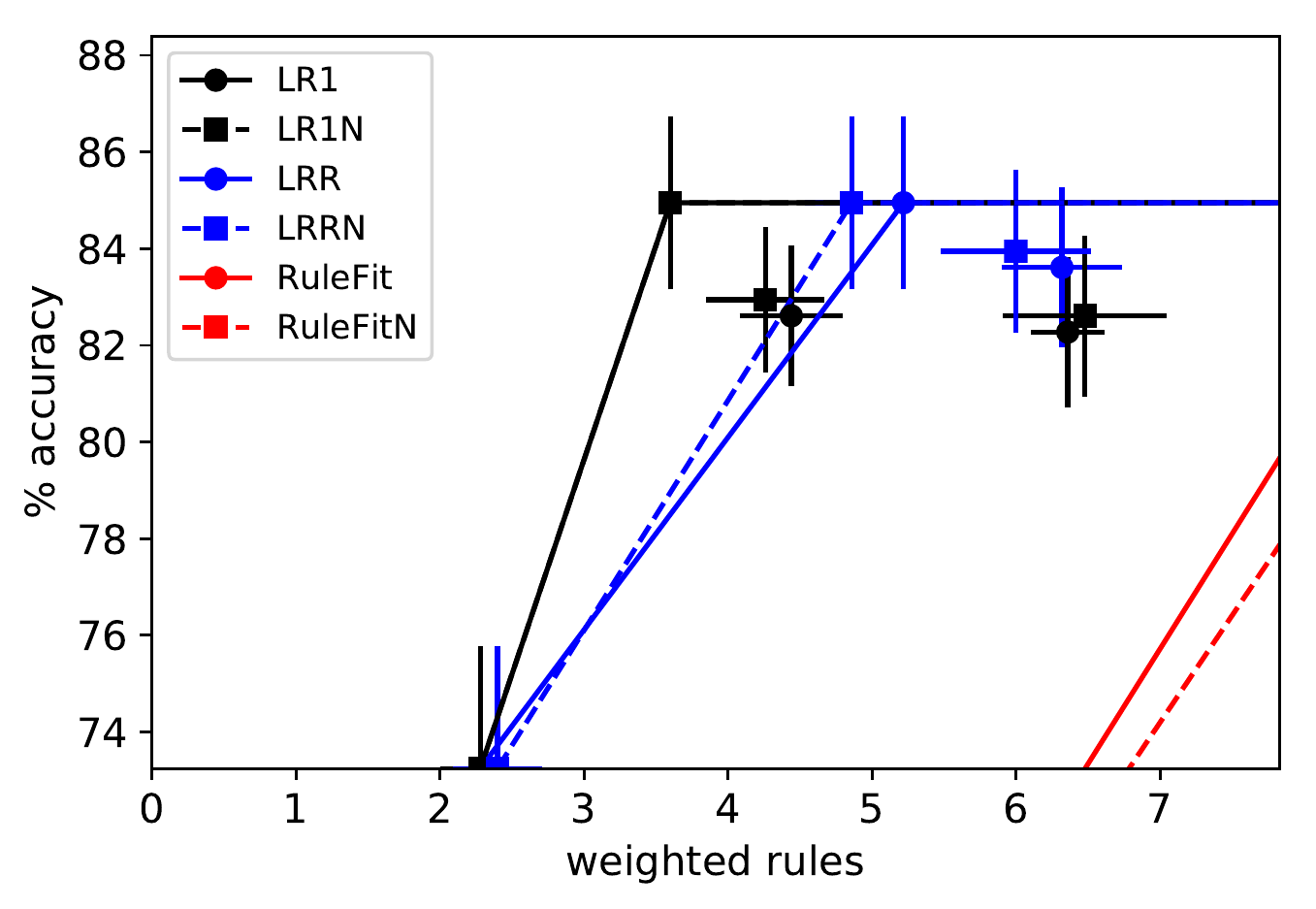}
  \caption{heart}
  \label{fig:paretoClassAcc:heart}
  \end{subfigure}
  \begin{subfigure}[b]{0.9\columnwidth}
  \includegraphics[width=0.9\columnwidth]{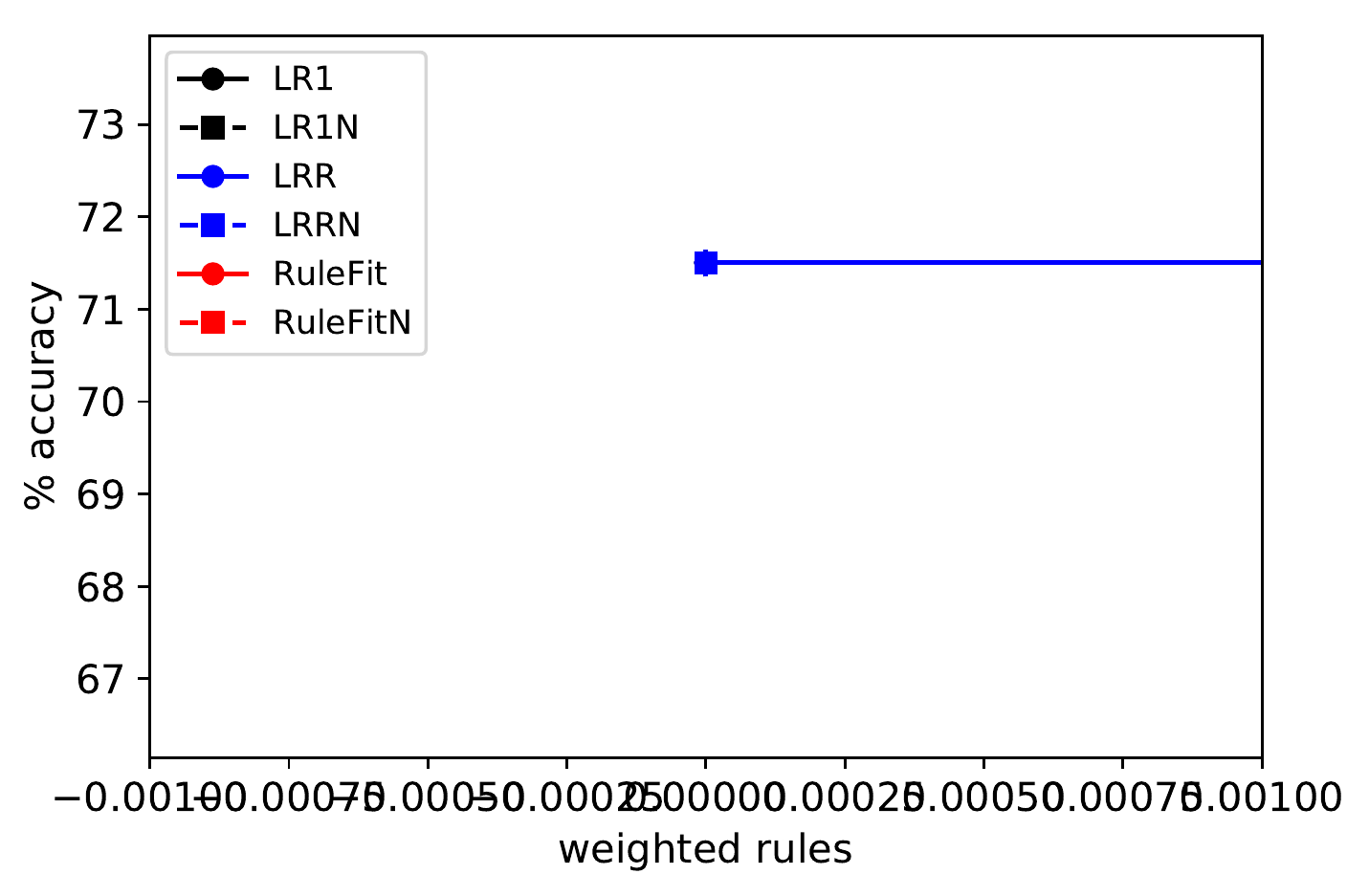}
  \caption{ILPD}
  \label{fig:paretoClassAcc:ILPD}
  \end{subfigure}
  \begin{subfigure}[b]{0.9\columnwidth}
  \includegraphics[width=0.9\columnwidth]{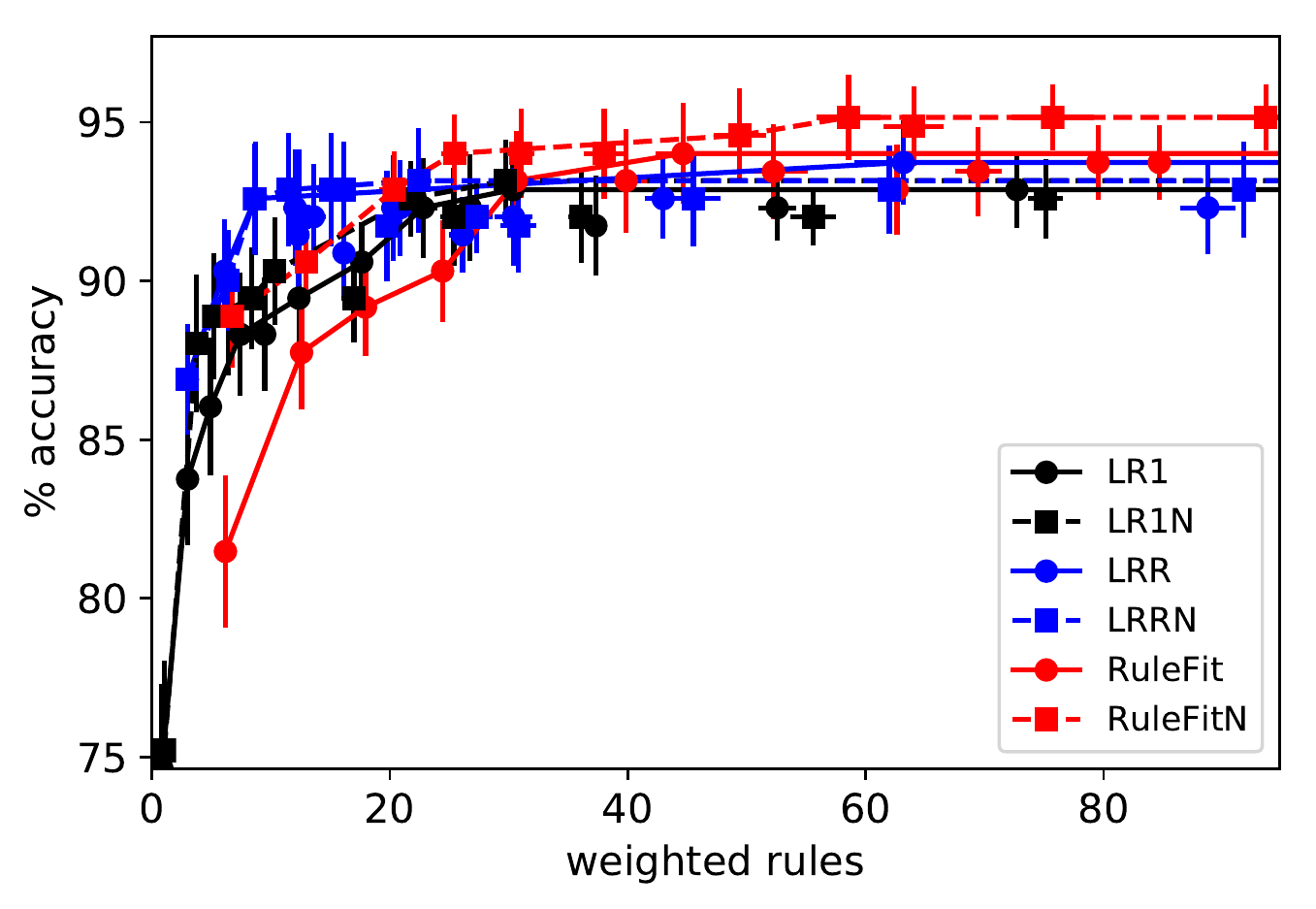}
  \caption{ionosphere}
  \label{fig:paretoClassAcc:ionosphere}
  \end{subfigure}
  \begin{subfigure}[b]{0.9\columnwidth}
  \includegraphics[width=0.9\columnwidth]{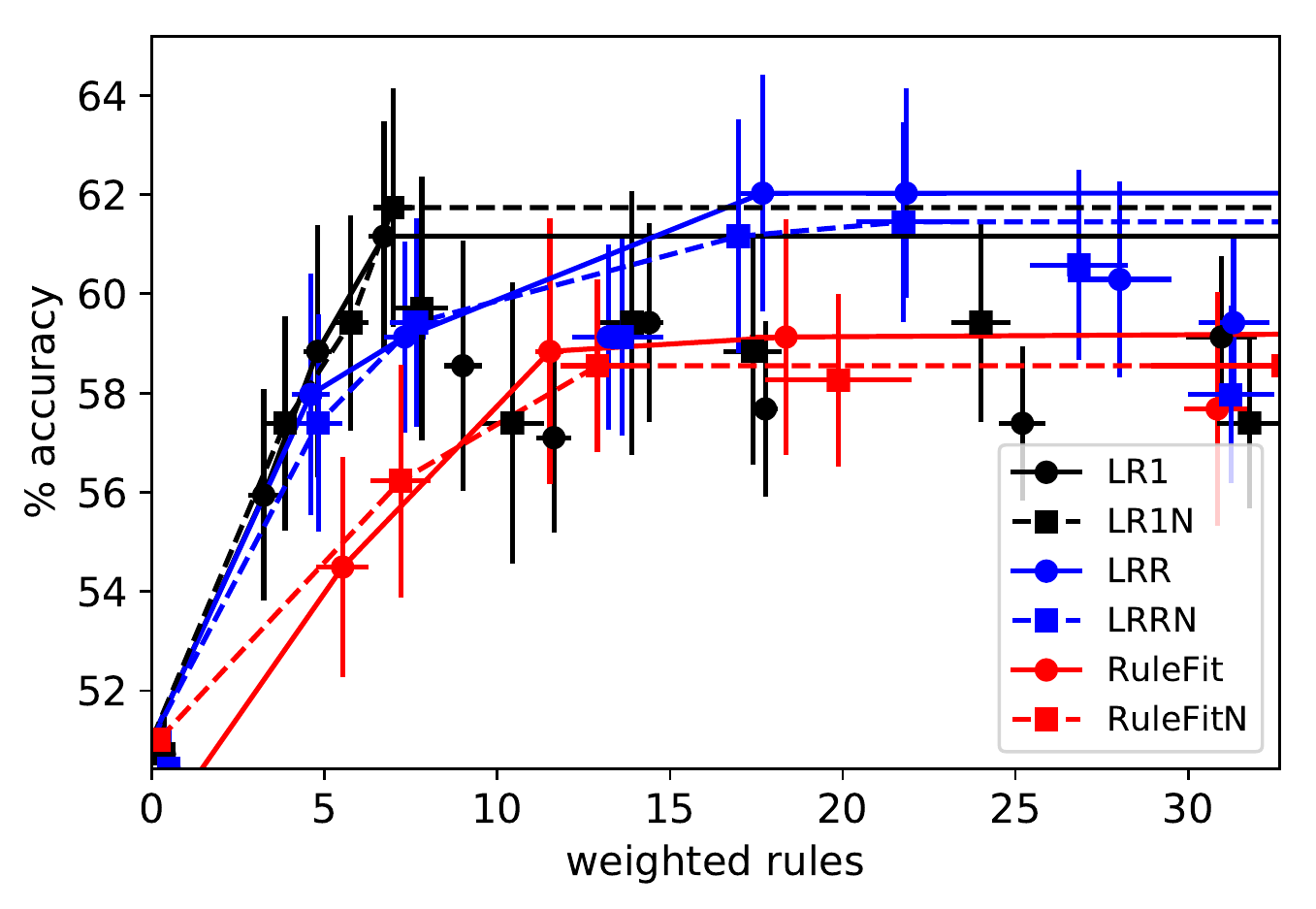}
  \caption{liver}
  \label{fig:paretoClassAcc:bupa}
  \end{subfigure}
  \begin{subfigure}[b]{0.9\columnwidth}
  \includegraphics[width=0.9\columnwidth]{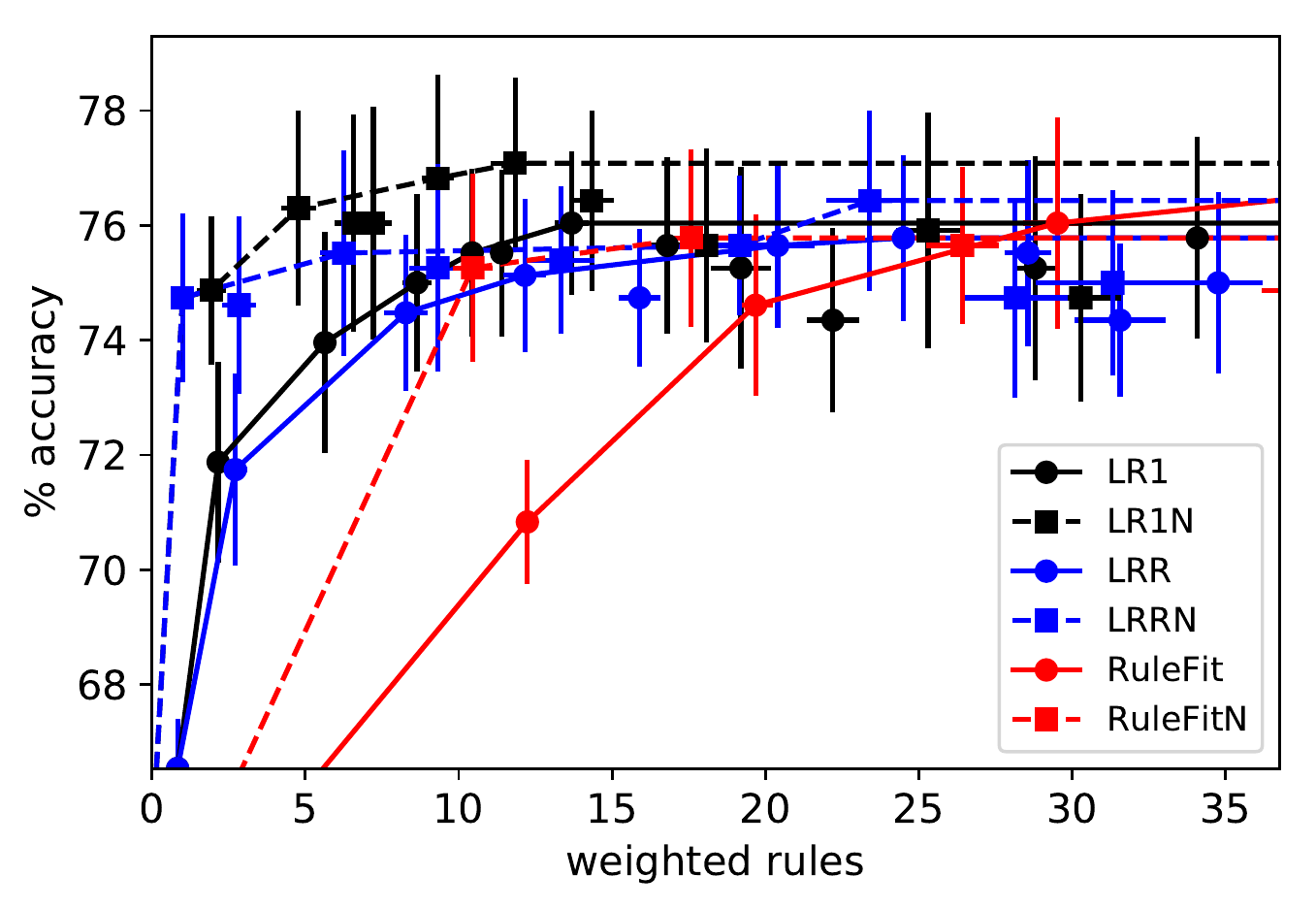}
  \caption{pima}
  \label{fig:paretoClassAcc:pima}
  \end{subfigure}
  \begin{subfigure}[b]{0.9\columnwidth}
  \includegraphics[width=0.9\columnwidth]{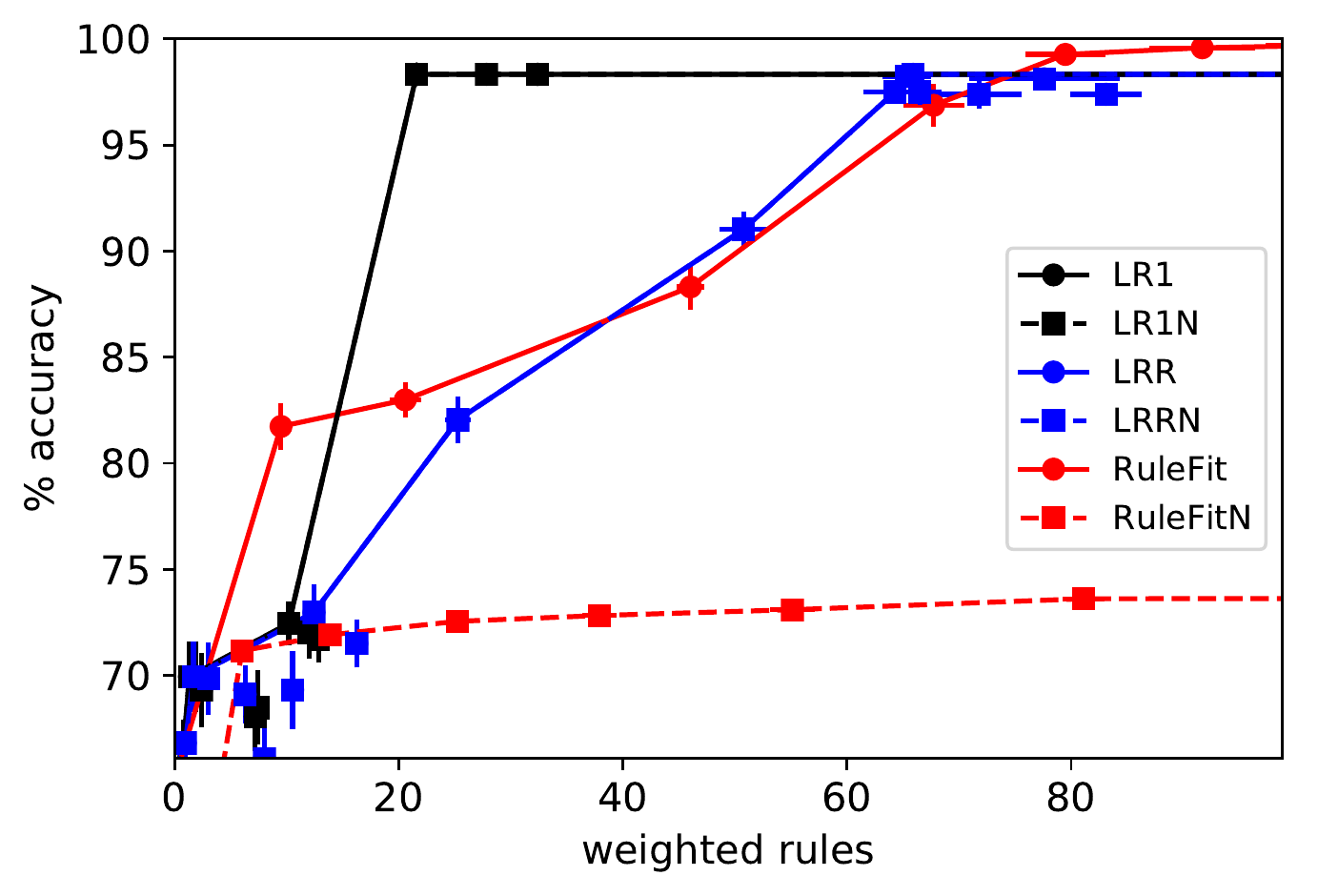}
  \caption{tic-tac-toe}
  \label{fig:paretoClassAcc:tic-tac-toe}
  \end{subfigure}
  \begin{subfigure}[b]{0.9\columnwidth}
  \includegraphics[width=0.9\columnwidth]{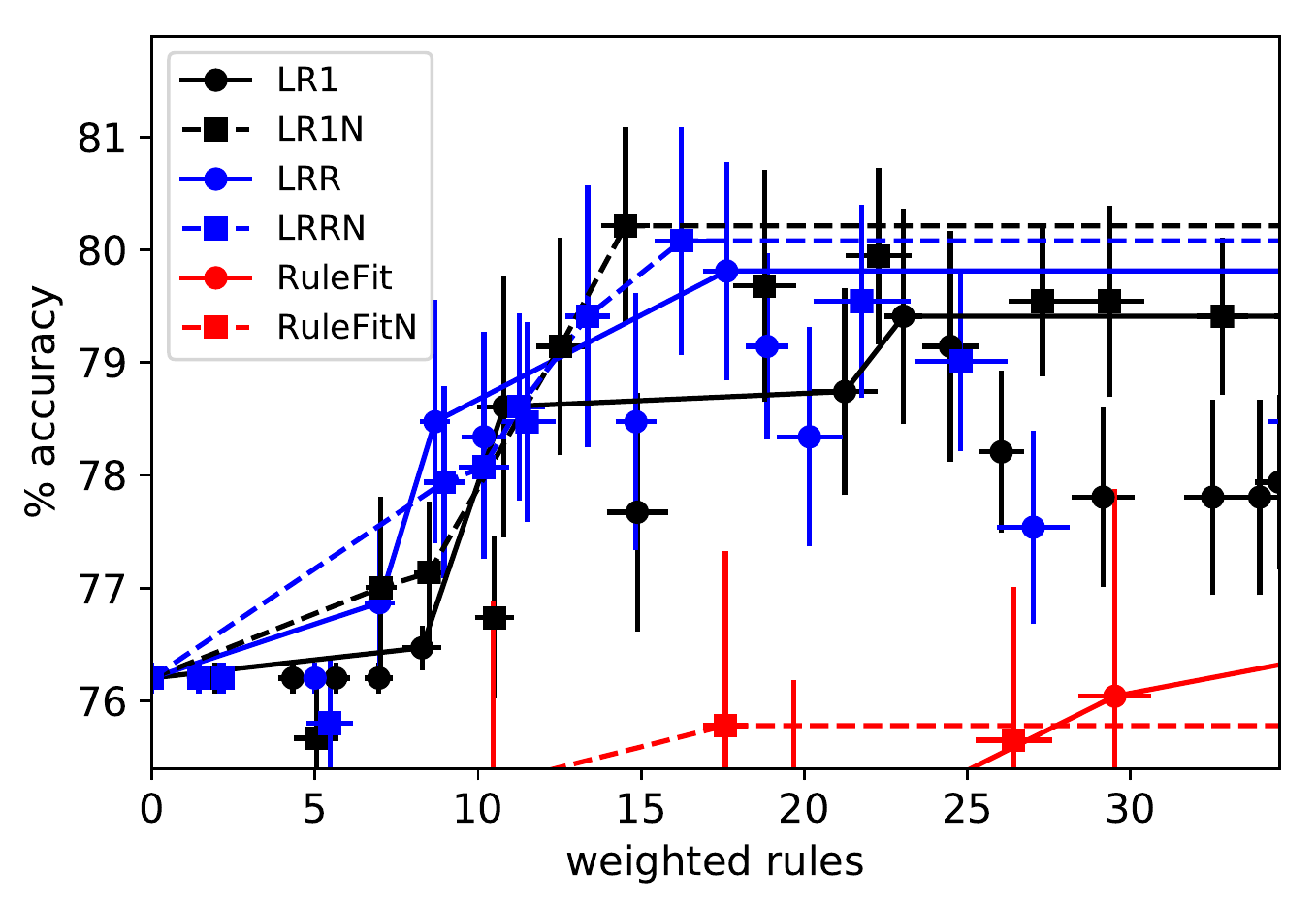}
  \caption{transfusion}
  \label{fig:paretoClassAcc:transfusion}
  \end{subfigure}
  \caption{Trade-offs between accuracy and weighted number of rules on classification datasets. Pareto efficient points are connected by line segments. Horizontal and vertical bars represent standard errors in the means.} 
  \label{fig:paretoClassAcc1}
\end{figure*}

\begin{figure*}[t]
  \centering
  \begin{subfigure}[b]{0.9\columnwidth}
  \includegraphics[width=0.9\columnwidth]{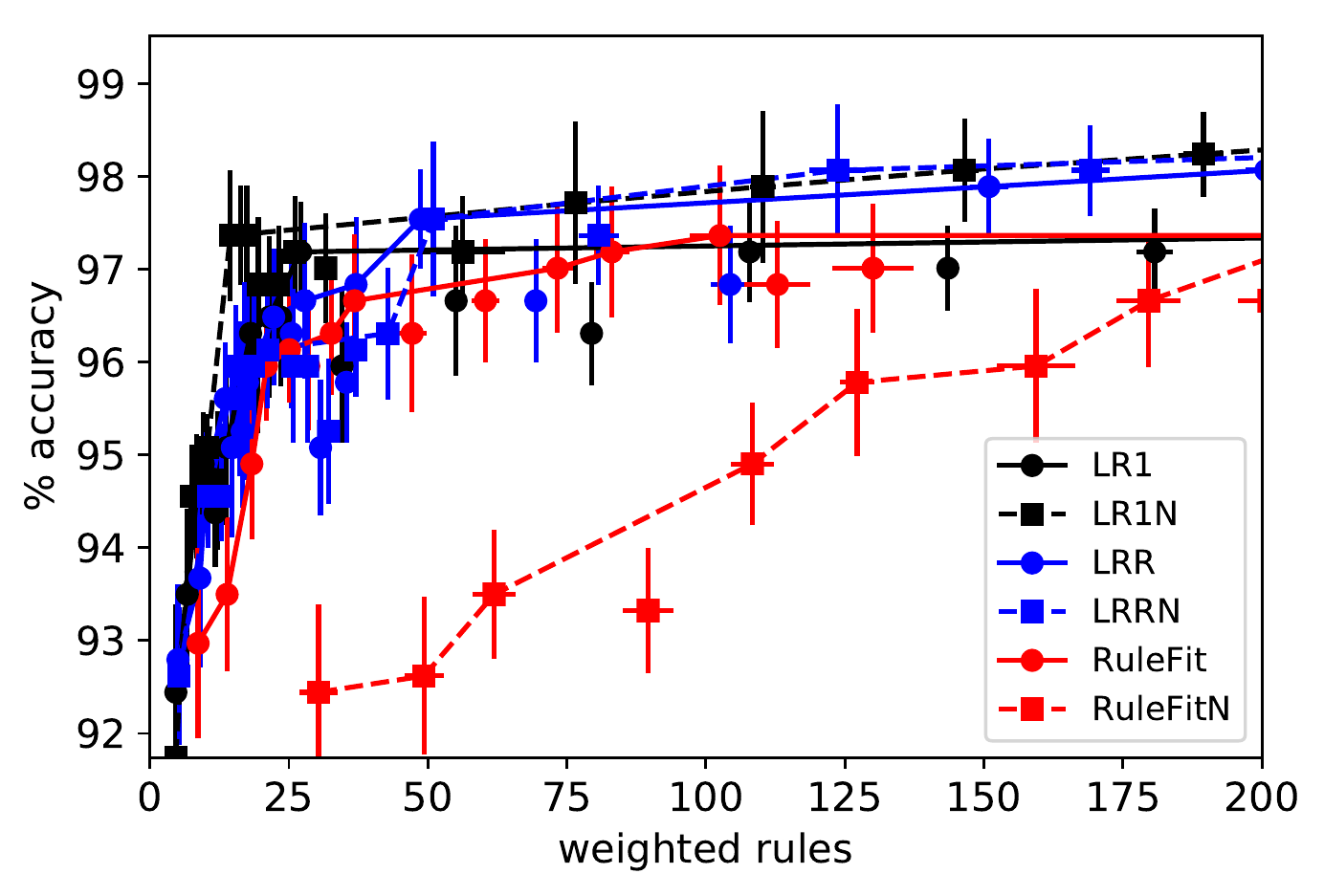}
  \caption{WDBC}
  \label{fig:paretoClassAcc:wdbc}
  \end{subfigure}
  \begin{subfigure}[b]{0.9\columnwidth}
  \includegraphics[width=0.9\columnwidth]{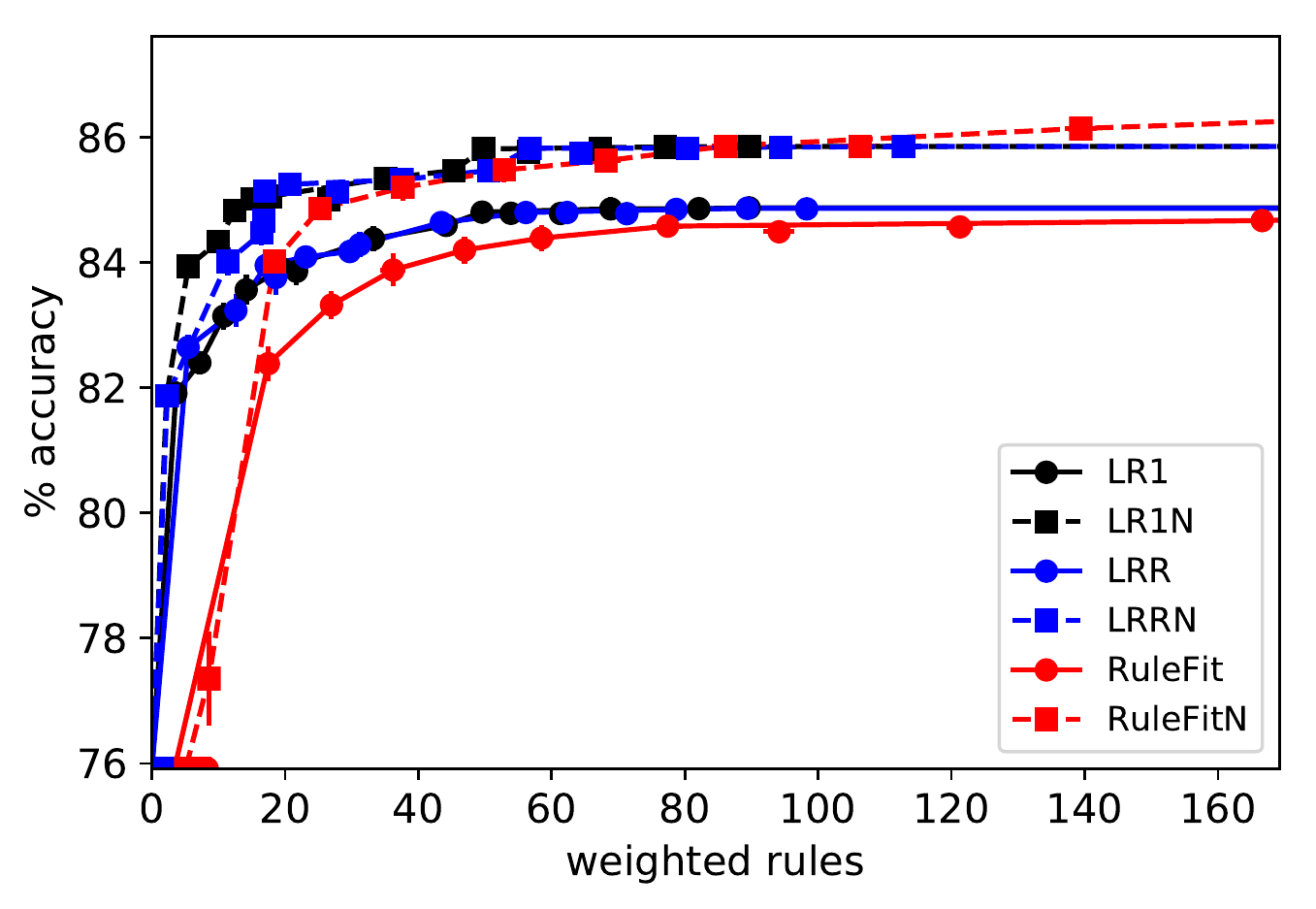}
  \caption{adult}
  \label{fig:paretoClassAcc:adult}
  \end{subfigure}
  \begin{subfigure}[b]{0.9\columnwidth}
  \includegraphics[width=0.9\columnwidth]{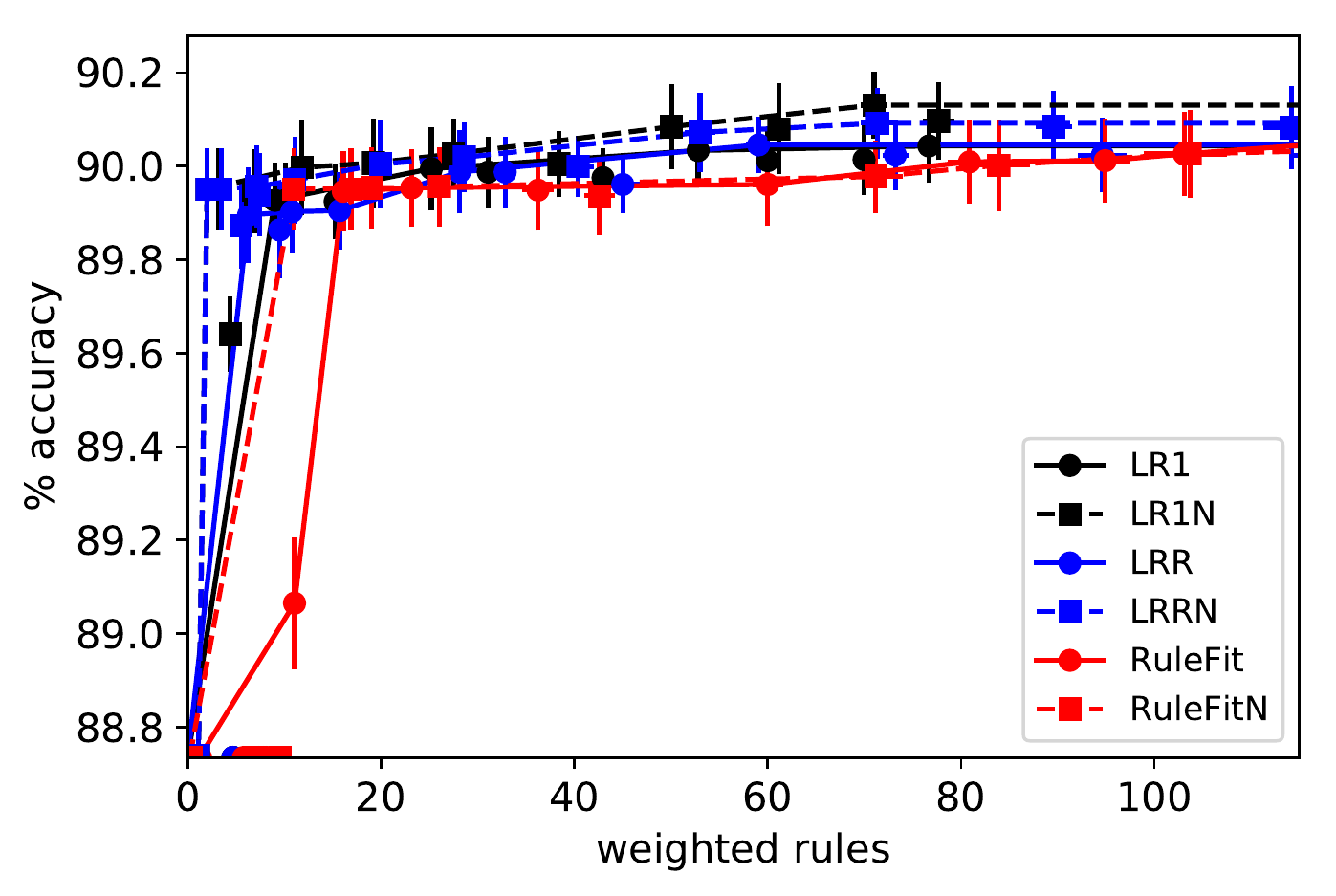}
  \caption{bank-marketing}
  \label{fig:paretoClassAcc:bank}
  \end{subfigure}
  \begin{subfigure}[b]{0.9\columnwidth}
  \includegraphics[width=0.9\columnwidth]{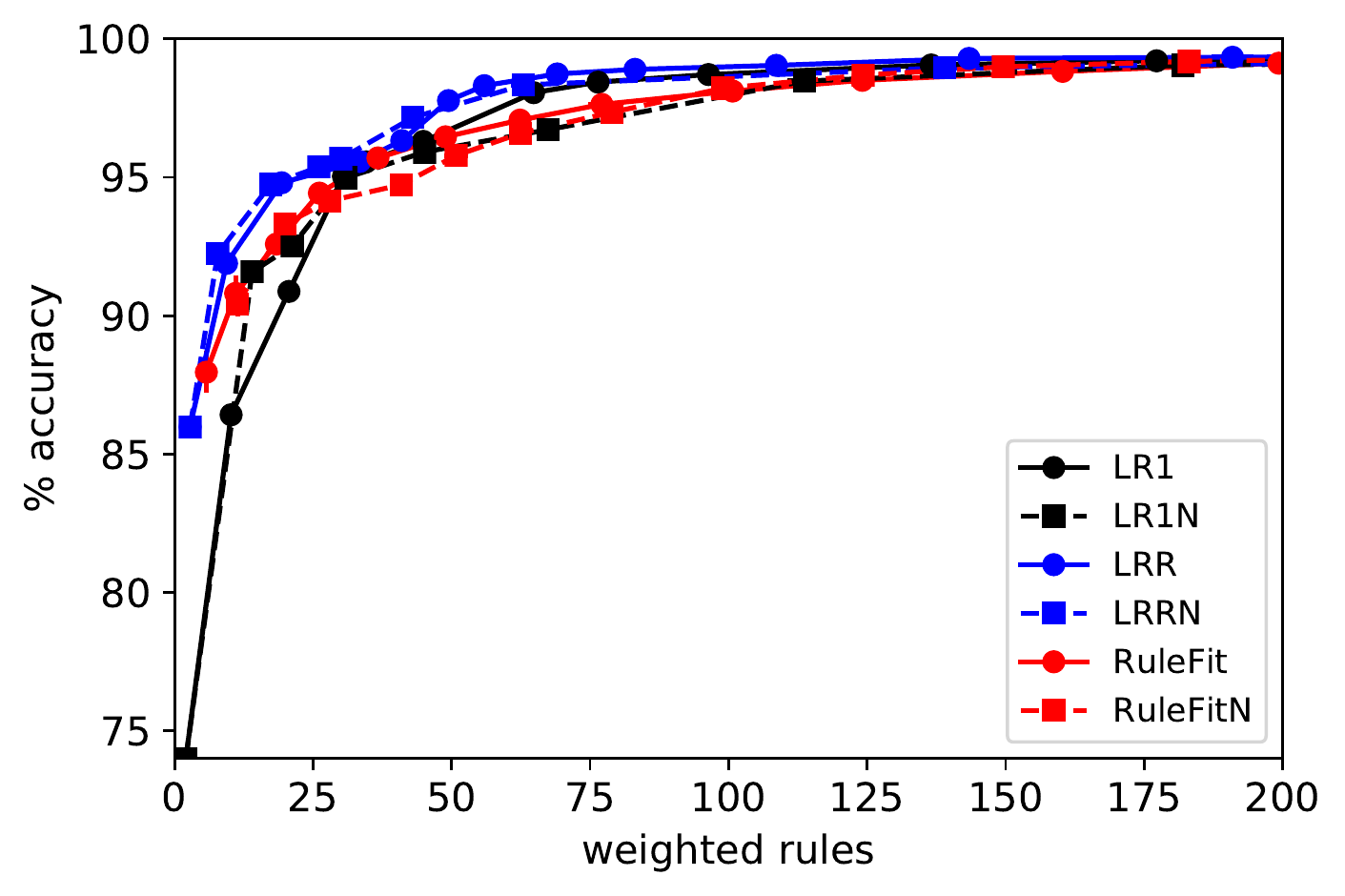}
  \caption{gas}
  \label{fig:paretoClassAcc:gas}
  \end{subfigure}
  \begin{subfigure}[b]{0.9\columnwidth}
  \includegraphics[width=0.9\columnwidth]{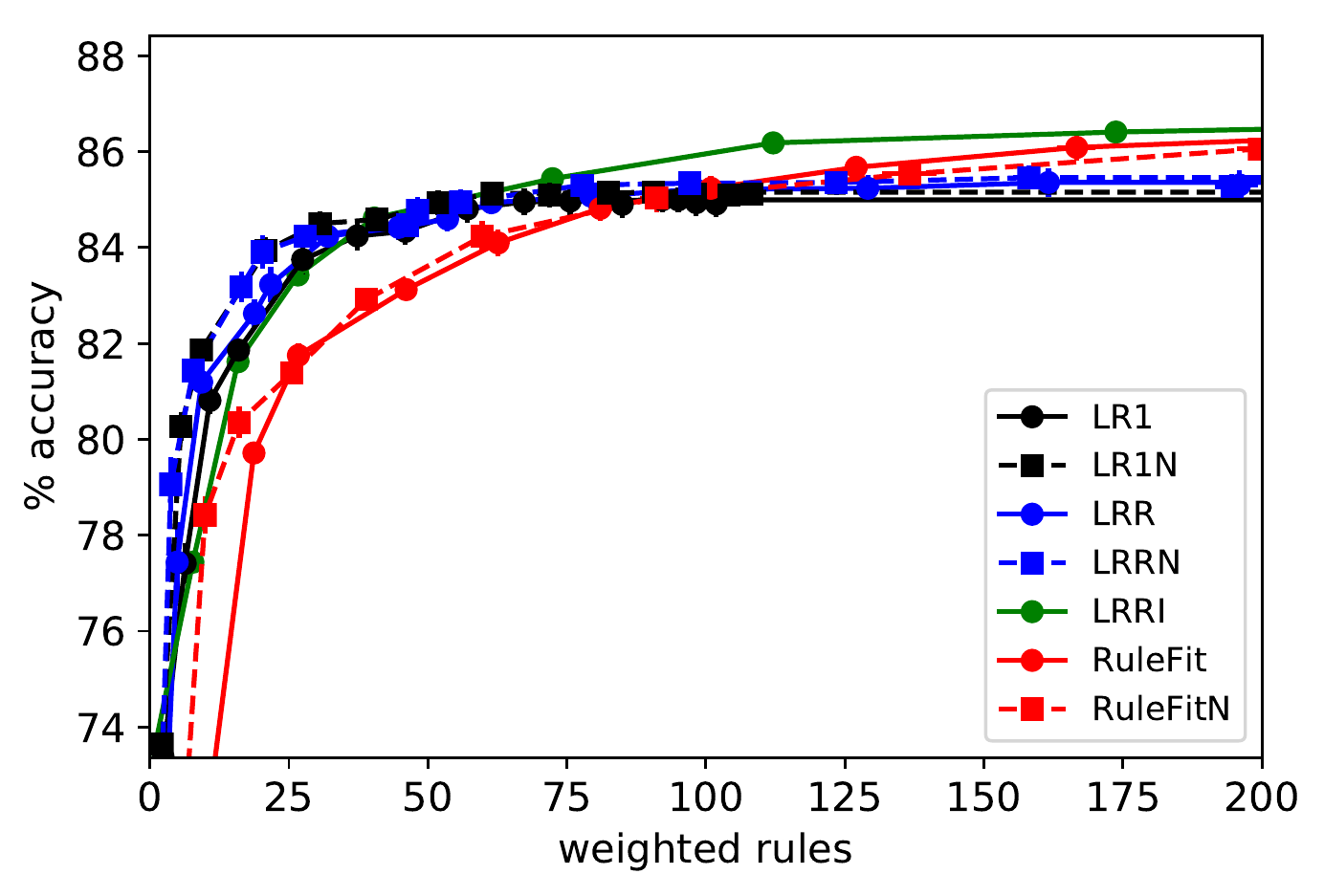}
  \caption{magic}
  \label{fig:paretoClassAcc:magic}
  \end{subfigure}
  \begin{subfigure}[b]{0.9\columnwidth}
  \includegraphics[width=0.9\columnwidth]{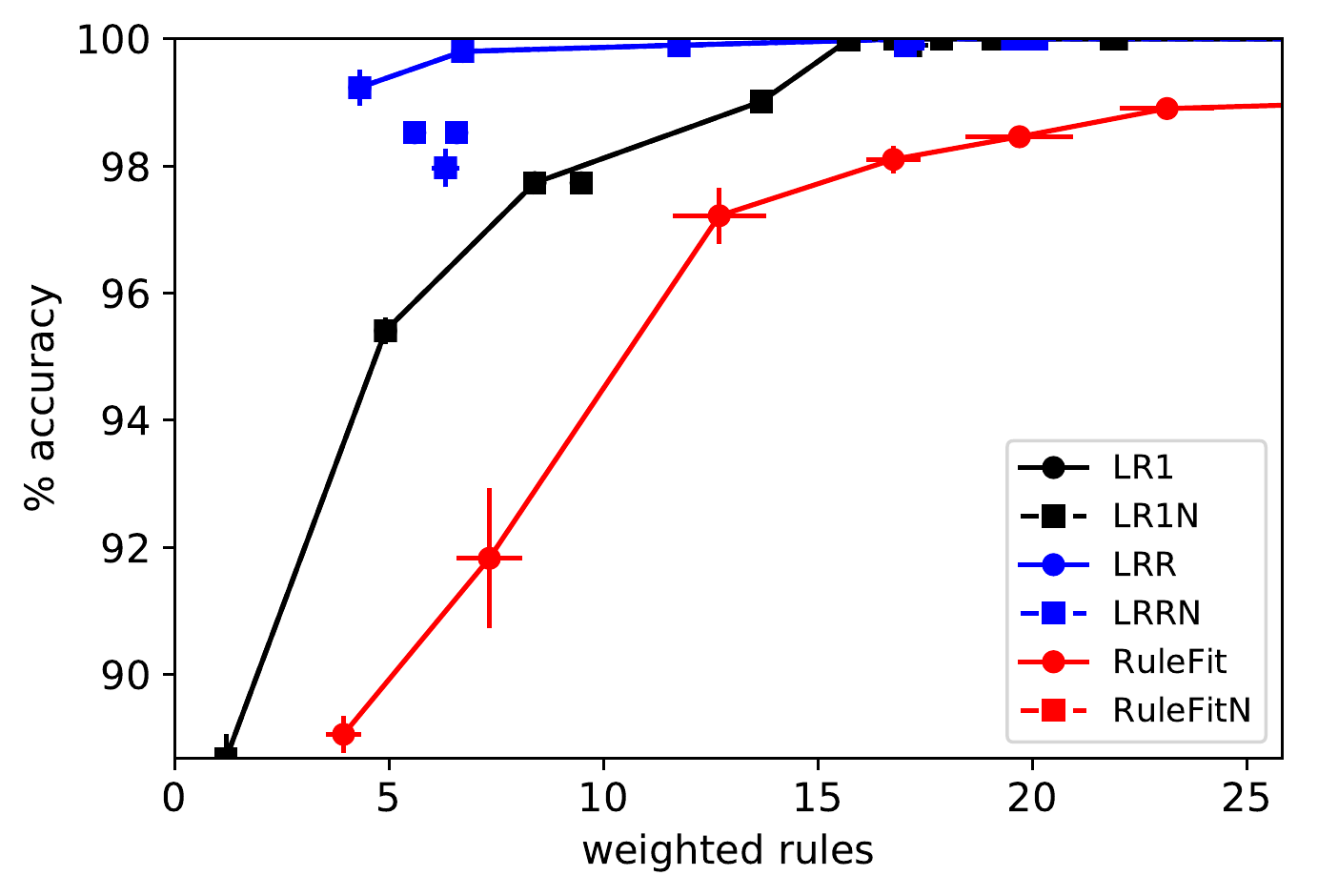}
  \caption{mushroom}
  \label{fig:paretoClassAcc:mushroom}
  \end{subfigure}
  \begin{subfigure}[b]{0.9\columnwidth}
  \includegraphics[width=0.9\columnwidth]{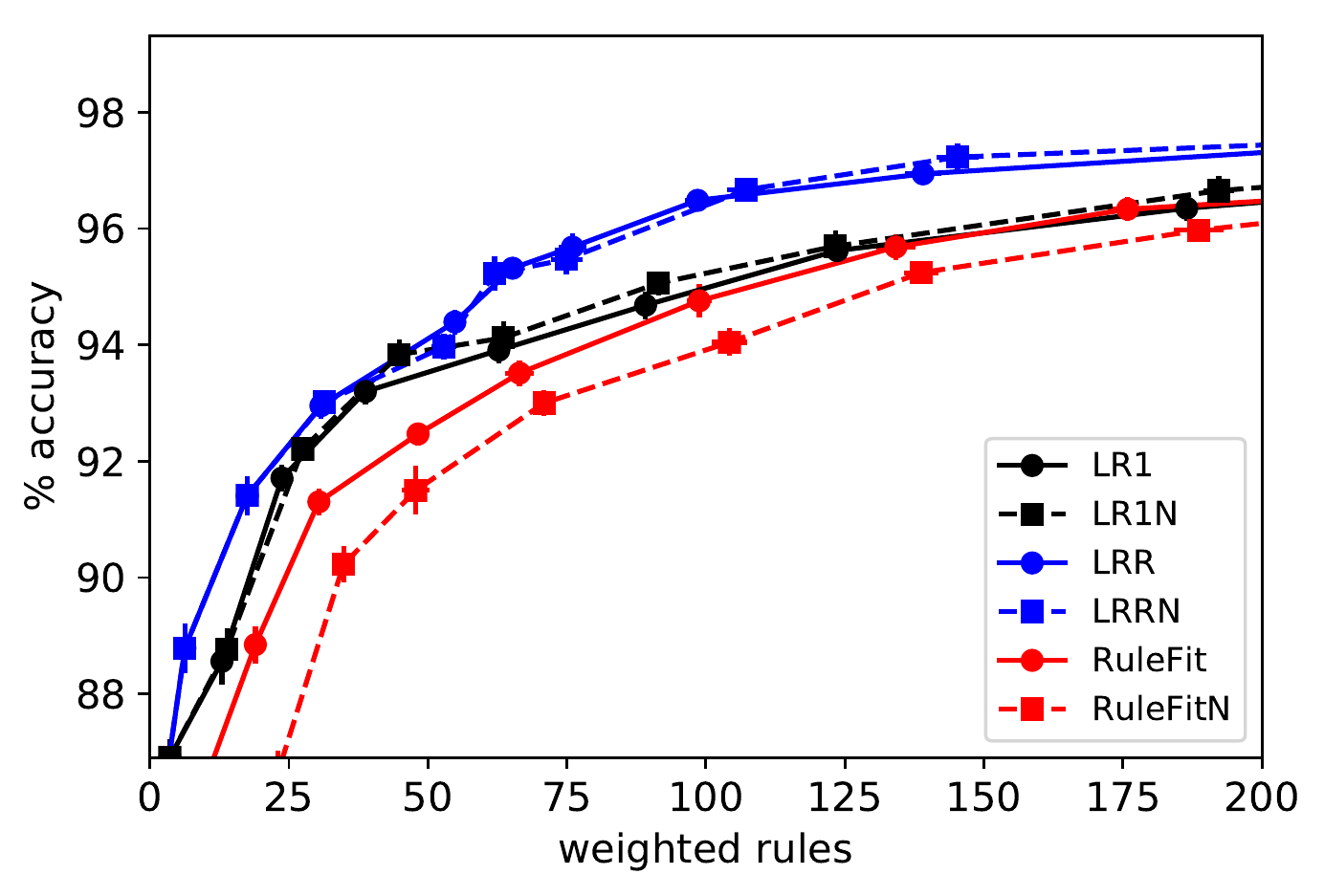}
  \caption{musk}
  \label{fig:paretoClassAcc:musk}
  \end{subfigure}
  \begin{subfigure}[b]{0.9\columnwidth}
  \includegraphics[width=0.9\columnwidth]{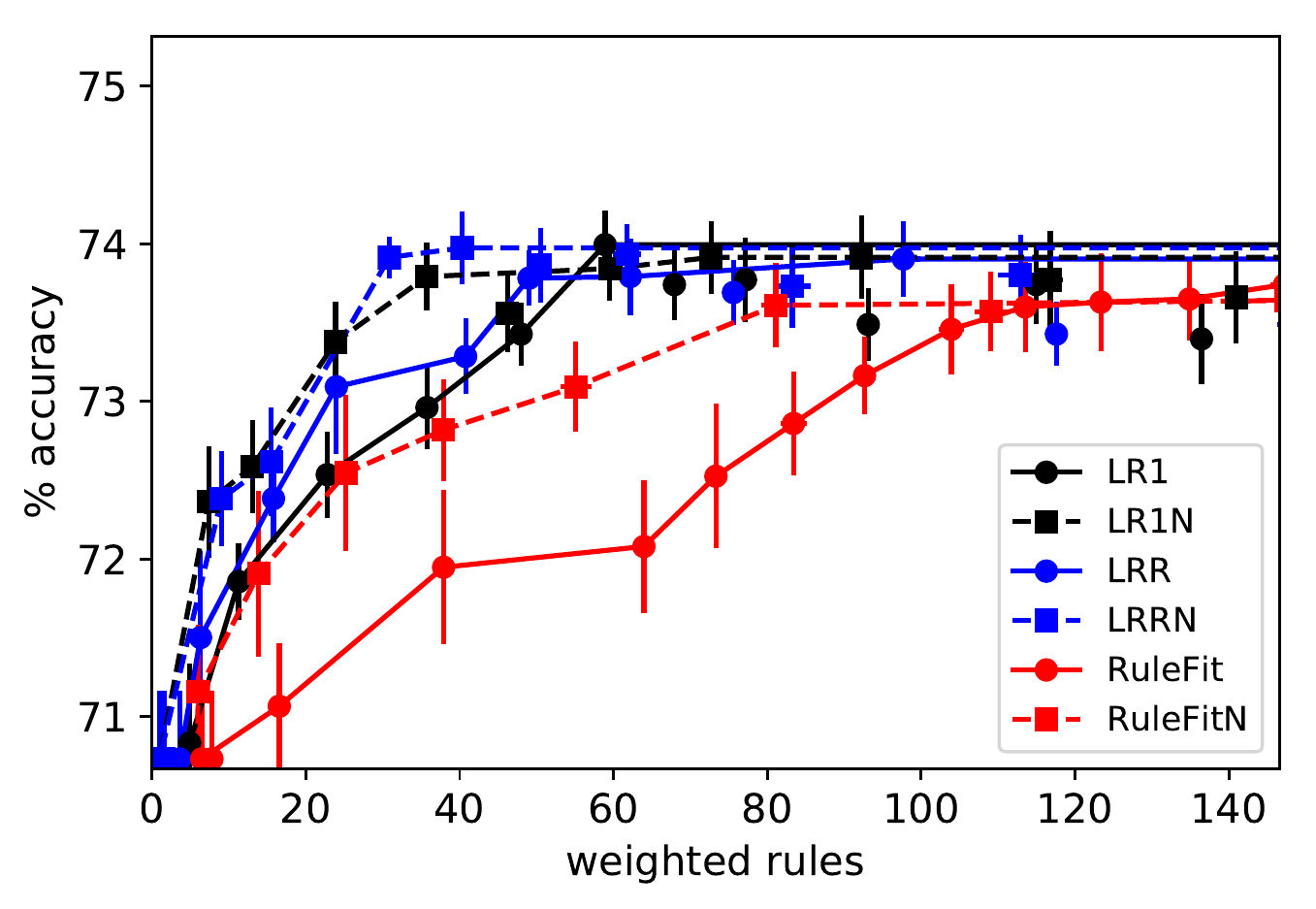}
  \caption{FICO}
  \label{fig:paretoClassAcc:FICO}
  \end{subfigure}
  \caption{Trade-offs between accuracy and weighted number of rules on classification datasets. Pareto efficient points are connected by line segments. Horizontal and vertical bars represent standard errors in the means.} 
  \label{fig:paretoClassAcc2}
\end{figure*}

Figures~\ref{fig:paretoClassBrier1}--\ref{fig:paretoClassAcc2} show trade-offs between Brier score and weighted rules and between accuracy and weighted rules for all $16$ classification datasets.

Tables~\ref{tbl:acc} and \ref{tbl:accWeighted} show mean test accuracies and corresponding complexities when the methods are optimized for accuracy. For Table~\ref{tbl:acc}, the Friedman statistic computed from the mean ranks is $9.74$ with a p-value of $0.202$, indicating no statistically significant differences in accuracy among the methods. For Table~\ref{tbl:accWeighted}, the Friedman statistic is $44.61$ (p-value $\sim 10^{-12}$). Post-hoc comparisons with LRRN as the reference show that RuleFit and RuleFitN are significantly more complex at the $0.05$ level using Holm's step-down procedure.

\begin{table*}[h]
\caption{Mean test accuracies (\%, standard error in parentheses). Best values in \textbf{bold}.}
\label{tbl:acc}
\begin{center}
\begin{small}
\begin{tabular}{lrrrrrrrr}
\toprule
dataset&\multicolumn{1}{c}{LR1}&\multicolumn{1}{c}{LRR}&\multicolumn{1}{c}{RuleFit}&\multicolumn{1}{c}{LR1N}&\multicolumn{1}{c}{LRRN}&\multicolumn{1}{c}{RuleFitN}&\multicolumn{1}{c}{GBT}&\multicolumn{1}{c}{SVM}\\
\midrule
banknote&$99.8$ ($0.1$)&$99.7$ ($0.1$)&$99.6$ ($0.1$)&$\mathbf{100.0}$ ($0.0$)&$99.9$ ($0.1$)&$99.9$ ($0.1$)&$99.7$ ($0.1$)&$99.9$ ($0.1$)\\
heart&$80.9$ ($1.6$)&$84.3$ ($2.0$)&$83.3$ ($1.3$)&$81.3$ ($1.8$)&$\mathbf{84.6}$ ($1.9$)&$83.3$ ($2.0$)&$82.3$ ($1.8$)&$82.6$ ($1.4$)\\
ILPD&$71.0$ ($1.1$)&$70.8$ ($0.5$)&$71.5$ ($0.1$)&$70.6$ ($0.9$)&$70.8$ ($0.7$)&$71.7$ ($1.1$)&$\mathbf{71.8}$ ($0.2$)&$71.7$ ($0.2$)\\
ionosphere&$91.2$ ($1.2$)&$91.2$ ($1.3$)&$93.4$ ($1.5$)&$91.7$ ($1.1$)&$90.9$ ($1.6$)&$94.3$ ($1.3$)&$91.2$ ($1.8$)&$\mathbf{94.9}$ ($1.4$)\\
liver&$\mathbf{61.2}$ ($2.0$)&$59.1$ ($2.2$)&$58.0$ ($2.2$)&$60.0$ ($2.6$)&$58.0$ ($2.7$)&$58.6$ ($2.1$)&$57.1$ ($2.5$)&$58.8$ ($2.7$)\\
pima&$75.5$ ($1.6$)&$75.1$ ($1.4$)&$75.5$ ($1.9$)&$\mathbf{77.7}$ ($1.3$)&$75.8$ ($1.6$)&$74.7$ ($1.9$)&$75.9$ ($1.9$)&$77.1$ ($2.0$)\\
tic-tac-toe&$98.3$ ($0.4$)&$98.0$ ($0.6$)&$\mathbf{100.0}$ ($0.0$)&$98.3$ ($0.4$)&$98.0$ ($0.6$)&$\mathbf{100.0}$ ($0.0$)&$99.1$ ($0.2$)&$98.3$ ($0.4$)\\
transfusion&$76.7$ ($0.3$)&$79.0$ ($0.9$)&$75.5$ ($1.9$)&$78.7$ ($0.7$)&$\mathbf{79.3}$ ($1.0$)&$74.7$ ($1.9$)&$76.6$ ($0.3$)&$76.9$ ($0.3$)\\
WDBC&$97.0$ ($0.6$)&$97.9$ ($0.5$)&$97.9$ ($0.4$)&$97.2$ ($0.7$)&$\mathbf{98.2}$ ($0.4$)&$96.8$ ($0.5$)&$95.6$ ($0.6$)&$98.1$ ($0.4$)\\
\midrule
adult&$84.9$ ($0.2$)&$84.9$ ($0.2$)&$84.8$ ($0.2$)&$85.8$ ($0.1$)&$85.9$ ($0.1$)&$\mathbf{87.0}$ ($0.2$)&$84.8$ ($0.2$)&$84.8$ ($0.1$)\\
bank-mkt&$88.7$ ($0.0$)&$90.0$ ($0.1$)&$88.7$ ($0.0$)&$88.7$ ($0.0$)&$\mathbf{90.1}$ ($0.1$)&$88.7$ ($0.0$)&$89.9$ ($0.1$)&$88.7$ ($0.0$)\\
gas&$99.5$ ($0.0$)&$99.6$ ($0.1$)&$99.5$ ($0.1$)&$\mathbf{99.6}$ ($0.0$)&$99.5$ ($0.1$)&$\mathbf{99.6}$ ($0.0$)&$99.4$ ($0.1$)&$99.5$ ($0.1$)\\
magic&$84.9$ ($0.3$)&$85.4$ ($0.3$)&$86.7$ ($0.2$)&$85.1$ ($0.3$)&$85.4$ ($0.2$)&$\mathbf{87.5}$ ($0.2$)&$87.2$ ($0.2$)&$87.4$ ($0.2$)\\
mushroom&$\mathbf{100.0}$ ($0.0$)&$\mathbf{100.0}$ ($0.0$)&$\mathbf{100.0}$ ($0.0$)&$\mathbf{100.0}$ ($0.0$)&$\mathbf{100.0}$ ($0.0$)&$\mathbf{100.0}$ ($0.0$)&$99.9$ ($0.1$)&$\mathbf{100.0}$ ($0.0$)\\
musk&$96.8$ ($0.5$)&$\mathbf{98.4}$ ($0.1$)&$97.6$ ($0.3$)&$96.1$ ($0.7$)&$98.4$ ($0.2$)&$97.8$ ($0.2$)&$94.5$ ($0.5$)&$97.6$ ($0.7$)\\
FICO&$73.8$ ($0.3$)&$73.8$ ($0.2$)&$73.8$ ($0.2$)&$\mathbf{74.0}$ ($0.2$)&$73.9$ ($0.2$)&$\mathbf{74.0}$ ($0.2$)&$73.3$ ($0.2$)&$72.4$ ($0.4$)\\
\midrule
mean rank&$5.25$&$4.31$&$4.84$&$4.16$&$3.78$&$\mathbf{3.69}$&$5.75$&$4.22$\\
\bottomrule
\end{tabular}
\end{small}
\end{center}
\vskip -0.1in
\end{table*}

\begin{table*}[h]
\caption{Mean weighted number of rules (standard error in parentheses) corresponding to Table~\ref{tbl:acc}. Best values in \textbf{bold}.}
\label{tbl:accWeighted}
\begin{center}
\begin{small}
\begin{tabular}{lrrrrrr}
\toprule
dataset&\multicolumn{1}{c}{LR1}&\multicolumn{1}{c}{LRR}&\multicolumn{1}{c}{RuleFit}&\multicolumn{1}{c}{LR1N}&\multicolumn{1}{c}{LRRN}&\multicolumn{1}{c}{RuleFitN}\\
\midrule
banknote&$32.3$ ($0.8$)&$47.2$ ($3.4$)&$57.8$ ($0.7$)&$\mathbf{16.4}$ ($0.4$)&$47.7$ ($1.6$)&$1124.9$ ($67.7$)\\
heart&$13.4$ ($2.7$)&$5.7$ ($0.6$)&$34.3$ ($0.9$)&$14.3$ ($2.1$)&$\mathbf{5.2}$ ($0.4$)&$59.4$ ($2.4$)\\
ILPD&$14.6$ ($3.7$)&$38.1$ ($25.5$)&$\mathbf{0.0}$ ($0.0$)&$14.7$ ($4.9$)&$1.9$ ($1.9$)&$2106.0$ ($30.3$)\\
ionosphere&$114.4$ ($28.5$)&$\mathbf{85.2}$ ($22.6$)&$1022.6$ ($64.9$)&$130.3$ ($23.7$)&$150.7$ ($49.3$)&$1225.6$ ($81.3$)\\
liver&$28.9$ ($5.8$)&$\mathbf{20.8}$ ($4.7$)&$66.7$ ($11.7$)&$25.7$ ($5.7$)&$34.7$ ($14.2$)&$89.7$ ($36.8$)\\
pima&$22.1$ ($2.3$)&$27.7$ ($1.8$)&$64.8$ ($1.1$)&$\mathbf{12.1}$ ($1.0$)&$15.5$ ($2.3$)&$3211.5$ ($83.3$)\\
tic-tac-toe&$\mathbf{21.6}$ ($0.0$)&$67.1$ ($3.5$)&$1640.7$ ($99.2$)&$\mathbf{21.6}$ ($0.0$)&$67.1$ ($3.5$)&$1640.7$ ($99.2$)\\
transfusion&$15.2$ ($4.3$)&$17.8$ ($1.1$)&$64.8$ ($1.1$)&$24.3$ ($2.5$)&$\mathbf{11.9}$ ($1.4$)&$3211.5$ ($83.3$)\\
WDBC&$145.7$ ($18.0$)&$283.6$ ($10.3$)&$809.4$ ($89.6$)&$\mathbf{86.1}$ ($12.6$)&$228.4$ ($30.3$)&$562.3$ ($83.3$)\\
\midrule
adult&$87.1$ ($1.6$)&$91.5$ ($4.8$)&$425.5$ ($35.0$)&$\mathbf{85.8}$ ($2.4$)&$94.2$ ($6.2$)&$719.9$ ($58.6$)\\
bank-mkt&$\mathbf{0.0}$ ($0.0$)&$68.6$ ($9.5$)&$\mathbf{0.0}$ ($0.0$)&$\mathbf{0.0}$ ($0.0$)&$83.6$ ($4.9$)&$0.2$ ($0.0$)\\
gas&$\mathbf{483.7}$ ($8.0$)&$678.2$ ($17.6$)&$2663.1$ ($235.9$)&$950.2$ ($12.9$)&$1259.4$ ($45.8$)&$2920.8$ ($125.7$)\\
magic&$\mathbf{93.1}$ ($2.2$)&$177.2$ ($17.5$)&$496.7$ ($5.9$)&$97.9$ ($2.9$)&$196.2$ ($25.0$)&$1656.0$ ($11.7$)\\
mushroom&$24.7$ ($0.6$)&$\mathbf{18.2}$ ($0.9$)&$927.9$ ($58.6$)&$24.7$ ($0.6$)&$\mathbf{18.2}$ ($0.9$)&$927.9$ ($58.6$)\\
musk&$\mathbf{263.0}$ ($39.3$)&$1002.0$ ($71.8$)&$1796.4$ ($326.1$)&$313.9$ ($101.6$)&$1079.7$ ($78.4$)&$2000.3$ ($314.8$)\\
FICO&$92.6$ ($5.9$)&$65.9$ ($3.4$)&$239.8$ ($2.5$)&$81.0$ ($5.2$)&$\mathbf{56.2}$ ($4.1$)&$183.4$ ($3.0$)\\
\midrule
mean rank&$\mathbf{2.25}$&$2.88$&$4.75$&$2.38$&$3.06$&$5.69$\\
\bottomrule
\end{tabular}
\end{small}
\end{center}
\vskip -0.1in
\end{table*}


\subsection{Regression}

\begin{figure*}[t]
  \centering
  \begin{subfigure}[b]{\columnwidth}
  \includegraphics[width=0.9\columnwidth]{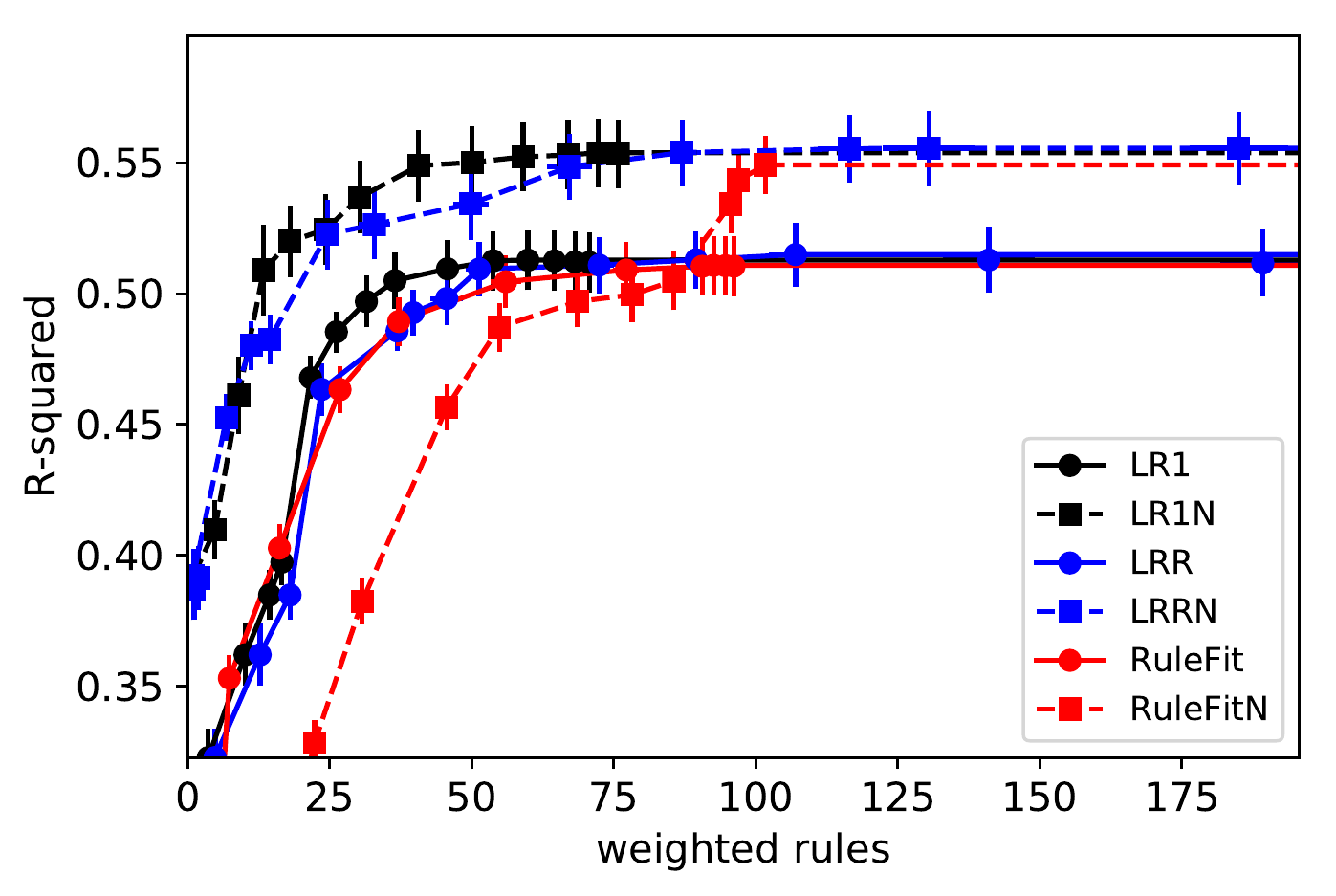}
  \caption{abalone}
  \label{fig:paretoRegress:abalone}
  \end{subfigure}
  \begin{subfigure}[b]{\columnwidth}
  \includegraphics[width=0.9\columnwidth]{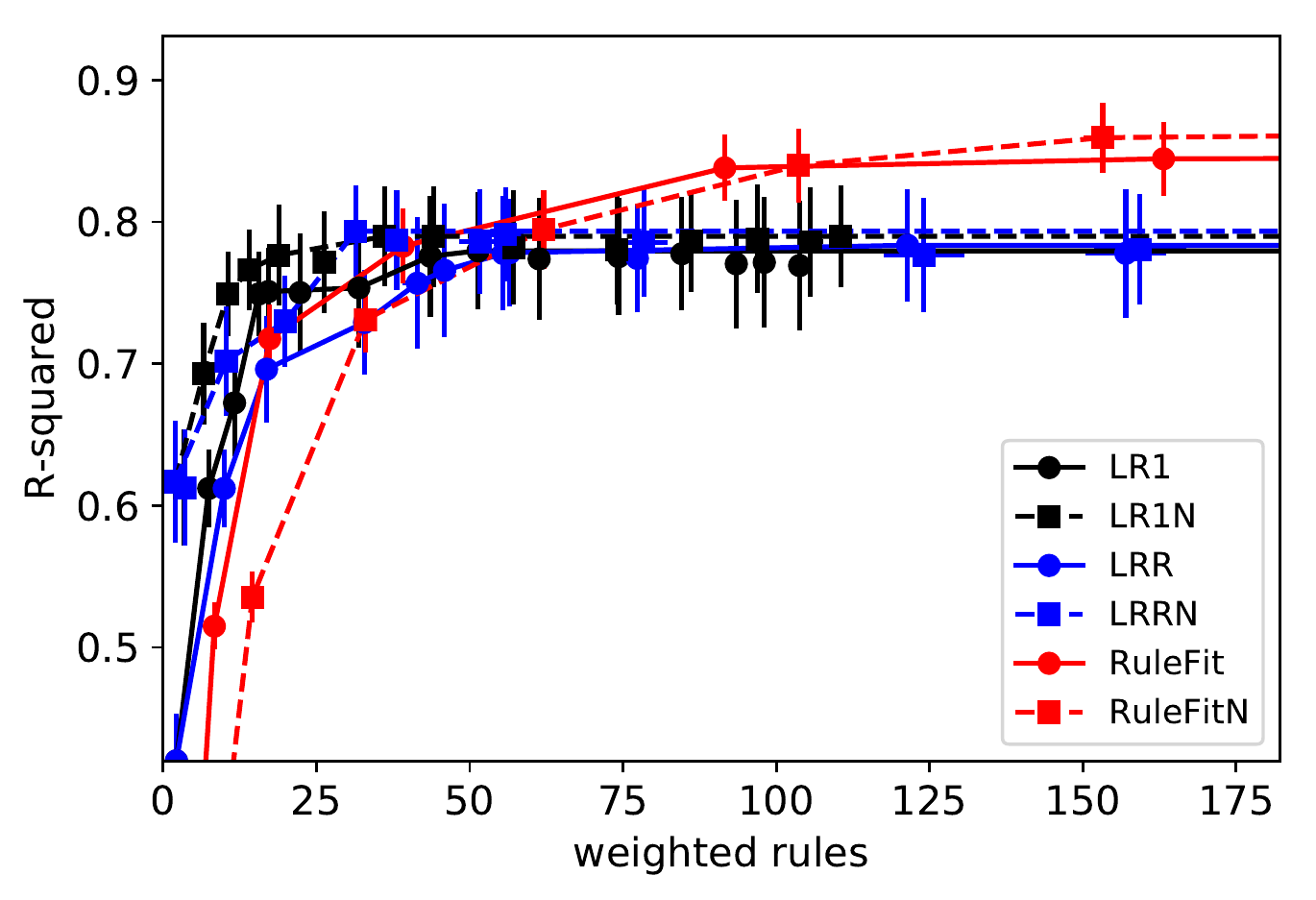}
  \caption{boston}
  \label{fig:paretoRegress:boston}
  \end{subfigure}
  \begin{subfigure}[b]{\columnwidth}
  \includegraphics[width=0.9\columnwidth]{paretoLinRR_R2_bike.pdf}
  \caption{bike}
  \label{fig:paretoRegress:bike}
  \end{subfigure}
  \begin{subfigure}[b]{\columnwidth}
  \includegraphics[width=0.9\columnwidth]{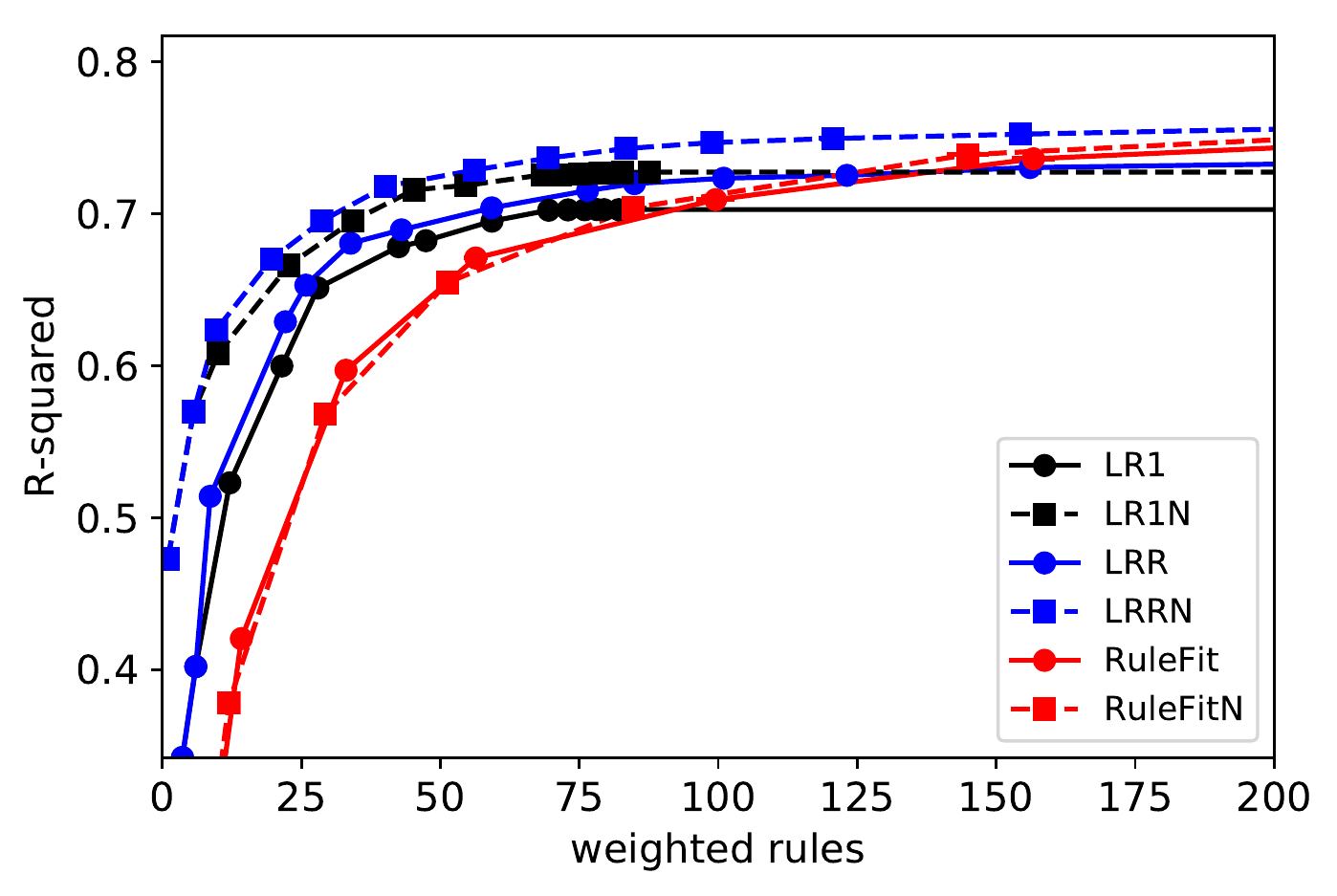}
  \caption{california}
  \label{fig:paretoRegress:california}
  \end{subfigure}
  \begin{subfigure}[b]{\columnwidth}
  \includegraphics[width=0.9\columnwidth]{paretoLinRR_R2_crime.pdf}
  \caption{crime}
  \label{fig:paretoRegress:crime}
  \end{subfigure}
  \begin{subfigure}[b]{\columnwidth}
  \includegraphics[width=0.9\columnwidth]{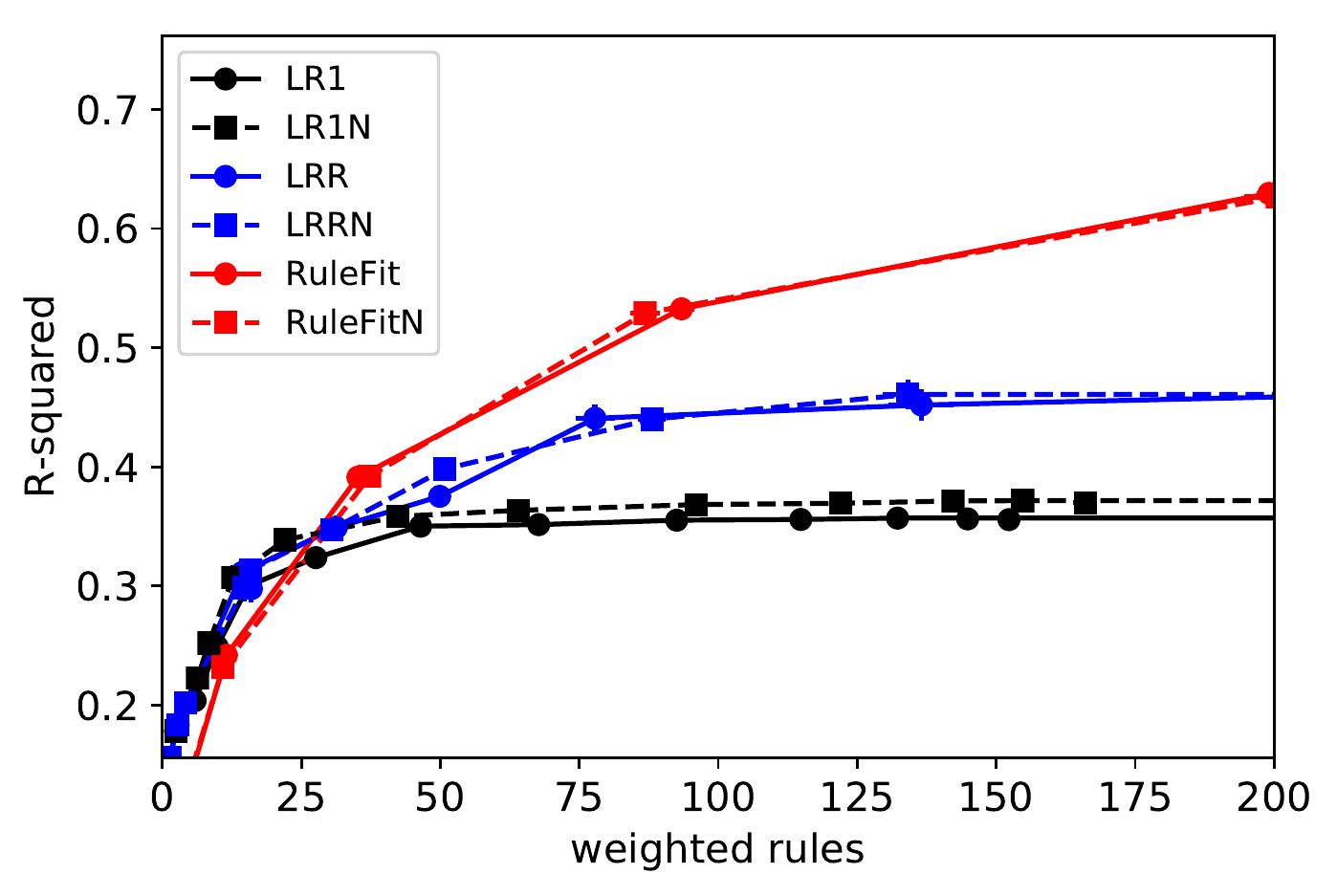}
  \caption{parkinsons}
  \label{fig:paretoRegress:parkinsons}
  \end{subfigure}
  \begin{subfigure}[b]{\columnwidth}
  \includegraphics[width=0.9\columnwidth]{paretoLinRR_R2_wine.pdf}
  \caption{wine}
  \label{fig:paretoRegress:wine}
  \end{subfigure}
  \begin{subfigure}[b]{\columnwidth}
  \includegraphics[width=0.9\columnwidth]{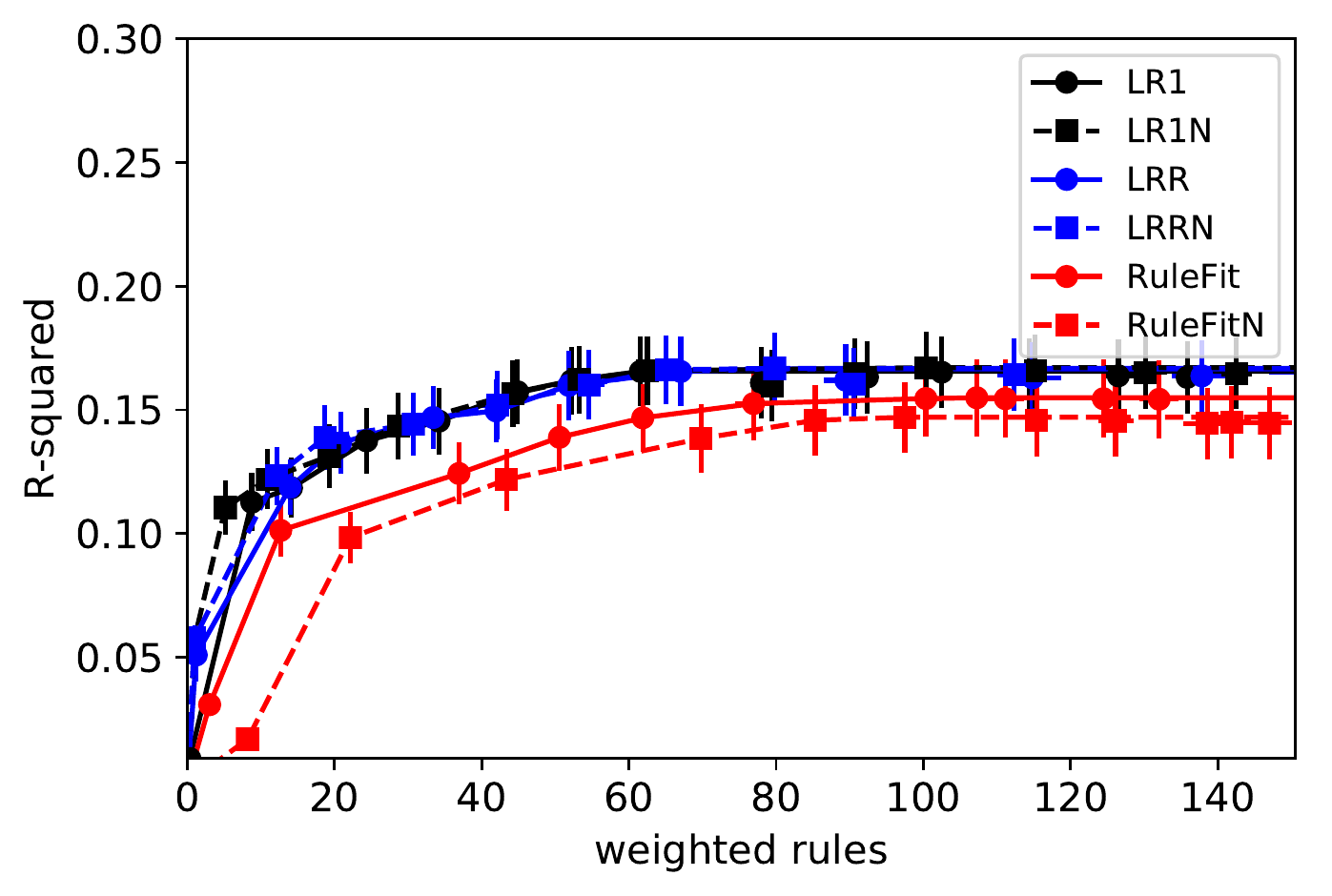}
  \caption{MEPS}
  \label{fig:paretoRegress:meps}
  \end{subfigure}
  \caption{Trade-offs between coefficient of determination $R^2$ and weighted number of rules on regression datasets. Pareto efficient points are connected by line segments. Horizontal and vertical bars represent standard errors in the means.} 
  \label{fig:paretoRegress2}
\end{figure*}

Figure~\ref{fig:paretoRegress2} shows the trade-off between $R^2$ and weighted rules for all $8$ regression datasets.

